\documentclass[a4paper,fleqn]{cas-dc}

\usepackage[numbers]{natbib}

\usepackage{graphicx}%
\usepackage{float}
\usepackage{multirow}%
\usepackage{amsmath,amssymb,amsfonts}%
\usepackage{amsthm}%

\usepackage[title]{appendix}%

\usepackage{xcolor}%
\usepackage{textcomp}%
\usepackage{manyfoot}%
\usepackage{booktabs}%

\usepackage{listings}%

\usepackage{comment}
\usepackage{tikz} 
\usepackage{enumitem}
\usepackage{multirow}
\usepackage{threeparttable}
\usepackage{adjustbox} 
\usepackage{longtable}
\usepackage{subcaption}
\usepackage[capitalise]{cleveref}
\usepackage{array}
\usepackage{url}
\usepackage{gensymb} 

\usepackage[ruled,vlined]{algorithm2e}
\crefname{algorithm}{Algorithm}{Algorithms}
\crefname{algocf}{Algorithm}{Algorithms}

\usepackage[T1]{fontenc}

\def\tsc#1{\csdef{#1}{\textsc{\lowercase{#1}}\xspace}}
\tsc{WGM}
\tsc{QE}

\begin{document}

\let\WriteBookmarks\relax
\def\floatpagepagefraction{1}
\def\textpagefraction{.001}
\shorttitle{Addressing Key Challenges of Adversarial Attacks and Defenses in the Tabular Domain}
\shortauthors{Y. Itzhakev et~al.}

\title[mode = title]{Addressing Key Challenges of Adversarial Attacks and Defenses in the Tabular Domain: A Methodological Framework for Coherence and Consistency}

\author[1]{Yael Itzhakev}[type=editor,
                        prefix=,
                        orcid=0009-0009-9854-8480
                        ]
\cormark[1]
\fnmark[1]
\ead{freidiya@post.bgu.ac.il}

\credit{Writing – review \& editing, Writing – original draft, Visualization, Data curation ,Investigation, Formal analysis, Validation, Software, Methodology, Conceptualization}

\affiliation[1]{organization={Department of Software and Information Systems Engineering},
                addressline={Ben-Gurion University of the Negev}, 
                city={Beer-Sheva},
                postcode={8410501}, 
                country={Israel}}

\author[1]{Amit Giloni}
\credit{Writing – review \& editing, Validation, Conceptualization}

\author[1]{Yuval Elovici}
\credit{Supervision}

\author[1]{Asaf Shabtai}
\credit{Supervision, Project administration}

\cortext[cor1]{Corresponding author}

\begin{abstract}
Machine learning models trained on tabular data are vulnerable to adversarial attacks, even in realistic scenarios where attackers only have access to the model's outputs. Since tabular data contains complex dependencies among features, it presents a unique challenge for adversarial samples which must maintain coherence and respect these dependencies to remain indistinguishable from benign data. Moreover, existing attack evaluation metrics—such as the success rate, perturbation magnitude, and query count-fail to account for this challenge.
To address these gaps, we propose a technique for perturbing dependent features while preserving sample coherence. In addition, we introduce Class-Specific Anomaly Detection (CSAD), an effective novel anomaly detection approach, along with concrete metrics for assessing the quality of tabular adversarial attacks. CSAD evaluates adversarial samples relative to their predicted class distribution, rather than a broad benign distribution. This ensures that subtle adversarial perturbations, which may appear coherent in other classes, are correctly identified as anomalies. We extend CSAD for importance-based anomaly detection by integrating SHAP explainability techniques to detect inconsistencies in model decision-making.
Our evaluation of adversarial sample quality incorporates both anomaly detection rates and importance-based assessments to provide a more comprehensive measure.
We evaluate various attack strategies, examining black-box query-based and transferability-based gradient attacks across four target models. Experiments on benchmark tabular datasets reveal key differences in the attacker's risk and effort and attack quality, offering insights into the strengths, limitations, and trade-offs faced by attackers and defenders. Our findings lay the groundwork for future research on adversarial attacks and defense development in the tabular domain.
\end{abstract}


\maketitle

\section{Introduction}\label{intro}

Machine learning (ML) models trained on tabular data are widely deployed in diverse industry sectors where they are used for a wide range of tasks, including credit risk assessment in finance~\cite{crouhy2000comparative,moro2011using}, the analysis of administrative claims and patient registries in healthcare~\cite{danese2019generalized}, and the prediction of energy consumption patterns in energy~\cite{lam2008principal,ramesh2010life}. However, recent research has highlighted their vulnerability to adversarial attacks, in which malicious data samples are used to mislead the model into producing incorrect outputs~\cite{ballet2019imperceptible,cartella2021adversarial,mathov2022not,grolman2022hateversarial}. Such attacks can even be conducted in realistic black-box settings, where the attacker has no knowledge of the model's internals and can only query the model to receive outputs. 

The core challenge in adversarial attacks on tabular data lies in maintaining realistic data relationships while achieving misclassification~\cite{mathov2022not, grolman2022hateversarial}. Unlike domains such as images, audio, or text, which contain homogeneous feature types with well-defined dependencies (spatial, temporal, or semantic respectively), the tabular domain presents unique complexities. Features in tabular datasets are heterogeneous, subject to diverse constraints, and exhibit complex interdependencies that must be preserved to avoid detection~\cite{cartella2021adversarial, mathov2022not, grolman2022hateversarial}.

Consider a loan approval scenario: an applicant may slightly increase their reported income to cross a decision threshold, changing the model's decision from `reject' to `approve.' However, this modification must remain consistent with other financial indicators to appear legitimate. This illustrates the fundamental challenge—crafting adversarial samples that are both effective and undetectable.

While extensive research has been performed on images~\cite{xu2020adversarial,zizzo2019adversarial,kaviani2022adversarial,puttagunta2023adversarial}, audio~\cite{lan2022adversarial}, text~\cite{xu2020adversarial,marulli2021exploring,goyal2023survey}, and graph data ~\cite{xu2020adversarial,sun2022adversarial}, the tabular domain remains relatively underexplored. This paper addresses three critical gaps in tabular adversarial research.

\textbf{First, research on preserving feature dependencies and coherence in adversarial attacks is limited.}

Models trained on tabular data often rely on interdependent features, where the values of some features (dependent features) are influenced by others (influencing features). Therefore, in adversarial attacks, perturbations to dependent features must align with the influencing features to preserve these constraints and interdependencies, maintaining coherence and consistency to avoid detection. For instance, the body mass index (BMI) is derived from height and weight~\cite{keys2014indices}, so when altering one of these influencing features (i.e., height or weight) during an attack, it is essential to ensure that the BMI value remains mathematically consistent with the changes.

In practice, correlations in tabular data can be complex. In the loan approval scenario, for example, a high credit score strongly correlates with approval, but for a low credit score, approval depends on other factors like income or debt, making the correlation conditional. In another practical case, such as employee performance ratings which depend on multiple factors such as project completion rates, client satisfaction scores, and peer feedback, the dependencies are more complex, and may be non-linear, indirect, or influenced by features unknown to the attacker. Preserving such dependencies during perturbation poses a significant challenge.

\textbf{Second, current attack evaluation methods are not adapted to tabular data.}
Existing approaches primarily focus on measuring the perturbation magnitude using $L_p$ norms~\cite{ballet2019imperceptible,cartella2021adversarial,mathov2022not}, which aim to indicate the detectability risk (i.e., the risk that the attack will be exposed). However, in the context of tabular data, limiting perturbation magnitudes alone is insufficient. Effective adversarial attacks must ensure that samples remain coherent, consistent, and in-distribution to evade detection. While previous work has acknowledged these challenges (see ~\cref{related_imperceptible}), there is a lack of concrete evaluation methods for quantifying the distinguishability of adversarial samples in the tabular domain.

\textbf{Third, empirical analysis of attack strategies in the tabular domain is limited.} Two common black-box attack strategies exist: query-based attacks, where attackers are assumed to have unlimited access to model predictions and iteratively optimize the adversarial samples by repeatedly querying the model~\cite{cartella2021adversarial}, and transferability-based attacks, where attackers lack direct access to the target model's outputs but train surrogate models using available data~\cite{mathov2022not,grolman2022hateversarial}. In the latter case, the attacker optimizes the adversarial sample to mislead the \emph{surrogate} model and then uses this sample to mislead the target model with a single query. 

Both attackers and defenders need to understand the practical trade-offs between these approaches, yet systematic comparison in the tabular domain remains limited. This understanding is crucial for both groups who use adversarial samples and may choose different attack strategies based on their goals, resources, and constraints;
\emph{Defenders} interested in assessing their model's vulnerability often use adversarial attacks for automated penetration testing or adversarial training, which involves creating adversarial samples to improve the model's robustness~\cite{chen2022improving,haroon2022adversarial}. Since defenders have full access to their model and can make unlimited queries, query-based attacks are a practical option.

In contrast, \emph{attackers} aiming to minimize the number of queries performed and avoid detection, might prefer transferability-based attacks that use gradient methods on neural networks, as they capture underlying feature interactions which is necessary for subtle perturbations. However, systematic comparison of these attack strategies in the tabular domain remains absent.

To address these gaps and advance adversarial research in the tabular domain, a novel and comprehensive methodological framework that combines attack design and evaluation tailored specifically to tabular data characteristics was used.

To address the first gap of maintaining coherence in dependent features, this research develops a novel technique for consistent perturbation, using \textit{regression models} to preserve realistic relationships in features with complex dependencies. This technique is designed for scenarios where these dependencies are not circular, ensuring that modifications maintain the inherent structure and correlations in the data.

To address the second gap of current evaluation methods' limitations, the paper proposes two complementary evaluation criteria for assessing the quality of adversarial samples. The first criterion detects anomalies in feature space values, quantifying structural deviations from the data distribution. The second examines how adversarial samples influence the model's decision-making process, employing SHapley Additive exPlanations (SHAP) \cite{lundberg2017unified}, a widely adopted explainability (XAI) technique \cite{belle2021principles}, to quantify inconsistencies in feature importance caused by adversarial perturbations. These measures address the limitations of relying solely on perturbation magnitude as an indicator of detectability.

Both criteria are employed using the proposed \textit{Class-Specific Anomaly Detection} (CSAD) approach, which evaluates of adversarial samples relative to the benign distribution of their \textit{predicted class}, rather than relying on global thresholds. This class-aware perspective improves sensitivity to inconsistencies in the tabular domain, where perturbations can be subtle and might be overlooked in global analyses.

To address the third gap of limited systematic comparison of attack strategies, a comprehensive evaluation of various black-box attack methods in the tabular domain is conducted. The evaluation integrates traditional metrics for computing attacker risk (success rate, query count, and $L_p$ norms) and attacker effort, complemented by the proposed quality criteria: sample anomaly detection and feature importance consistency. This thorough evaluation provides insights into both attack effectiveness and adversarial sample quality that can guide the development of more robust attacks and defenses.

The experiments performed in this research aim to answer the following research questions:
\begin{itemize}
\item \textbf{RQ\#1 } How do the perturbed samples of different attack strategies differ in terms of the attacker's risk and effort? 
\item \textbf{RQ\#2 } To what extent do query-based attacks reflect the behavior of transferability-based gradient attacks?
\item \textbf{RQ\#3 } For both benign samples and perturbed samples, is the attacked model’s decision-making process aligned with the expected feature importance?
\end{itemize}

\begin{figure*}
\centering
\includegraphics[width=0.97\textwidth]{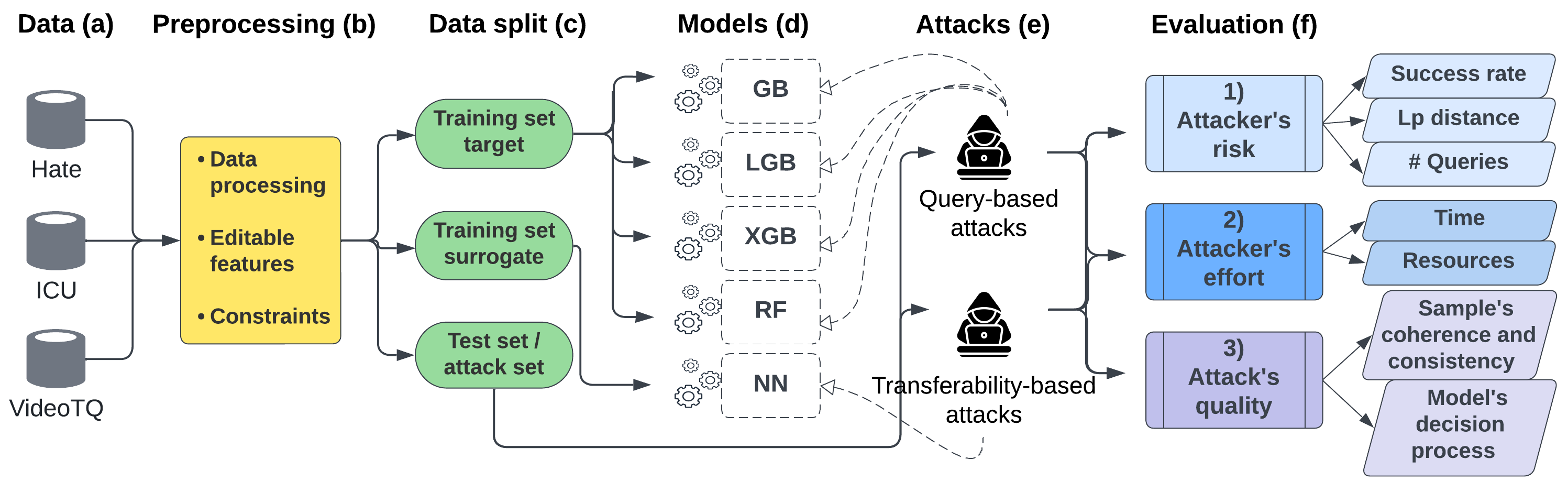}
\caption{An overview of the paper's methodological framework.}
\label{fig:flow}
\end{figure*}

The evaluation performed examines seven attack strategies across four tree-based classification models using three datasets: the Hateful Users on Twitter~\cite{ribeiro2018characterizing} and Intensive Care Unit~\cite{hanberger2005intensive} public datasets, which are benchmark datasets in the tabular domain, and a proprietary Video Transmission Quality dataset. The examined attack strategies consist of two query-based attacks (boundary and HopSkipJump~\cite{brendel2017decision,chen2020hopskipjumpattack}) and five transferability-based gradient attacks with varying feature selection techniques~\cite{mathov2022not,grolman2022hateversarial}. The process illustrated in ~\cref{fig:flow}.

The contributions of this work can be summarized as follows:
\begin{enumerate}
\sloppy
    \item Proposing a novel technique for perturbing interdependent features in tabular data, leveraging regression models to preserve coherence and ensure feature consistency.
    \item Introducing CSAD, an approach that assesses anomalies relative to class-specific distributions rather than global thresholds, which improves sensitivity to feature inconsistencies.
    \item Developing comprehensive dual evaluation criteria based on anomaly assessment of the sample values and the feature importance, providing a principled framework for quantifying adversarial sample quality and its impact on a model's decision-making process.
    \item Systematically comparing adversarial attack strategies' performance factors in the tabular domain, providing critical insights into their effectiveness, limitations, and corresponding defense strategies. To the best of our knowledge, this is the first study to do this.
\end{enumerate}

The remainder of this paper is organized as follows: \cref{background} provides an overview of related works. \Cref{proposed_approach}, which describes the methodological framework, is followed by \cref{sec:exp_setup} and \cref{sec:Results}, which present the experimental setup and provide a thorough analysis of the examined query- and transferability-based attack behavior. \Cref{sec:discussion} analyzes the trade-offs between different attack strategies, examines the implications for both attackers and defenders, and evaluates the effectiveness of detection mechanisms in tabular data domains. Finally, \cref{conc} concludes the paper, providing key insights and future research directions for attackers and defenders in tabular data domains.

\section{Background and Related Work}\label{background}

\subsection{Adversarial Attacks in the Tabular Domain} \label{tabular_attacks}

Szegedy \textit{et al.}~\cite{szegedy2013intriguing} introduced the concept of an adversarial attack in which an attacker provides a malicious data sample aimed at misleading the ML model and causing it to produce incorrect output.
Their adversarial sample was crafted by performing subtle changes to an image, which were imperceptible to the human eye but misled the target model.
However, research conducted in tabular subdomains, such as fraud detection and recommendation systems~\cite{ballet2019imperceptible,cartella2021adversarial,deldjoo2021survey}, has highlighted  critical challenges within the tabular domain. These challenges include the constraints of editable and non-editable features (e.g., identification number, date of birth), data imbalance, and the presence of noncontinuous features (e.g., categorical features). The authors of these studies also mentioned the complexity of perturbing discrete values while maintaining the semantic integrity of the sample, as well as the challenge of capturing nuances when designing attacks. This discussion was further elaborated on in~\cite{mathov2022not}.

Most recent adversarial attacks operate under black-box conditions, where attackers lack prior knowledge of the target model's architecture and parameters, possessing only query access~\cite{mahmood2021back}. Two main types of black-box attacks are query-based attacks and transferability-based attacks.
In query-based attacks, including decision-based methods like the boundary Attack and HopSkipJump~\cite{brendel2017decision,chen2020hopskipjumpattack}, attackers iteratively adjust sample values based on model output until the optimization objective is fulfilled - misleading the target model to produce the desired output. These attacks typically begin with large initial modifications to the original sample, causing incorrect predictions, then optimize the modified sample to remain as close as possible to the original while maintaining the deception.
In contrast, transferability-based attacks leverage surrogate models to craft an adversarial sample applying a white-box attack on the surrogate model, and then utilize the adversarial sample crafted to attack the target model in a single query~\cite{papernot2017practical}.

Recent developments in tabular adversarial attacks have focused on addressing domain-specific constraints identified in earlier work. Cartella et al.  ~\cite{cartella2021adversarial} addressed the limitation of applying image-based attacks to tabular data by modifying decision-based attacks to incorporate tabular data validity and editability constraints. Building on this foundation, Mathov et al. ~\cite{mathov2022not} tackled the challenge of maintaining feature correlations, proposing an untargeted, transferability-based gradient attack that incorporats neural network (NN) embedding functions to preserve feature correlations and ensure value consistency by minimizing the $L_2$ distance and adjusting correlated features. Grolman et al.~\cite{grolman2022hateversarial} extended this approach by integrating feature importance into the perturbation process, selecting the most important features for perturbation. Ju et al. ~\cite{ju2022robust} further advanced the field by addressing feature distribution challenges, developing a transferability-based approach that considers feature distribution in the perturbation process, although validity constraints remained unaddressed.

\subsection{Evaluation Challenges and Anomaly Detection} \label{related_imperceptible}

Current evaluation frameworks for adversarial attacks in various domains rely primarily on four dimensions: attack success rates~\cite{goodfellow2014explaining,ballet2019imperceptible}; the number of queries made to the target model~\cite{ilyas2018black,wang2020towards}; computational efficiency, measured by the time required to create the adversarial sample, as well as the amount of data and time needed to train the surrogate model~\cite{alecci2023your}; and perturbation magnitude, typically quantified through $L_{p}$ norms such as $L_{0}$, $L_{2}$, and $L_{\infty}$~\cite{goodfellow2014explaining,kurakin2018adversarial,fezza2019perceptual,ballet2019imperceptible,mathov2022not,grolman2022hateversarial} or the average error between the original and adversarial samples~\cite{gomez2023vaasi}.

In domains such as computer vision, perturbation magnitude is often used as an indicator of the imperceptibility of an adversarial sample. Specifically, an adversarial sample is considered indistinguishable from a benign sample if the human eye cannot perceive any difference from the original~\cite{szegedy2013intriguing,goodfellow2014explaining,papernot2016limitations,kurakin2018adversarial}. In such cases, measuring perturbation magnitude through norms such as $L_{2}$ or $L_{\infty}$ is meaningful, because these norms capture relative changes in pixel values that directly influence visual detectability.

However, there are inherent limitations when employing perturbation magnitude as an evaluation measure in the tabular domain. While $L_{0}$, which counts the number of modified features, is particularly relevant, as it reflects the practical feasibility and potential cost of modifying specific attributes in real-world structured datasets, the applicability of $L_{2}$ and $L_{\infty}$ is restricted. These norms are only meaningful when all input features are numeric and share a common scale. More importantly, unlike images, where human visual inspection can directly assess perturbations, identifying such changes in tabular data requires domain expertise~\cite{ballet2019imperceptible}.

Existing works in the tabular domain have attempted to address this by involving domain experts who examine a small subset of ``important'' features, selected based on their significance to the model, to detect manipulations~\cite{ballet2019imperceptible,cartella2021adversarial}. However, this approach is neither objective, scalable, nor practical, as it requires manual assessment of each sample individually and does not account for the fact that altering even a small number of features can cause inconsistencies with the rest of the sample, making it distinguishable from benign data distributions.

Moreover, in the tabular domain, imperceptibility should not rely on the perturbation size but should instead critically depend on maintaining coherence and preserving the natural relationships and dependencies between features. Existing metrics fail to capture a sample's coherence and feature consistency, both of which are essential for determining the true distinguishability of tabular adversarial samples.

To address these evaluation challenges, we propose using anomaly detection as a practical and scalable approach for the objective assessment of the distinguishability of adversarial samples in the tabular domain. The underlying assumption is that adversarial samples lacking structural coherence or consistency with benign data distributions are more likely to exhibit anomalous patterns, making them both distinguishable and detectable. However, the selection of suitable anomaly detection techniques for tabular data requires careful consideration of domain-specific constraints and data characteristics.

Traditional anomaly detection methods developed for image domains often prove inadequate for tabular data. Convolutional autoencoders (CAEs) and generative adversarial networks (GANs), while effective for spatial correlations, struggle to capture the complex multivariate dependencies characteristic of tabular data \cite{borisov2022deep}. Classical statistical approaches like t-tests and the ANOVA (analysis of variance) assume normality and independence, limiting their applicability to tabular data domains where variables might not follow normal distributions and exhibit complex, non-linear relationships\cite{fisher1971design}. Simple univariate approaches, such as z-score (standard score) thresholding, focus on individual feature deviations but fail to account for multivariate dependencies \cite{sakurada2014anomaly}. Similarly, principal component analysis (PCA)-based methods assume linear relationships, limiting their effectiveness in detecting non-linear dependencies in high-dimensional datasets\cite{sakurada2014anomaly}.

More suitable approaches for tabular data include autoencoders (AEs), which learn latent representations to identify deviations from expected patterns~\cite{sakurada2014anomaly}, and isolation forest (IF) algorithms, which detect anomalies by isolating points in a partitioned feature space. However, while IF excels at detecting structurally shifted or distant outliers, it may exhibit reduced sensitivity to some subtle inconsistencies that may characterize sophisticated adversarial attacks~\cite{xu2023deep}.

It should be noted that the anomaly detection framework proposed here serves as a methodological foundation for assessing adversarial sample quality and can be extended to incorporate additional anomaly detection techniques as needed. Alternative approaches such as support vector machines (SVMs) with one-class classification, local outlier factor (LOF) methods, or more recent deep learning-based anomaly detection techniques could similarly be integrated into this evaluation framework. The choice of specific anomaly detection methods can be adapted based on dataset characteristics, computational constraints, and the particular nature of adversarial attacks under investigation, providing flexibility for future research applications.

In this study, IF and AEs were selected as primary anomaly detection baselines, motivated by their complementary strengths and established effectiveness in the tabular domain. IF provides computational efficiency and strong performance on high-dimensional data without requiring distributional assumptions, making it suitable for diverse tabular datasets. Conversely, AEs can model complex feature interactions and distributional shifts that may characterize sophisticated adversarial perturbations. This dual approach enables the comprehensive evaluation of adversarial sample quality across different types of anomalous patterns.

\subsection{Research Gap and Contributions.}

\cref{tab:related_works} summarizes related studies that developed adversarial attacks, taking into account the unique constraints of the tabular domain.
Despite significant advances in tabular adversarial attacks, prior research shares three critical limitations that this work addresses:

\begin{table*}
\caption{Related works that developed adversarial attacks, taking into account the unique constraints of the tabular domain.}
\centering
\begin{tabular}{@{}>
{\raggedright\arraybackslash}p{0.15\textwidth}>{\raggedright\arraybackslash}p{0.35\textwidth}>{\raggedright\arraybackslash}p{0.18\textwidth}>{\raggedright\arraybackslash}p{0.25\textwidth}@{}}

\toprule
\textbf{Paper \newline(Attack Type)} & \textbf{Key Contribution} & \textbf{Evaluation Metrics} & \textbf{Limitation} \\
\midrule

\cite{ballet2019imperceptible} \newline(White-box) & 
Incorporates feature validity constraints and minimizes perturbations on important features to reduce imperceptibility & 
Success rate, \newline weighted $L_2$ norm (scaled by feature importance), \newline distance to nearest neighbor &
Drops non-ordered categorical features; no evaluation of feature coherence \\\midrule

\cite{cartella2021adversarial} \newline(Black-box) & 
Extends the weighted $L_2$ norm from \cite{ballet2019imperceptible} to minimize perturbations on both important and unimportant (irrelevant) features; adapts query-based attacks with validity constraints and custom thresholds for unbalanced datasets & 
Success rate, \newline important feature change count, \newline validity satisfaction &
Limited to validity constraints; no coherence evaluation \\\midrule

\cite{mathov2022not} \newline(White-box, Transferability) & 
Uses NN embeddings to minimize $L_2$ distance while preserving feature correlations & 
Success rate, \newline $L_0$ distance &
No evaluation of semantic coherence \\\midrule

\cite{grolman2022hateversarial} \newline(White-box, Transferability) & 
Extends previous work by incorporating feature importance for perturbation selection, reducing $L_0$ distance & 
Success rate, \newline $L_0$ distance &
No evaluation of semantic coherence \\\midrule

\cite{ju2022robust} \newline(Transferability) & 
Develops a transferability-based approach considering feature distributions across different models & 
Success rate, \newline transferability improvement &
No validity or coherence constraints; focused solely on transferability performance\\

\bottomrule
\end{tabular}
\label{tab:related_works}
\end{table*}

\textbf{Evaluation Framework Limitations:} Existing approaches lack objective metrics for assessing adversarial sample quality in the tabular domain, relying on subjective expert judgment or conventional metrics that fail to examine sample coherence and consistency. This work develops comprehensive evaluation criteria based on anomaly detection assessment of sample values and feature importance, providing a principled framework to quantify adversarial sample quality and its impact on model decision-making processes.

\textbf{Anomaly Detection Limitations:} Existing anomaly detection methods assess samples against global data distributions, failing to account for the fact that in tabular data, small adversarial changes may not be anomalous relative to their original class but can be anomalous relative to the target class they impersonate after the attack. We introduce CSAD, which assesses anomalies relative to class-specific distributions rather than global thresholds, better capturing the nuanced nature of tabular adversarial perturbations. This approach is further elaborated in \cref{csad}.

\textbf{Incomplete Comparative Analysis:} Recent research lacks systematic comparison of adversarial attack strategies' performance factors in the tabular domain. This work provides a comprehensive comparison of attack effectiveness and limitations across multiple attack types and datasets, offering valuable insights into their relative strengths and weaknesses.

By establishing these foundations, this research enables more informed attack design and evaluation development in tabular machine learning domains, contributing novel evaluation methodologies and comprehensive empirical insights to the field.

\section{Methodological Framework for Addressing Challenges in Adversarial Attacks and Evaluation in the Tabular Domain}\label{proposed_approach}

In this section, we introduce the black-box query- and transferability-based attacks evaluated, along with the modifications made in order to deal with the tabular constraints. We also present our novel technique for perturbing dependent features, as well as two criteria for evaluating the quality of adversarial attack samples. These criteria focus on the distinguishability of the adversarial samples and their impact on the target model's internal behavior in the tabular domain.

\subsection{Modifying Adversarial Attacks to Address Tabular Constraints.}
In our evaluation (see \cref{subsec:attack_setup}), we examine two decision-based attacks: the boundary attack~\cite{brendel2017decision} and the HopSkipJump attack~\cite{chen2020hopskipjumpattack}, which were adjusted to address tabular constraints and evaluated by ~\cite{cartella2021adversarial}.
We implement these attacks by applying additional constraints based on feature distributions (e.g., limiting values according to statistical information) and making adjustments to ensure validity (e.g., [0,1] values for binary features and ordinal values for categorical features) and editability (guided by domain knowledge).

In addition, we perform transferability-based gradient attacks based on the architecture proposed in~\cite{mathov2022not,grolman2022hateversarial}, which is a sophisticated attack designed to craft valid and consistent adversarial samples.

The details and pseudocode for the boundary and HopSkipJump attacks, as well as the transferability-based attacks, adapted to take into account the constraints of tabular data, are provided in Appendix~\ref{appendix:appendixA}.

\subsection{Perturbing Dependent Features.}\label{regretion_models}

To ensure coherence and maintain feature consistency when crafting adversarial samples, we introduce a novel technique for perturbing dependent features. Specifically, for each dependent feature, we train a regression model on the benign samples, excluding the target feature itself. This model then predicts the value of the target feature based on the values of the other features in the modified sample.

During the adversarial attack process, when perturbations are applied to each dependent feature, the corresponding regression model is queried to predict its updated value, ensuring that the dependent feature remains consistent with the rest of the modified sample. This approach ensures that the perturbations respect consistency between features, thereby maintaining the overall coherence of the sample.

This method addresses scenarios where feature dependencies are known to exist, but the exact computational relationship is either unavailable to the attacker or only partially observable in the dataset. For instance, features such as \textit{`apache\_3j\_bodysystem'} and \textit{`apache\_3j\_diagnosis'} in the ICU dataset are components of clinical scoring systems (e.g., APACHE - Acute Physiology and Chronic Health Evaluation). These features are derived from multiple physiological and clinical attributes, some of which are not included in the dataset, creating semantic dependencies on other inputs through complex and often unknown relationships. Since the attacker cannot reproduce the precise calculations used in practice, the regression-based approach provides an effective approximation.

When the dependency structure is unknown or latent, an additional preliminary analysis stage is required to identify these relationships before applying the regression-based adjustment. Such identification can be performed using statistical dependency measures, feature clustering, or model-based approaches (e.g., probabilistic graphical models). Without this step, the regression-based perturbation technique cannot be reliably applied.

The framework is inherently modular and allows flexibility in the regression model selection based on dataset characteristics. For datasets with linear feature relationships, linear regression provides computational efficiency with satisfactory predictive accuracy. When complex, non-linear relationships exist, more sophisticated models such as polynomial regression or random forests can be employed to better capture feature interactions, although at increased computational cost. Any suitable dependency modeling technique can be substituted for the regression step, including non-linear or ensemble-based models, to better capture complex interactions and adapt to datasets with complex correlation structures.

This technique focuses on cases where features have complex or unknown dependencies on other variables, while assuming that the features with dependencies are not directly interdependent among themselves. More challenging scenarios arise when multiple features depend on each other simultaneously, forming mutual or circular dependencies. Handling such cases would require an additional methodology to identify the existence of these circular dependencies, determine the specific group of mutually dependent features, and develop solutions that enable coordinated modification of multiple interdependent features; this is an open challenge that remains for future research.

The application of this technique should be selective, based on the characteristics of the dataset. In our evaluation, we identified clear feature dependencies in the ICU dataset that warranted this approach. However, for datasets like the Hateful Users on Twitter and the Video Transmission Quality, our domain analysis revealed no features that are strictly dependent on other unknown or unobserved variables, making the regression-based adjustment unnecessary for these cases.

\sloppy
In our implementation of both query- and transferability-based attack strategies, the attack process includes an additional step in which the dependent features, as identified by domain knowledge, are adjusted according to the predictions from the regression models. This step helps maintain feature consistency, improve realism, and reduce the likelihood of generating detectable anomalies.

\subsection{Evaluating the Quality of Adversarial Attacks}
\fussy

The quality of an adversarial attack is measured by its ability to evade detection, which depends on how well the crafted samples maintain coherence and consistency with benign data.
To assess this, we propose two evaluation criteria:
\begin{enumerate}
\item Feature Space Coherency – Quantifying the distinguishability of adversarial samples from benign data distributions.
\item \sloppy Model Interpretation Stability – Analyzing the extent to which adversarial samples influence the internal decision-making process of the target model when queried.
\end{enumerate}

While \textit{Feature Space Coherency} captures statistical deviations from the benign distribution, \textit{Model Interpretation Stability} provides a complementary perspective by examining how adversarial samples influence the model’s internal reasoning. This aspect is particularly important in cases where adversarial perturbations are subtle and remain close to the benign data in the input space 
yet still cause shifts in the model's attribution of feature importance. Such changes may not be detectable through input-level analysis alone.

Both criteria leverage a key contribution of our work: the introduction of Class-Specific Anomaly Detection (CSAD). Unlike traditional one-class anomaly detection methods that assume all benign samples belong to a single class or distribution, CSAD evaluates anomalies relative to the specific class to which the adversarial sample is assigned by the target model. This class-aware evaluation enables CSAD to account for class-dependent variations in benign data, resulting in more effective anomaly detection.

\textbf{The CSAD Approach.}\label{csad}
In the CSAD approach, we evaluate anomalies \textbf{separately} for each target class. For instance, given a dataset with three classes, we train three distinct anomaly detection models—one for each class. Each model is trained exclusively on benign samples from its respective class, ensuring that the unique statistical properties of each class-specific distribution are captured.
When an adversarial sample is generated, it is evaluated using the anomaly detection model corresponding to its \textbf{predicted class}. For example, if an adversarial sample is classified as class 1 by the target model, it is assessed using the anomaly detection model trained on benign samples from class 1.

This approach addresses a key limitation of traditional anomaly detection techniques, where adversarial samples appear anomalous within the general data distribution, but not within their assigned class. Adversarial attacks introduce subtle perturbations that might preserve the statistical characteristics of the \textbf{original} sample while misleading the model. By design, adversarial samples can still resemble their true class and remain indistinguishable from other benign samples in that class. However, when analyzed in the context of the \textbf{target class} (the class to which they are misclassified), these perturbations may exhibit distinct anomalies. By leveraging class-specific distributions, CSAD enables more precise differentiation between adversarial and benign samples, capturing anomalies that would otherwise go undetected.

In addition to its conceptual advantages, the CSAD approach is computationally efficient in both the training and inference phases. During inference, CSAD maintains constant computational complexity, as each sample is evaluated by a single class-specific model, identical to traditional approaches.
The training phase reveals distinct computational advantages that depend both on the computational complexity of the underlying anomaly detection algorithm and on the class distribution in the dataset.

For algorithms with super-linear complexity, i.e., complexity of order $O(n^\alpha)$ with $\alpha>1$, such as one-class SVM with RBF kernels ($O(n^2)$ to $O(n^3)$)~\cite{scholkopf2001estimating}, CSAD can reduce computational overhead compared to traditional approaches that train a single model on the full dataset. In the balanced case, where each of the $k$ classes contains $\approx n/k$ samples, training $k$ models requires
$O(k \cdot (n/k)^\alpha) = O(k^{1-\alpha} \cdot n^\alpha)$ operations for $\alpha \geq 1$, 
which is lower than the $O(n^\alpha)$ cost of the traditional approach for $\alpha>1$. In unbalanced datasets, the computational cost improvement decreases as class imbalance grows, and in the extreme case where one class dominates (i.e., contains most of the samples), the computational cost becomes equivalent to that of the traditional method. A detailed derivation is provided in Appendix~\ref{appendix:appendixB}.

For algorithms with linear training complexity, such as IF ($O(n)$)~\cite{liu2008isolation} 
and AEs ($O(n)$)~\cite{sakurada2014anomaly}, partitioning the data into $k$ class-specific subsets 
does not change the total cost in either balanced or imbalanced scenarios, since the per class training cost 
scales linearly with the subset size. Training these $k$ models requires $O(n)$ operations in total, matching the computational complexity of the traditional single model approach. CSAD therefore maintains computational parity for linear models while retaining its conceptual advantages for class-specific anomaly detection. Nevertheless, its applicability is inherently constrained by the sample size requirements of the underlying anomaly detection method.

While this work focuses on tabular data, the core principle of CSAD—evaluating anomalies relative to the distribution of benign samples within the predicted class—is general and potentially applicable to other data modalities such as images, text, or graphs. However, extending CSAD to these domains requires the selection and adaptation of anomaly detection methods that align with the specific characteristics and constraints of the data. For example, in tabular data, subtle perturbations may remain close to the original distribution but become anomalous for the predicted class, while in other domains, the notions of anomaly and feature dependencies may differ. Additionally, the ensemble or consensus approach used in CSAD for tabular data—where separate anomaly detection models are trained per class—may not always provide similar benefits in other domains. In some cases, especially with less complex or more homogeneous datasets, a single global anomaly detection model might suffice, as in traditional approaches. Evaluating when and how class-specific modeling improves detection performance in different data modalities remains an open question that must be addressed before expanding CSAD application beyond tabular data.

\textbf{Feature Space Coherency.}\label{anomaly_metric} \sloppy
To evaluate the distinguishability of adversarial samples, we propose measuring the anomaly detection rate-the percentage of samples identified as anomalies for each attack. The anomaly detection rate metric serves as an indicator of how different adversarial samples are from the benign data distribution.
We compute the anomaly detection rate using two well-established methods from the anomaly detection domain:
\textit{i)} Isolation Forest (IF) – A tree-based algorithm that identifies anomalies based on how easily a sample can be isolated in a feature space, and
\textit{ii)} Autoencoder (AE) – A neural network-based model that reconstructs benign samples and detects anomalies based on the reconstruction error.

In the IF algorithm, a sample is considered anomalous if it is classified as an outlier. In the AE model, an anomaly is identified when its reconstruction error exceeds a threshold, defined in~\cref{equation:ae_threshold}:
\begin{equation}
\label{equation:ae_threshold}
\begin{split}
Threshold(x_c,\:AE_c) = mean(test_{c,\:r.\,error})\,+ \\ 2*std(test_{c,\:r.\,error})
\end{split}
\end{equation}
where $x_c$ is a sample classified as class $c$, $AE_c$ is the AE trained on benign samples of class $c$, and $test_{c,r.,error}$ represents the reconstruction error on a validation set of class $c$ during the AE's training. This thresholding approach follows a well-established practice in reconstruction-based anomaly detection using AEs \cite{nelson1984shewhart, shewhart1931economic,shan2022abnormal}. Such formulations aim to separate benign and anomalous samples based on the statistical distribution of reconstruction errors. Following this approach, we adopt $k=2$, based on empirical tuning to balance sensitivity and false positives across datasets and attack scenarios.

The AE model has a symmetric architecture with a single hidden layer of 64 neurons serving as the embedding layer between the encoder and decoder and ReLU activations. 
The model was trained using the Adam optimizer with a learning rate of $1e-3$, a weight decay of $1e-8$, for 10 epochs, and the mean square error (MSE) loss. This design aligns with common practices in reconstruction-based anomaly detection, where lightweight models have demonstrated good performance with reduced computational demands and lower risk of overfitting, thereby contributing to training stability~\cite{torabi2023practical}.
The IF method was implemented with 100 estimators and $max\_features$ set to one, without bootstrapping. For comparison, the contamination parameter in the IF model was set to match the false positive rate (FPR) obtained by the AE according to the threshold defined in \cref{equation:ae_threshold}, corresponding to the proportion of benign samples with a reconstruction error above this threshold.
Other anomaly detection methods can be considered depending on the dataset characteristics and attack patterns.

\textbf{Model Interpretation Stability.}\label{shap_metric}
 To quantify interpretation stability, we assess how adversarial samples influence the target model's decision-making process using SHAP~\cite{lundberg2017unified}, a widely used explainability technique that provides a unified measure of feature importance in ML models. SHAP estimates each feature’s contribution to a prediction, computed from the attacked model’s outputs.
Specifically, for each class $c$ and feature $f$, we define the normal range as $[\min(\text{SHAP}_{f,c}), \max(\text{SHAP}_{f,c})]$ based on the SHAP values observed in benign training samples assigned to class $c$. An adversarial sample predicted as class $c$ is considered to exhibit an anomalous SHAP value for feature $f$ if its value falls outside this class-specific range.

Based on this definition, we introduce two SHAP-based importance metrics:
(1) \textit{Importance-Based Anomaly Detection Rate} – the percentage of adversarial samples where at least one feature’s SHAP value falls outside the benign range (i.e., is considered anomalous); and
(2) \textit{Average Anomalous SHAP Features per Sample} – the average number of features per adversarial sample whose SHAP values fall outside the benign range.
Lower values for these metrics indicate higher-quality adversarial samples, as they remain consistent with the model’s internal reasoning despite altering predictions.

Importantly, we have extened CSAD to to SHAP-based anomaly detection; we compute SHAP-based anomalies within the context of each class, ensuring that SHAP value deviations are analyzed relative to the specific class distribution, rather than across the entire dataset.
The SHAP values were extracted using SHAP TreeExplainer.

\textbf{Leveraging Anomaly Detection for Defense.}
The Importance-Based Anomaly Detection Rate and Average Anomalous SHAP Features per Sample methodologies not only assess adversarial quality but also serve as defense mechanisms.
Since adversarial samples often exhibit statistical or interpretability inconsistencies, the anomaly detection rate—whether computed directly or through SHAP values—serves as a robust indicator for detecting adversarial attacks. Our class-specific approach further enhances the ability to distinguish between benign and adversarial samples, offering a stronger defense against adversarial manipulation.

\section{Experimental Setup} \label{sec:exp_setup}

\subsection{Datasets}\label{subsec-dataset}

We conducted our evaluation on two publicly available, real-life tabular datasets and an additional proprietary tabular dataset. These datasets contain a relatively large number of samples and features, as well as a variety of feature types, to ensure robust and reliable results.

\begin{enumerate}
\item \emph{The Hateful Users on Twitter (Hate) dataset}~\cite{ribeiro2018characterizing}. 
This dataset contains 4,971 records of English-speaking Twitter users that were manually annotated as hateful or non-hateful users. 
This dataset is unbalanced and includes 544 records from class 1 (i.e., hateful users). 
The dataset comprises 115 numerical and categorical features, with 109 of them being mutable (i.e., can be changed by the attacker). 

\item \emph{Intensive Care Unit (ICU) dataset\footnote{ICU:\url{kaggle.com/competitions/widsdatathon2020/data}}}~\cite{hanberger2005intensive}.
This dataset, which contains the medical records of 83,978 patients who were admitted to the intensive care unit, is used to predict the mortality of the patients. 
The dataset is unbalanced and includes 7,915 records from class 1 (i.e., patients who died while in the ICU). 
The dataset comprises 74 numerical and categorical features, 45 of which are mutable and five of which contain values that are dependent on other features. 

\item \emph{Video Transmission Quality (VideoTQ) dataset.}
This proprietary commercial dataset was obtained from RADCOM,\footnote{\url{radcom.com/}} a company specializing in service assurance for telecom operators, and it contains records of 54,825 streaming service media broadcasts. 
This dataset is used to predict a transmission's resolution (i.e., high or low resolution). 
The dataset is unbalanced and includes 10,000 records from class 1 (i.e., high resolution).
The dataset comprises 23 numerical and categorical features, nine of which are mutable.
\end{enumerate}

\subsection{Dataset Preprocessing}

The basic preprocessing pipeline was designed following the established methodology of Mathov et al.~\cite{mathov2022not}, which defines standard and widely accepted preprocessing practices for tabular data. This approach was deliberately chosen to ensure fair comparison between attack methods, given our focus on evaluating and comparing adversarial attack behaviors and the coherence of the adversarial samples they generate. By maintaining a consistent preprocessing pipeline across all datasets, we ensure that any variations in results are attributable to the attack behavior itself rather than inconsistencies in data preparation. This standardized approach also facilitates direct comparability with previous works that used similar preprocessing on the same datasets. Additional dataset-specific adaptations were applied where appropriate, to accommodate unique characteristics while preserving the overall consistency of the evaluation framework.

In accordance with the implementation of a standard preprocessing pipeline across all datasets, we systematically eliminated outliers, dropped a group of features with a correlation exceeding $90\%$ and randomly included just one of them, and excluded instances with more than $80\%$ missing data. In addition, unique features, as well as features for which over $75\%$ of the values were identical, were removed. Categorical features were label-encoded, following the methodology of Mathov et al.~\cite{mathov2022not} and Grolman et al.~\cite{grolman2022hateversarial}.
While one-hot encoding is the standard approach for nominal features, as it preserves equal semantic distance between categories, these works opted for label encoding due to constraints imposed by adversarial attack requirements. Specifically, applying perturbations to one-hot encoded features can lead to invalid states, where multiple elements of the same categorical feature may simultaneously flip their value from zero to one, violating the inherent constraint of mutual exclusivity and resulting in unrealistic adversarial samples. Label encoding maintains the single feature perturbation paradigm, ensuring that adversarial perturbations generate semantically valid examples~\cite{mathov2022not,grolman2022hateversarial}.

The following additional preprocessing steps were applied based on the unique characteristics of each dataset:

For the \textbf{Hate dataset,} we followed the methodological procedure of Grolman et al.~\cite{grolman2022hateversarial}, which included removing the \textit{`user\_id'} feature and excluding features that contain GloVe representations~\cite{pennington2014glove} of other features (which include \textit{`glove'} in their names), resulting in a distilled set of 311 features.
Following the same methodology, we trained four models on the dataset: XGBoost~\cite{chen2016xgboost}, gradient boosting~\cite{friedman2001greedy}, LightGBM~\cite{ke2017lightgbm}, and random forest~\cite{breiman2001random}, and selected the top-40 most important features for each model based on the Pearson correlation coefficient, ensuring that only meaningful and relevant features were retained. All of the selected features had importance scores exceeding 0.0001, whereas the remaining features exhibited near-zero importance scores.
The final processed dataset consists of 115 features and 4,971 samples.

For the \textbf{ICU dataset,} we followed the preprocessing methodology outlined in a publicly available Kaggle implementation.\footnote{\url{kaggle.com/code/binaicrai/fork-of-fork-of-wids-lgbm-gs}} This included both the basic preprocessing steps described above and additional steps proposed in the Kaggle solution. Specifically, the training and test sets provided by Kaggle were combined to increase the amount of available data, as the original training set contained a substantial proportion of missing values, while the test set was comparatively more complete.
Missing values were handled through a multi-step imputation process. Samples for which the \textit{`bmi'} feature was missing were filled in using the BMI formula based on the values in the \textit{`height'} and \textit{`weight'} features. Missing values in the following features were imputed using regression models to predict values for each feature: \textit{`age,' `height,' `weight,' `apache\_4a\_hospital\_death\_prob,'} and \textit{`apache\_4a\_icu\_death\_prob.'} Following the imputation process described in Mathov et al. (2022), the remaining numeric features were imputed using median values, and categorical features were imputed using modal values to avoid filling in invalid or non-integer values and ensure consistency with the original feature distribution. 

We acknowledge the limitations of using the Kaggle-based preprocessing, specifically the potential introduction of bias due to merging the training and test sets. This approach may alter the data distribution and affect class balance; however, given that the original training set was already highly unbalanced and contained substantial missing data, we considered this trade-off acceptable. Furthermore, the selected preprocessing pipeline demonstrated strong classification performance in the Kaggle setting on the same task as in this paper, supporting its suitability for our objective.

For the \textbf{VideoTQ dataset,} we removed 15 records that had missing values and converted all timestamp features into separate date and time formats.

For all datasets, the editable and immutable features were determined and selected by a domain expert. 
All datasets were split into training and test sets, with $75\%$ of the samples serving as the training set.
Oversampling was performed by duplicating the training set samples in class 1 to address the significant class imbalance in the datasets.
The oversampled training set was split into a target training set and a surrogate training set (for the attacker's use in the transferability-based attacks).
The test set is also used to create adversarial samples (i.e., the attack set) as illustrated in~\cref{fig:flow} and further detailed in \cref{subsec-eval_setup}.

\subsection{Classification Models Used for Evaluation}

For our evaluation, we trained four different ML classification models commonly used in real-world tabular data: XGBoost~\cite{chen2016xgboost} (XGB), gradient boosting~\cite{friedman2001greedy} (GB), LightGBM~\cite{ke2017lightgbm} (LGB), and random forest~\cite{breiman2001random} (RF) (see~\cref{fig:flow}).
The hyperparameters for each model were manually tuned to fulfill two key objectives: (i) to achieve high classification performance on each dataset, and (ii) to promote model robustness by encouraging the model to rely on a diverse set of features instead of overfitting to a small number of highly correlated features.
The XGBoost target models were trained using 90, 70, and 300 estimators; a max depth of 3, 8, and 5; and a learning rate of 1, 0.1, and 1 for the Hate, ICU, and VideoTQ datasets respectively. 
The gradient boosting target models were trained using 40, 500, and 300 estimators; a max depth of 7, 6, and 5; and a learning rate of 2.5, 0.01, and 1 for the Hate, ICU, and VideoTQ datasets respectively. 
The LightGBM target models were trained using 300, 200, and 200 estimators; a max depth of 7, 8, and 8; and a learning rate of 1, 0.1, and 0.1 for the Hate, ICU, and VideoTQ datasets respectively. 
The random forest target models were trained using 100, 500, and 500 estimators and a max depth of 4, 9, and 9 for the Hate, ICU, and VideoTQ datasets respectively.

\subsection{Attack Configuration}\label{subsec:attack_setup}

In our comparison of different types of attacks, we used two query-based attacks and five transferability-based gradient attacks that varied in terms of their feature selection techniques.
The boundary attack was performed with an epsilon of one, a delta of one, maximal iterations of 3000, the number of trials set at 20, and an adaptation step of one.
The HopSkipJump attack was performed with the L2 norm, maximal iterations of 50, the max-eval parameter set at 10,000, the init-eval set at 500, and an init size of 100.
The parameters of the boundary and HopSkipJump attacks were identical for all datasets, except for the maximal iteration number; for the VideoTQ dataset this was set at 1000 in the HopSkipJump attack.

The transferability-based attacks require the attacker to train a surrogate model in order to generate adversarial samples that are transferred to the target black-box model (illustrated in~\cref{fig:flow}).
In our evaluation, all surrogate models were neural networks in which an embedding layer was used to try to maintain the consistency of the crafted samples, following the established approaches~\cite{mathov2022not,grolman2022hateversarial}. The surrogate models were comprised of two components: a sub-model for embedding and a sub-model for classification, both of which were fine-tuned for optimized performance.

The embedding sub-model had an input size matching the number of features in the input sample, followed by a dense layer with 256 neurons and ReLU and PReLU activations for the Hate and VideoTQ datasets respectively. For the ICU dataset, the embedding sub-model had three dense layers with 256, 128, and 64 neurons and ReLU activation for all layers. The output size of all embedding sub-models was set at 16, i.e., the embedding size.
The classification sub-model had a dense layer with 16 neurons (matching the embedding size), with ReLU activation for the Hate and ICU datasets and PReLU activation for the VideoTQ datasets, followed by a dropout layer with a dropout rate of 0.1. All surrogate models were trained using the binary cross-entropy loss function~\cite{liu2017learning} and the Adam optimizer~\cite{kingma2014adam} with a learning rate of 0.2 for the Hate and VideoTQ datasets and 0.1 for the ICU dataset.
Training was conducted for a maximum of 70 epochs, with early stopping applied based on the validation loss (patience = 2 epochs). The model parameters from the epoch yielding the lowest validation loss were retained.

After training, the attacker performs the adversarial attack against the surrogate model using specific feature selection techniques. We implemented four attacks that select features for perturbation based on their importance scores, differing only in the model used to compute the feature importance (via SHAP values): XGBoost, gradient boosting, LightGBM, or random forest, each trained separately for this purpose. Additionally, we implemented a random-based selection attack that selects features uniformly at random. In all attacks, $k=2$ features are selected per iteration along with $n=1$ correlated features (identified using the Pearson correlation coefficient~\cite{weisstein2006correlation}), continuing until an adversarial sample is found or the maximum allowed $L_0$ perturbation $\lambda$ is reached. In our implementation $\lambda$ is set to be the number of editable features.

The adversarial objective underlying these attacks is to generate adversarial samples that mislead the surrogate model while preserving similarity to the original input in the latent space. This is achieved 
by minimizing the adversarial loss function, defined as:
\begin{equation}
\label{equation:objective}
\begin{split}
L_{adv}(x_{adv}, x, y) = -\mathrm{BCE}(M'(x_{adv}), y) \\ + \alpha \left\| \phi(x_{adv}) - \phi(x) \right\|_{2}
\end{split}
\end{equation}

where $y$ is the true label of $x$, $\mathrm{BCE}$ denotes the binary cross-entropy loss, $M'$ is the surrogate model, $\phi(\cdot)$ is the embedding sub-model that outputs the latent representation (embedding), and $\alpha=1$ is the regularization parameter. The first term promotes misclassification by maximizing the classification error of $x_{adv}$, whereas the second term ($L_2$ regularization) seeks to preserve similarity by minimizing the Euclidean distance between the latent representations of $x_{adv}$ and $x$ which correspond to the output of the embedding sub-model $\phi(\cdot)$.

The adversarial sample $x_{adv}$ is obtained by solving the optimization problem
\begin{equation}
\label{equation:optimization}
\begin{split}
x_{adv} = \arg\min_{x_{adv}} L_{adv}(x_{adv}, x, y)
\end{split}
\end{equation}

Optimization is performed using the Adam optimizer with manually tuned learning rates of 1.0 for the Hate and VideoTQ datasets and 20.0 for the ICU dataset.

The detailed procedure for feature selection, perturbation computation, and constraint handling is provided in Appendix~\ref{appendix:appendixA}. The full attack design and theoretical framework are described in detail in~\cite{mathov2022not,grolman2022hateversarial}.

\subsection{Regression Models}\label{regresion_models}

\sloppy
Of the datasets used in our evaluation, only the ICU dataset contains dependent features whose relationships with other attributes are known to exist but are not explicitly defined or fully observable in the data. Therefore, we applied the regression-based perturbation technique introduced in \cref{proposed_approach} exclusively to this dataset.
Specifically, we identified four dependent features that require regression-based adjustment: \textit{`apache\_3j\_bodysystem,'} \textit{`apache\_3j\_diagnosis,'} \textit{`d1\_mbp\_invasive\_max,'} and \textit{`d1\_mbp\_invasive\_min.'} For each of these features, we trained a separate regression model using the same dataset employed for training the surrogate models.

The regression models were implemented using gradient boosting with 200 estimators and a maximum depth of six. However, different learning rates were required for optimal performance: 0.1 for the \textit{`apache\_3j\_bodysystem'} and \textit{`apache\_3j\_diagnosis'} features, and 0.01 for the \textit{`d1\_mbp\_invasive\_max'} and \textit{`d1\_mbp\_invasive\_min'} features.
These hyperparameters were determined through manual tuning to optimize model performance for each feature, with the different learning rates reflecting the varying complexity and scale of the target features. Since satisfactory performance was achieved across all models, no further hyperparameter optimization was conducted.

\subsection{Threat Model}
\fussy
We assume that the attacker can query the target model and has an unlimited query budget for query-based attacks and a query budget of one for transferability-based attacks; in each query to the target model, the attacker obtains the confidence score for the prediction.
In addition, we assume that the attacker has no prior knowledge about the specific model architecture or ML algorithm used. 
The attacker also has no access to the internal parameters of the model (such as weights and biases) and cannot calculate any gradients related to the model.
Furthermore, the attacker does not have access to the target model's training data; however, we assume that the attacker has a surrogate dataset derived from the same distribution, meaning that it contains the same features, in the same order, as the training data.

\subsection{Evaluation Setup}\label{subsec-eval_setup}
\sloppy

To ensure a robust and meaningful evaluation of adversarial effectiveness, we constructed the attack set by filtering the test set to include only samples that were correctly classified by both the target and surrogate models prior to the attack. This guarantees that any misclassification following the perturbation can be directly attributed to the adversarial manipulation rather than to pre-existing model errors. After filtering, $85\%$, $69\%$, and $91\%$ of the test samples were retained for the Hate, ICU, and VideoTQ datasets, respectively. We then randomly selected a balanced number of samples from each class, resulting in attack sets of 182, 1,000, and 1,000 samples, respectively for the three datasets.

For the purpose of attack evaluation, we assess only adversarial samples that successfully fooled the target model (or both the surrogate and target models in the case of transferability-based attacks). This allows us to analyze the properties of adversarial samples that actually compromise model reliability, as failed attacks do not pose a real threat to downstream users.

When analyzing intrinsic properties of adversarial samples, such as $L_p$ distances or anomaly scores, we compare them to the distribution of benign (non-adversarial) training samples. Importantly, this benign reference set includes all training samples, regardless of whether they were correctly classified. This methodological choice captures the full variability of natural data, reflecting a realistic deployment scenario in which even misclassified samples influence the model's perceived class distribution. As a result, anomaly scores are computed relative to a distribution that detectors might actually encounter in practice.

\subsection{Evaluation Metrics}\label{subsec-metrics}

The main objective of our evaluation is to answer the research questions presented in \cref{intro} and analyze the characteristics of and differences between black-box query- and transferability-based adversarial attacks with respect to three key factors: the attacker's risk (RQ\#1), the attacker's effort (RQ\#1), and the attack quality (RQ\#2, RQ\#3).
To perform a thorough analysis, we used the evaluation metrics commonly used in the literature and the metrics proposed in this work.

To assess the \textbf{attacker's risk} we used three metrics (illustrated in~\cref{fig:flow}):
\textit{i)} the percentage of samples that successfully misled the model (attack \textit{success rate});
\textit{ii)} the perturbation magnitude computed based on the average number of modified features ($L_0$), which reflects practical constraints such as the cost and feasibility of changing specific features in real-world tabular data, and the average Euclidean distance between the original and adversarial samples ($L_2$), included to support comparison with prior work and to demonstrate its limitations in capturing detectability in heterogeneous tabular datasets;
and \textit{iii)} the average number of queries required.

We exclude the $L_\infty$ norm since, like $L_2$, it depends heavily on feature scaling and poorly reflects perturbation detectability or practical impact in tabular domains with mixed feature types and scales~\cite{mathov2022not}. Furthermore, $L_\infty$ is not used in prior work on adversarial attacks in tabular domains, reflecting its lack of applicability and relevance in this context.

To assess the \textbf{attacker's effort} we used three metrics (illustrated in~\cref{fig:flow}):
\textit{i)} the time it took to craft an adversarial sample; 
\textit{ii)} the number of data samples needed to execute the attack;
and \textit{iii)} the memory resources and computing capabilities required to craft the adversarial sample and perform the optimization process.

The \textbf{attack quality} is reflected in the crafted adversarial sample, which were assessed using three metrics (illustrated in~\cref{fig:flow}):
\textit{i)} the \textit{anomaly detection rate} of adversarial samples compared to benign samples, using IF and AE models, where a higher rate indicates less consistency and coherence (RQ\#2);
\textit{ii)} the percentage of samples with at least one feature with an anomalous SHAP value (\textit{Importance-Based Anomaly Detection Rate}) (RQ\#3); and \textit{iii)} the average anomalous SHAP features per sample (RQ\#3).

All three quality metrics are computed using our CSAD approach (see \cref{anomaly_metric}), which evaluates adversarial deviations relative to the distribution of benign samples from the same predicted class. In this context, lower values across these metrics indicate higher-quality adversarial samples, since they reflect less abnormal behavior within the predicted class. While feature space anomalies (metric i) are model-agnostic and computed once per attack, SHAP-based anomalies (metrics ii and iii) depend on the model’s internal reasoning and are therefore assessed separately for each target model.

\subsection{Experimental Environment Setup}

All experiments were performed on an Intel Core i7-10700 CPU at 2.90 GHz processor with the Windows 10 Pro operating system, an Intel UHD Graphics 630 graphics card, and 32 GB of memory. 
The code used in the experiments was written using Python 3.10.9 and the following Python packages:
PyTorch 2.0.0, adversarial-robustness-toolbox 1.13.0~\cite{nicolae2018adversarial}, pandas 1.5.2, NumPy 1.23.5, TensorFlow 2.10.0, and Keras 2.10.0.

\section{Experimental Results}\label{sec:Results}

\subsection{Models' Performance}

\cref{tab:models_performence} summarizes the performance of all target and surrogate models examined, presenting the accuracy, F1 score, precision, and recall values obtained by each model on the test set.

\begin{table}
\caption{Target and surrogate models' performance on the test set.}
\centering
\resizebox{1\columnwidth}{!}{
\begin{tabular}{@{}llccccc@{}}
\toprule
\multirow{2}{0.07\textwidth}{\textbf{Dataset}} &
\multirow{2}{0.07\textwidth}{\textbf{Metric}} &
\multicolumn{5}{c}{\textbf{Models}}\\
\cmidrule(l){3-7}
& &\textbf{GB} & 
\textbf{LGB} & 
\textbf{XGB} & 
\textbf{RF} & 
\textbf{Surrogate} \\ \midrule
\textbf{Hate}   & Accuracy        & 0.91        & 0.92         & 0.88         & 0.91        & 0.95 \\
                 & F1 Score        & 0.91        & 0.92         & 0.87         & 0.91        & 0.95 \\
                 & Precision       & 0.93        & 0.94         & 0.94         & 0.89        & 0.92 \\
                 & Recall          & 0.89        & 0.91         & 0.82         & 0.92        & 1.00  \\ \midrule
\textbf{ICU}    & Accuracy        & 0.80        & 0.80          & 0.77         & 0.78        & 0.78  \\
                 & F1 Score        & 0.79        & 0.81         & 0.75         & 0.77        & 0.79 \\
                 & Precision       & 0.82        & 0.84         & 0.82         & 0.81        & 0.81 \\
                & Recall          & 0.77        & 0.74         & 0.69         & 0.74        & 0.74 \\ \midrule
\textbf{VideoTQ} & Accuracy        & 0.98        & 0.99         & 0.98         & 0.98        & 0.96 \\
               & F1 Score        & 0.99        & 1.00          & 0.99         & 0.99        & 0.97 \\
               & Precision       & 0.91        & 0.97         & 0.96         & 0.93        & 0.86 \\
               & Recall          & 0.98        & 0.99         & 0.96         & 0.98        & 0.93 \\ \bottomrule 
\end{tabular}}
\label{tab:models_performence}
\end{table}

As can be seen, all models were high-performing.
The target models for the Hate dataset obtained accuracy values ranging from [$88\%-91\%$], F1 scores ranging from [$87\%-92\%$], precision values ranging from [$89\%-94\%$], and recall values ranging from [$82\%-92\%$].
The target models for the ICU dataset obtained accuracy values ranging from [$77\%-80\%$], F1 scores ranging from [$75\%-81\%$], precision values ranging from [$81\%-84\%$], and recall values ranging from [$69\%$-$77\%$].
The target models for the VideoTQ dataset obtained accuracy values ranging from [$98\%-99\%$], F1 scores ranging from [$99\%-100\%$], precision values ranging from [$91\%-97\%$], and recall values ranging from [$96\%$-$99\%$].

\subsection{Attacker's Risk}\label{subsec:results_risk}
\begin{figure*}[!htbp]
    \centering
    
    \begin{subfigure}[b]{0.42\textwidth}
        \centering
        \includegraphics[width=\textwidth]{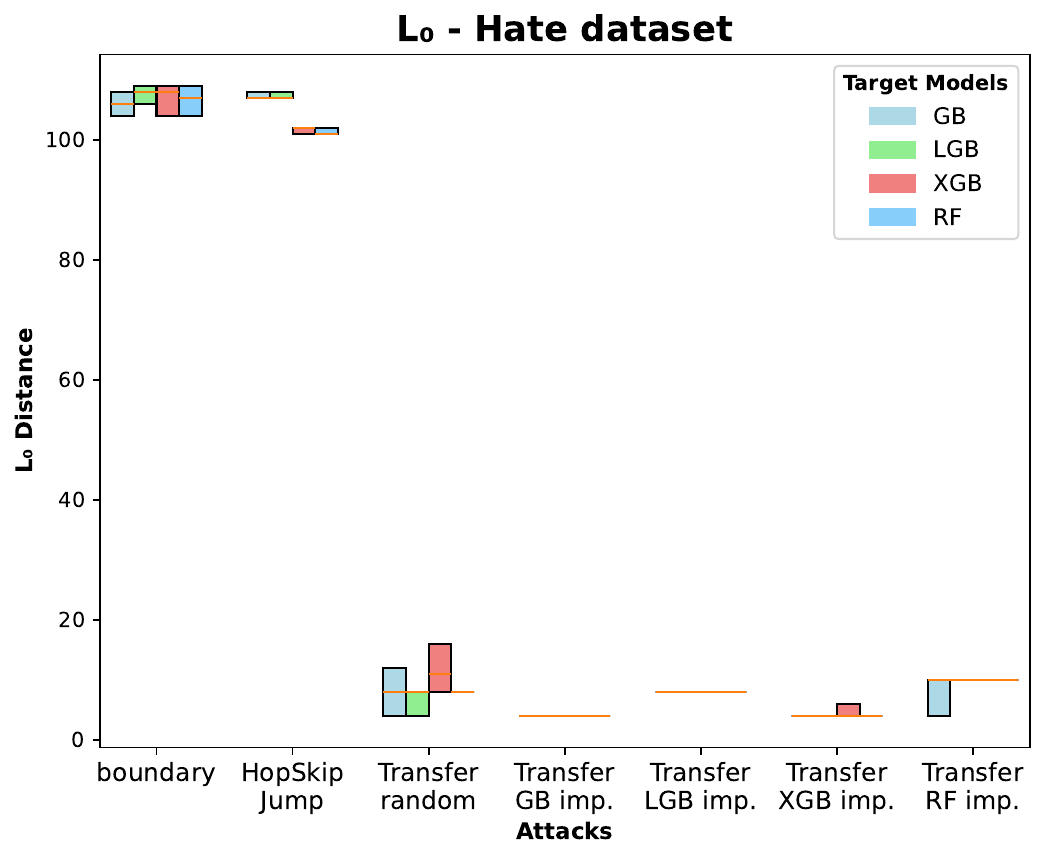}
        \label{fig:l0_hate}
    \end{subfigure}
    \begin{subfigure}[b]{0.42\textwidth}
        \centering
        \includegraphics[width=\textwidth]{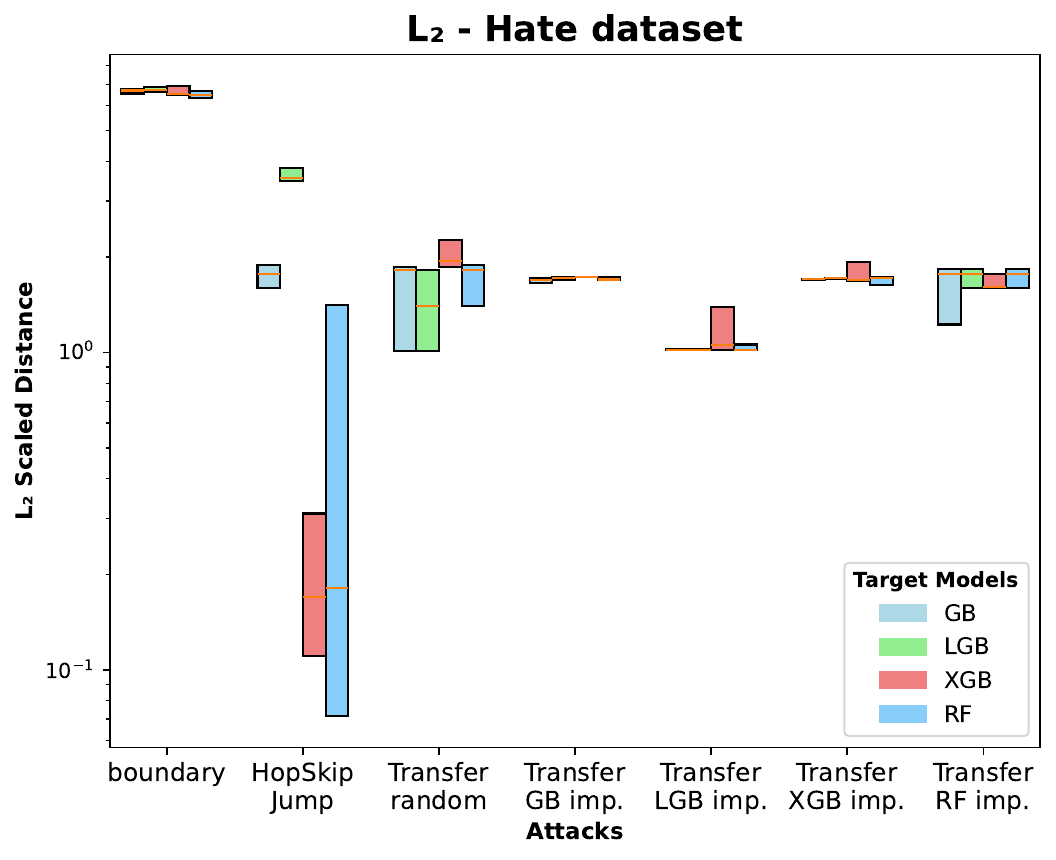}
        \label{fig:l2_hate}
    \end{subfigure}
    \begin{subfigure}[b]{0.42\textwidth}
        \centering
        \includegraphics[width=\textwidth]{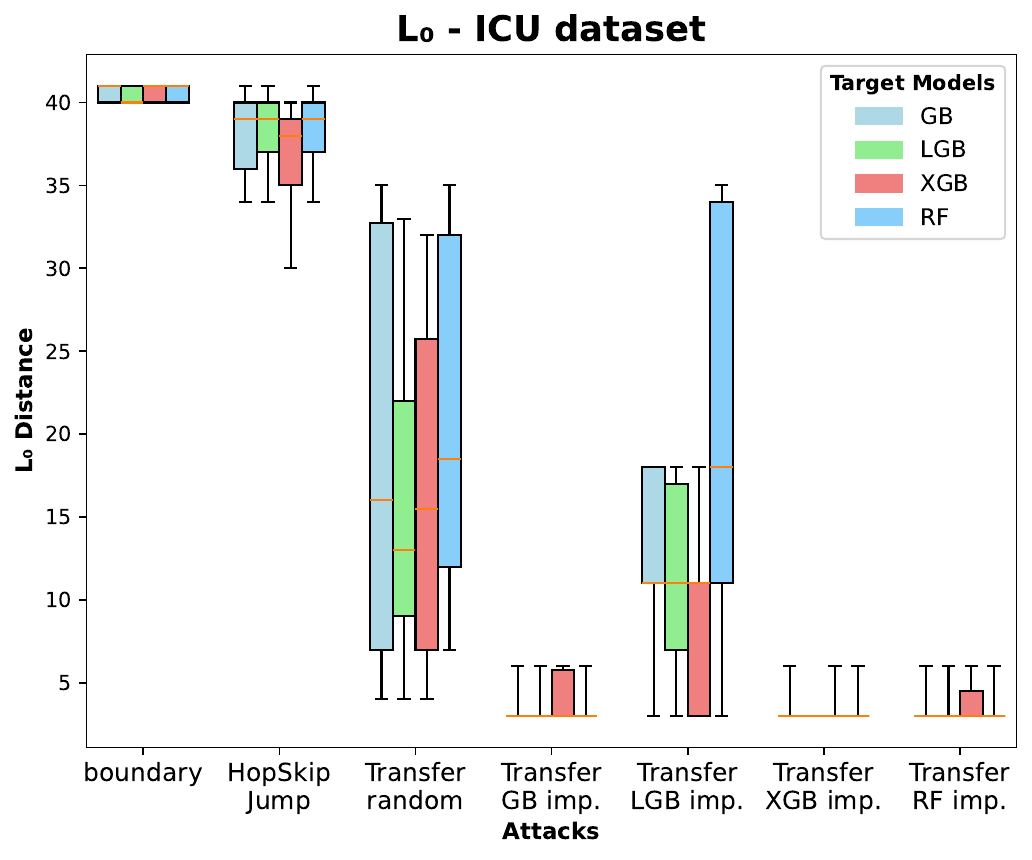}
        \label{fig:l0_icu}
    \end{subfigure}
    \begin{subfigure}[b]{0.42\textwidth}
        \centering
        \includegraphics[width=\textwidth]{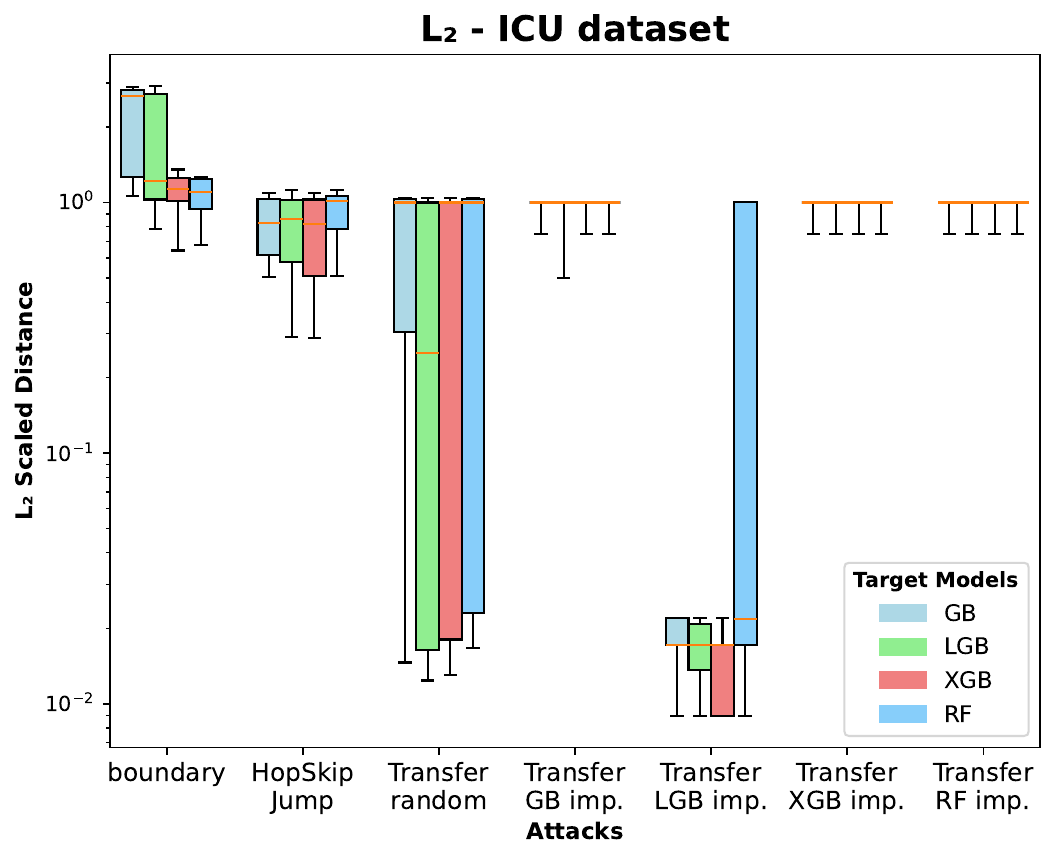}
        \label{fig:l2_icu}
    \end{subfigure}
    \begin{subfigure}[b]{0.42\textwidth}
        \centering
        \includegraphics[width=\textwidth]{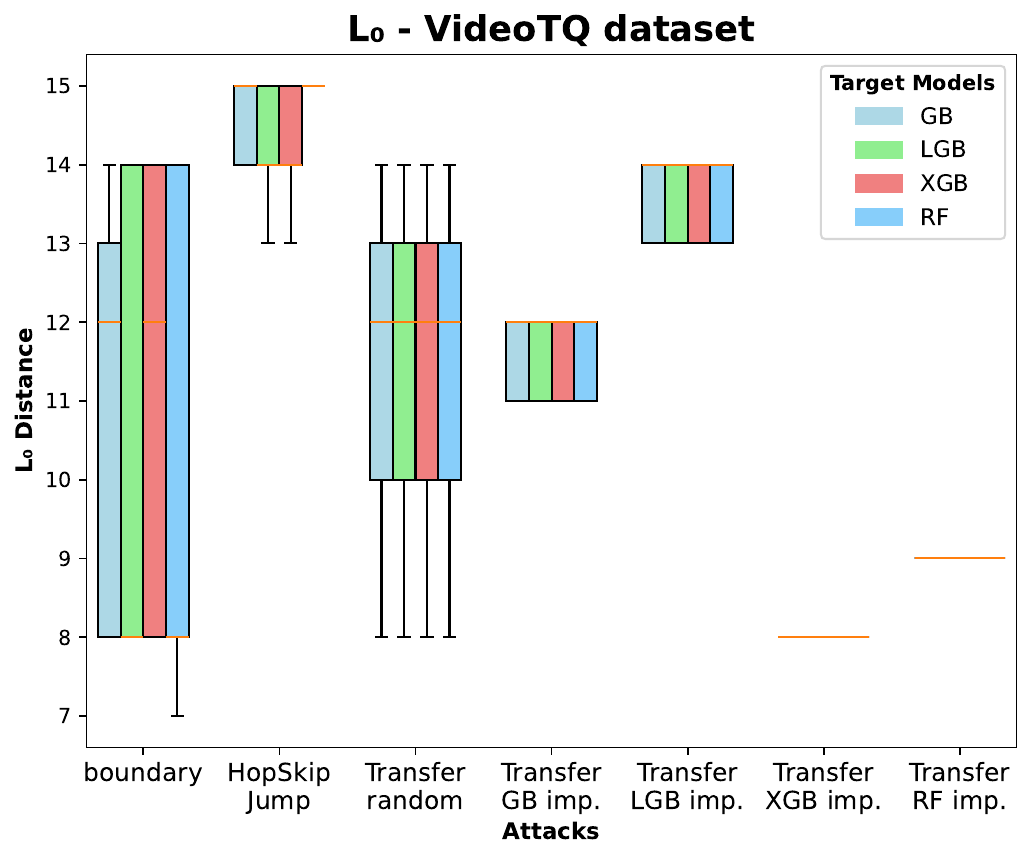}
        \label{fig:l0_VideoTQ}
    \end{subfigure}
    \begin{subfigure}[b]{0.42\textwidth}
        \centering
        \includegraphics[width=\textwidth]{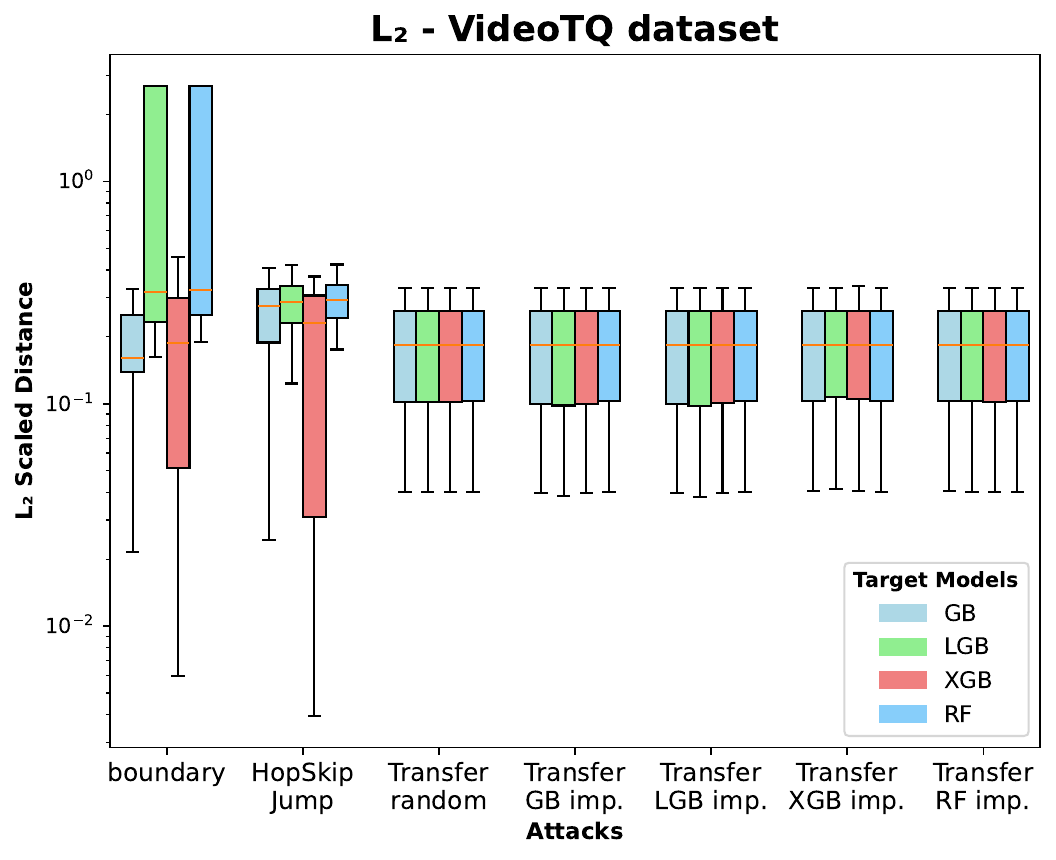}
        \label{fig:l2_VideoTQ}
    \end{subfigure}
    
    \caption{\textbf{Attacker's risk:} number of changed features (\textit{$L_0$ distance}) and distortion size (\textit{$L_2$ distance}), across query- and transferability-based attacks.}
    \label{fig:lp_norm}
\end{figure*}

\sloppy
\cref{tab:compare_sr,tab:compare_queries} and~\cref{fig:lp_norm} present the results for the attacker’s risk metrics (see~\cref{subsec-metrics}) for the query- and transferability-based attacks performed against different target models on each dataset.

\begin{table*}
\fussy
\caption{\textbf{Attacker's risk}: performance of query- and transferability-based attacks on different target models, evaluated using the following metrics: Target Success Rate (\textit{SR}): percentage of adversarial samples that successfully mislead the target model;
Surrogate Success Rate (\textit{Surrogate SR}): percentage of adversarial samples that successfully mislead the surrogate model;
Transfer Success Rate (\textit{Transfer SR}): percentage of successful adversarial samples that transfer from the surrogate model to the target model (computed as the ratio of the Surrogate SR);
Overall Success Rate (\textit{Overall SR}): percentage of adversarial samples that mislead both the surrogate and target models (computed based on the entire attack set).
Note that importance-based feature selection techniques are referred to as \textit{Imp.} in the table.}
\centering
\resizebox{\textwidth}{!}{
    \begin{tabular}
    {@{}p{0.07\textwidth}p{0.055\textwidth}p{0.06\textwidth}p{0.09\textwidth}ccccc@{}}
    \toprule
    \multirow{3}{0.07\textwidth}{\textbf{Dataset}} &
    \multirow{3}{0.06\textwidth}{\textbf{Target Model}} &
    \multicolumn{2}{c}{\textbf{Query Attacks}} &
    \multicolumn{5}{c}{\textbf{Transferability Attacks}} \\ 
    \cmidrule(l){3-4} \cmidrule(l){5-9} 
      &&
      \textbf{boundary} &
      \textbf{HopSkipJump} &
      \textbf{random} &
      \textbf{GB imp.} &
      \textbf{LGB imp.} &
      \textbf{XGB imp.} &
      \textbf{RF imp.} \\ 
      ~ & ~ & \multicolumn{2}{c}{SR (\%)} & \multicolumn{5}{c}{Surrogate SR (\%) / Transfer SR (\%) / Overall SR (\%)} \\ \toprule
         \textbf{Hate} & GB & \centering100  & \centering100  & 98.9 / 13.9 / 13.7 & 98.9 / 70.6 / 69.8 & 98.9 / 44.4 / 44.0 & 98.9 / 83.3 / 82.4 & 98.9 / 86.7 / 85.7 \\
        ~ & LGB & \centering100 & \centering100  & 98.9 / 7.8 / 7.7   & 98.9 / 71.1 / 70.3 & 98.9 / 45.6 / 45.1 & 98.9 / 73.9 / 73.1 & 98.9 / 85.0 / 84.1 \\
        ~ & XGB & \centering100  & \centering100  & 98.9 / 12.2 / 12.1 & 98.9 / 70.0 / 69.2 & 98.9 / 36.7 / 36.3 & 98.9 / 68.3 / 67.6 & 98.9 / 71.7 / 70.9 \\
        ~ & RF & \centering100  & \centering100  & 98.9 / 2.8 / 2.7   & 98.9 / 36.7 / 36.3 & 98.9 / 20.6 / 20.3 & 98.9 / 54.4 / 53.8 & 98.9 / 73.3 / 72.5 \\ \midrule 
        \textbf{ICU} & GB & \centering98.6 & \centering98   & 87.4 / 9.4 / 8.2   & 87.8 / 45.4 / 39.9 & 87.4 / 25.4 / 22.2 & 88.6 / 41.9 / 37.1 & 88.4 / 45.8 / 40.5 \\
        ~ & LGB & \centering98.7 & \centering97.9 & 87.4 / 12.8 / 11.2 & 87.8 / 35.8 / 31.4 & 87.4 / 29.1 / 25.4 & 88.6 / 34.2 / 30.3 & 88.4 / 36.4 / 32.2 \\
        ~ & XGB & \centering99.9 & \centering98.2 & 87.4 / 24.5 / 21.4 & 87.8 / 55.2 / 48.5 & 87.4 / 41.3 / 36.1 & 88.6 / 50.5 / 44.7 & 88.4 / 55.3 / 48.9 \\
        ~ & LGB & \centering99.2 & \centering99.2 & 87.4 / 11.4 / 10.0 & 87.8 / 44.9 / 39.4 & 87.4 / 18.4 / 16.1 & 88.6 / 42.8 / 37.9 & 88.4 / 45.4 / 40.1 \\ \midrule 
        \textbf{VideoTQ} & GB & \centering82.6 & \centering57.6 & 67.6 / 73.7 / 49.8 & 67.6 / 74.4 / 50.3 & 67.6 / 74.3 / 50.2 & 67.6 / 73.5 / 49.7 & 67.6 / 73.7 / 49.8 \\
        ~ & LGB & \centering73.7 & \centering53   & 67.6 / 74.1 / 50.1 & 67.6 / 74.7 / 50.5 & 67.6 / 74.9 / 50.6 & 67.6 / 73.4 / 49.6 & 67.6 / 74.0 / 50.0 \\
        ~ & XGB & \centering81.7 & \centering67.9 & 67.6 / 74.1 / 50.1 & 67.6 / 74.3 / 50.2 & 67.6 / 74.1 / 50.1 & 67.6 / 73.2 / 49.5 & 67.6 / 73.8 / 49.9 \\
        ~ & RF & \centering65.9 & \centering50   & 67.6 / 74.0 / 50.0 & 67.6 / 74.0 / 50.0 & 67.6 / 74.0 / 50.0 & 67.6 / 74.0 / 50.0 & 67.6 / 74.0 / 50.0 \\ \bottomrule
        \end{tabular} }

\label{tab:compare_sr}
\end{table*}

\begin{table}
\caption{\textbf{Attacker's risk:} average number of \textit{queries} required to execute query-based attacks.}
\centering
\resizebox{\columnwidth}{!}{
\begin{tabular}{@{}llcccc@{}}
\toprule
\multirow{2}{*}{\textbf{Dataset}} & 
\multirow{2}{*}{\textbf{Attack}} & 
\multicolumn{4}{c}{\textbf{Target Model}} \\  \cmidrule(l){3-6}
 &   &  \textbf{GB} & \textbf{LGB}  & \textbf{XGB}  & \textbf{RF} \\ \midrule
Hate  & Boundary  & 292.5    & 302.0    & 275.4    & 309.8    \\
& HopSkipJump     & 120260.0 & 120246.6 & 120243.4 & 120361.2 \\ \midrule
ICU   & Boundary  & 902.3    & 2909.9   & 910.9    & 3531.7   \\ 
 & HopSkipJump    & 119231.3 & 119402.4 & 119642.7 & 121632.6 \\ \midrule
VideoTQ  & Boundary & 59604.1  & 57627.6  & 58860.6  & 60173.1  \\
        & HopSkipJump & 16310.9  & 118374.4 & 103314.8 & 108680.1 \\ \bottomrule
\end{tabular}}

\label{tab:compare_queries}
\end{table}

\cref{tab:compare_sr} presents details on the following attack success rates:
Target Success Rate (SR): The percentage of adversarial samples that successfully mislead the target model;
Surrogate Success Rate (Surrogate SR): The percentage of adversarial samples that successfully mislead the surrogate model;
Transfer Success Rate (Transfer SR): The percentage of adversarial samples that successfully transfer from the surrogate model to the target model, calculated as the ratio of the Surrogate SR. Overall Success Rate (Overall SR): The percentage of adversarial samples that successfully mislead both the surrogate and target models, computed as the ratio of the entire attack set.

As detailed in \cref{tab:compare_sr}, the query-based attacks have consistently high success rates (SR) on all target models across the Hate and ICU datasets. For example, on the Hate and ICU datasets, both attacks achieve success rates ranging between [$98\%-100\%$] across all target models.
In comparison, the success rate varies more for the transferability-based attacks. The success of these attacks relies on the performance of the surrogate models and the transferability of adversarial samples. For instance, on the Hate dataset, despite successfully fooling surrogate models in $98.9\%$ of cases, the transferability-based attacks demonstrated limited effect on the target models, with success rates falling below $12\%$ when employing random feature selection (Transfer random). However, when using importance-based feature selection (Transfer imp.), the success rates improved, ranging between [$20\%-86\%$].
A similar phenomenon can be observed on the ICU dataset, where transferability-based attacks continued to perform variably, depending on the feature selection method employed.
On the VideoTQ dataset, unlike the other datasets, transferability-based attacks had more consistent performance, achieving a $\sim$$50\%$ success rate overall. In contrast, the query-based attacks displayed more variability, with success rates ranging between [$65.9\%-82.6\%$] for the boundary attack and between [$50\%-67.9\%$] for the HopSkipJump attack.

According to pairwise proportions z-tests with Holm correction, 
the differences in success rates between query- and transferability-based attacks 
were statistically significant across all datasets ($p<0.05$). 
Non-significant results were observed only within the transferability group, 
specifically between variants that achieved similarly moderate success 
(e.g., GB imp. vs LGB imp. importance-based selection attacks on Hate and ICU, 
and the different transfer variants on VideoTQ). 
Full pairwise results are provided in Appendix~\ref{appendix:appendixC}.

\cref{tab:compare_queries} presents the average number of queries required to generate adversarial samples for query-based attacks. As can be seen, the examined query-based attacks require a significant number of queries, ranging from hundreds to hundreds of thousands. In contrast, transferability-based attacks require just a single query for all datasets.

\cref{fig:lp_norm} illustrates the distribution of both the number of modified features ($L_0$ ) and the Euclidean distance ($L_2$) between the adversarial attacks and the original samples. 

As can be seen in the figure, for most target models on the Hate and ICU datasets the examined query-based attacks consistently require more feature changes (i.e., larger $L_0$ distances) than the transferability-based attacks. For instance, on the Hate dataset transferability-based attacks modify approximately $7$ features, whereas query-based attacks modify around $103$.

However, the VideoTQ dataset exhibits a less consistent pattern, with the boundary attack shows substantial variation in $L_0$ values, including samples with considerably lower $L_0$ distances than those of transferability-based attacks. This suggests that, in some cases, attacks required more feature modifications to succeed on this dataset.

In terms of $L_2$ distance, the two query-based attacks differ. Across the Hate and ICU datasets, the boundary attack generally causes larger distortions compared to transferability-based attacks. However, the VideoTQ dataset exhibits a different pattern: the boundary attack achieves notably lower $L_2$ distances on the GB and XGB target models than the transferability-based attacks. In contrast, while the HopSkipJump attack typically outperforms most transferability-based attacks by achieving lower median $L_2$ distances on most target models, it performs less effectively on the VideoTQ dataset, showing higher median $L_2$ distances on several target models.

Statistical validation using pairwise Mann–Whitney U tests with Holm correction confirmed these patterns across datasets. 
For both $L_0$ and $L_2$, query-based attacks produced significantly larger perturbations than transferability-based attacks ($p<0.05$), consistent with the visual trends in \cref{fig:lp_norm}. 
Non-significant differences appeared only within the transferability group, where multiple variants produced similarly small distortions, and on the VideoTQ dataset, where the boundary attack in some cases reached $L_0$ and $L_2$ values comparable to those of transfer-based attacks. 
Detailed test results are provided in Appendix~\ref{appendix:appendixC}.

In conclusion, while query-based attacks tend to achieve higher attack success rates, they have higher operational costs due to the large number of queries required. On the other hand, transferability-based attacks, despite their lower success rates on the target models, present a lower risk in terms of query overhead and perturbation efficiency, as demonstrated by their reduced $L_0$ and $L_2$ distances, especially when feature-importance-based selection is applied.

\subsection{Attacker's Effort}
To assess the attacker's effort, we compared the performance of the query- and transferability-based attacks in terms of the computational time, the required amount of data, and the computational resources needed(see \cref{subsec-metrics}).

With respect to the \textit{computational time}, the attacker's objective is to minimize the amount of time it takes to craft an adversarial sample.
In query-based attacks, the computational time includes: 
\textit{i)} the time it takes to query the target model (denoted as $\alpha$);
and \textit{ii)} the time it takes to perform a single optimization iteration (denoted as $\beta$).
The total amount of time it takes to craft an adversarial sample is calculated in~\cref{equation:total_time_query}:
\begin{equation}
\centering
\label{equation:total_time_query}
Time(x,M) = nt^2\alpha\beta
\end{equation}
where $x$ is a data sample, $M$ is the target model, $n$ is the number of queries required to craft a successful adversarial sample, and $t$ is the time it takes to perform a single query.
When considering solely the query-related aspects of the required time, query-based attacks are at a disadvantage compared to transferability-based attacks, since many more queries to the target model are required (see \cref{subsec:results_risk}).
However, in transferability-based attacks, there is another aspect to consider -- the surrogate model's training and querying time.
The total amount of time it takes to craft an adversarial sample is calculated in~\cref{equation:total_time_t}, where we assume that the time it takes to perform a query to the target model is identical to the time required to perform a query to the surrogate model.

\begin{equation}
\label{equation:total_time_t}
Time(x,M) = mt^2\alpha\beta+T_{surrogate}+t
\end{equation}
where $x$ is a data sample, $M$ is the target model, $m$ is the number of queries to the surrogate model required to craft a successful adversarial sample, and $T_{surrogate}$ is the time it takes to train the surrogate model.
In transferability-based attacks that rely on an additional feature importance model, the total time must also include the time it takes to train that model.

In our evaluation, we found that the total amount of time it took to produce a single adversarial sample in query-based attacks was, on average, $518.4$ seconds.
However, in transferability-based attacks, it was $\sim$$243.5673$ seconds: $\sim$$240.49$ seconds to train the surrogate and feature importance models, $0.0773$ seconds to generate the adversarial sample using the surrogate model, and $\sim$$3$ seconds to query the target model.
Based on this, we can conclude that although transferability-based attacks require training additional models, the total computational time required to craft a single adversarial sample is shorter than in query-based attacks.

With respect to the \textit{amount of data} required, the attacker's objective is to minimize the amount of data.
In query-based attacks, the only data required by the attacker is the original sample that the adversarial sample will be based on, however in transferability-based attacks, the attacker must also possess a surrogate dataset to train the surrogate and feature importance models.
Based on this, we can conclude that transferability-based attacks require more data samples than query-based attacks.

In our evaluation, the surrogate model was trained with the exact same amount of data used to train the target models ($3,320$, $62,848$, and $33,727$ samples from the Hate, ICU, and VideoTQ datasets respectively), and achieved satisfactory performance.

With respect to \textit{computational resources}, the attacker's objective is to minimize the amount of computational resources needed.
In query-based attacks, the required resources include computational power for the adversarial sample generation process and storage to store the original and adversarial samples and the byproducts of the process (e.g., the target model's response), which can be substantial when dealing with a large number of queries.
In contrast, transferability-based attacks require the resources mentioned above, as well as additional storage and computational resources to train and store the surrogate and feature importance models.
Based on this, we can conclude that transferability-based attacks require more computational resources than query-based attacks.

In conclusion, while query-based attacks require less data, transferability-based attacks require less computational time and resources. However, they are limited by lower success rates compared to query-based attacks, as discussed in \cref{subsec:results_risk}.

\subsection{Attack Quality}

In our evaluation, we assessed query- and transferability-based attacks in terms of the quality of the adversarial samples produced. This quality can be examined by considering both the adversarial sample's coherence and its impact on the target model's decision-making process.
To measure the coherence of adversarial samples, we propose using metrics based on AE and IF models. To evaluate the impact on the target model's decision-making process, we introduced two new metrics derived from the SHAP framework (see \cref{subsec-metrics}).

\textbf{Feature Space Coherency.}\label{subsub:coherence}

\begin{figure*}[!htbp]
  \centering

  \begin{subfigure}[b]{0.48\textwidth}
    \centering
    \includegraphics[width=\linewidth]{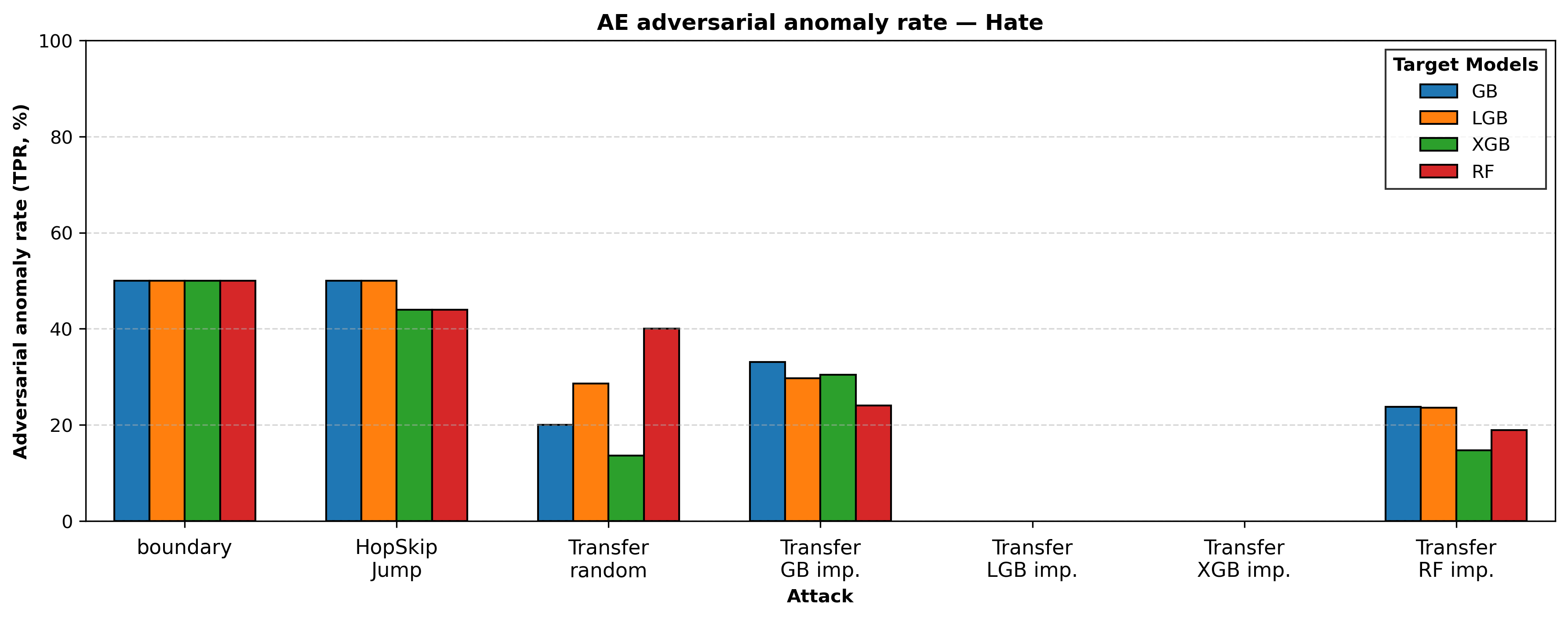}
    \caption{Hate — AE}
    \label{fig:anomaly-hate-ae}
  \end{subfigure}\hfill
  \begin{subfigure}[b]{0.48\textwidth}
    \centering
    \includegraphics[width=\linewidth]{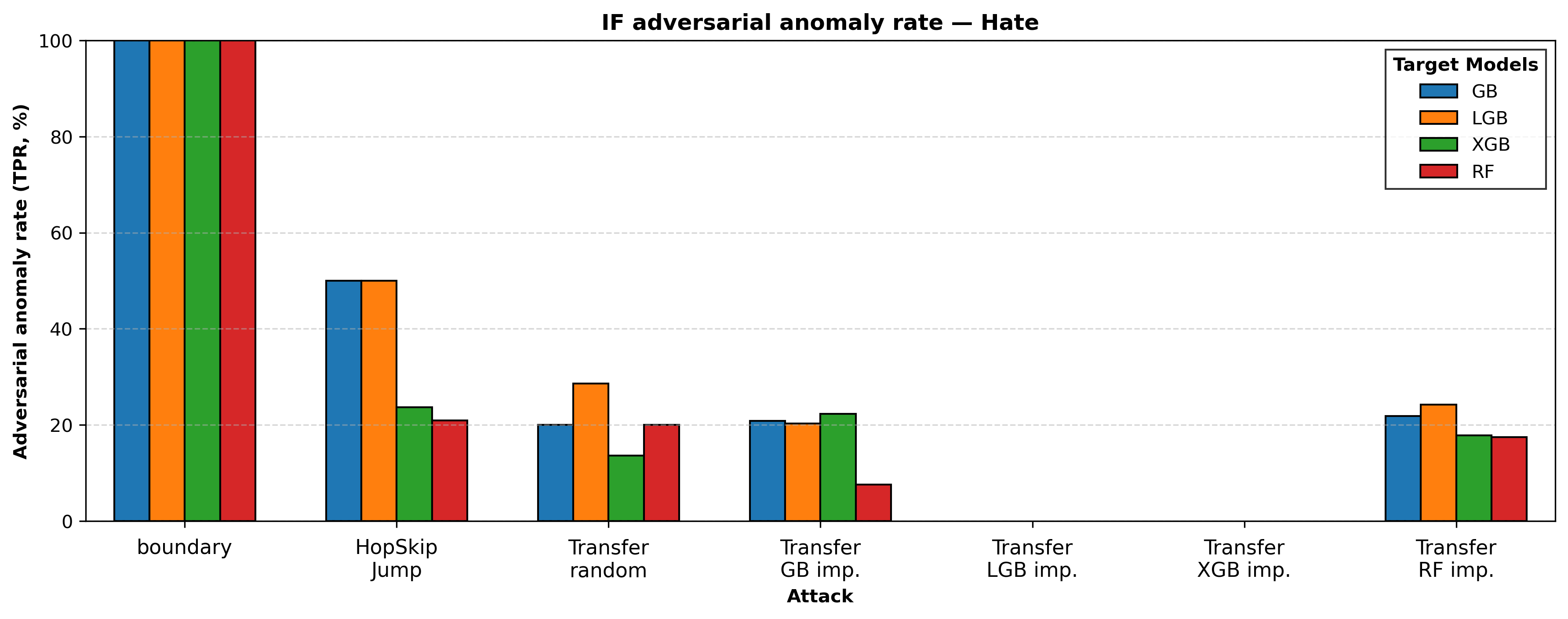}
    \caption{Hate — IF}
    \label{fig:anomaly-hate-if}
  \end{subfigure}

  \vspace{0.6em}

  \begin{subfigure}[b]{0.48\textwidth}
    \centering
    \includegraphics[width=\linewidth]{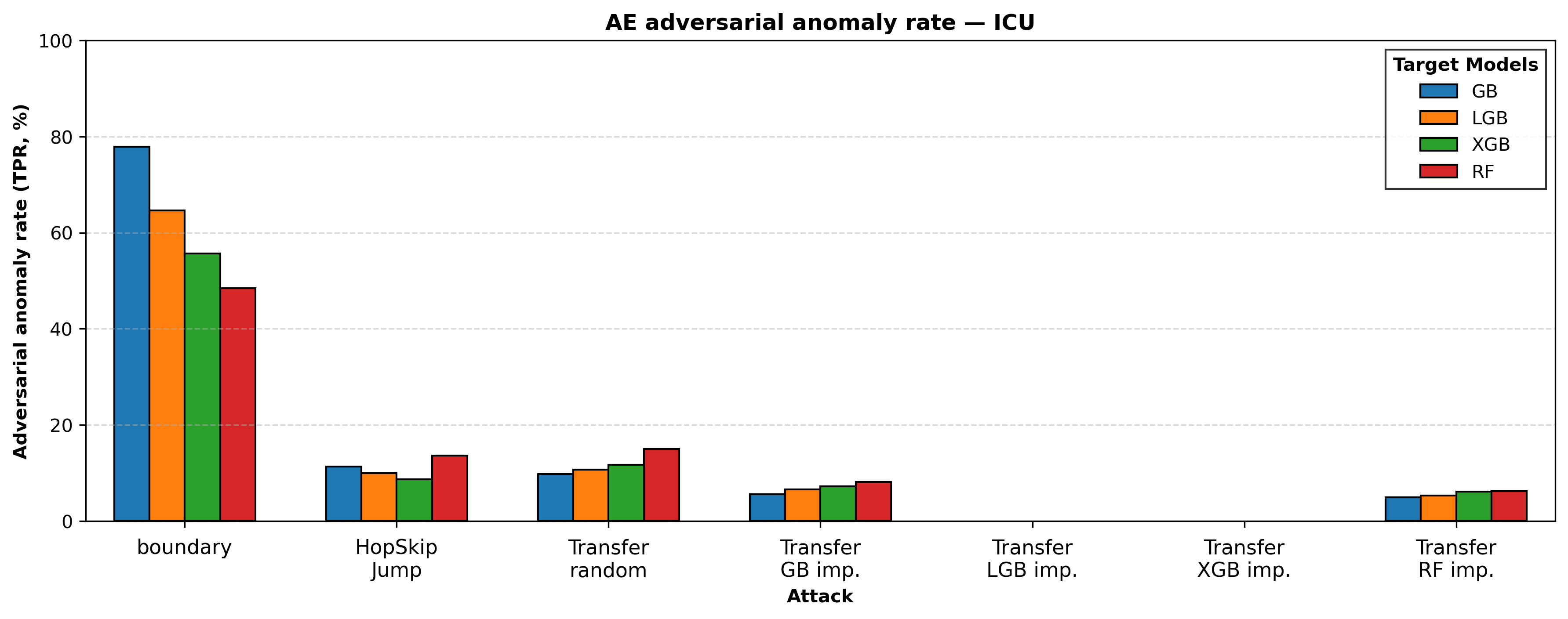}
    \caption{ICU — AE}
    \label{fig:anomaly-icu-ae}
  \end{subfigure}\hfill
  \begin{subfigure}[b]{0.48\textwidth}
    \centering
    \includegraphics[width=\linewidth]{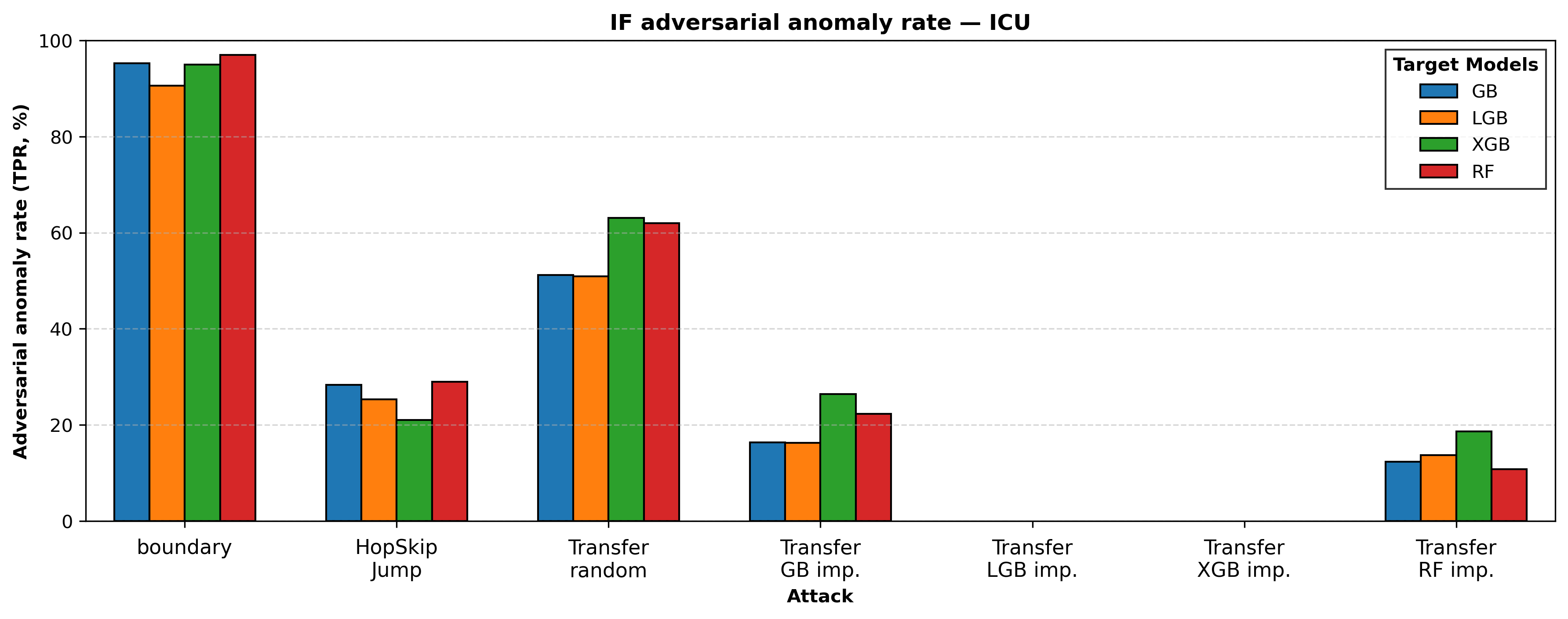}
    \caption{ICU — IF}
    \label{fig:anomaly-icu-if}
  \end{subfigure}

  \vspace{0.6em}

  \begin{subfigure}[b]{0.48\textwidth}
    \centering
    \includegraphics[width=\linewidth]{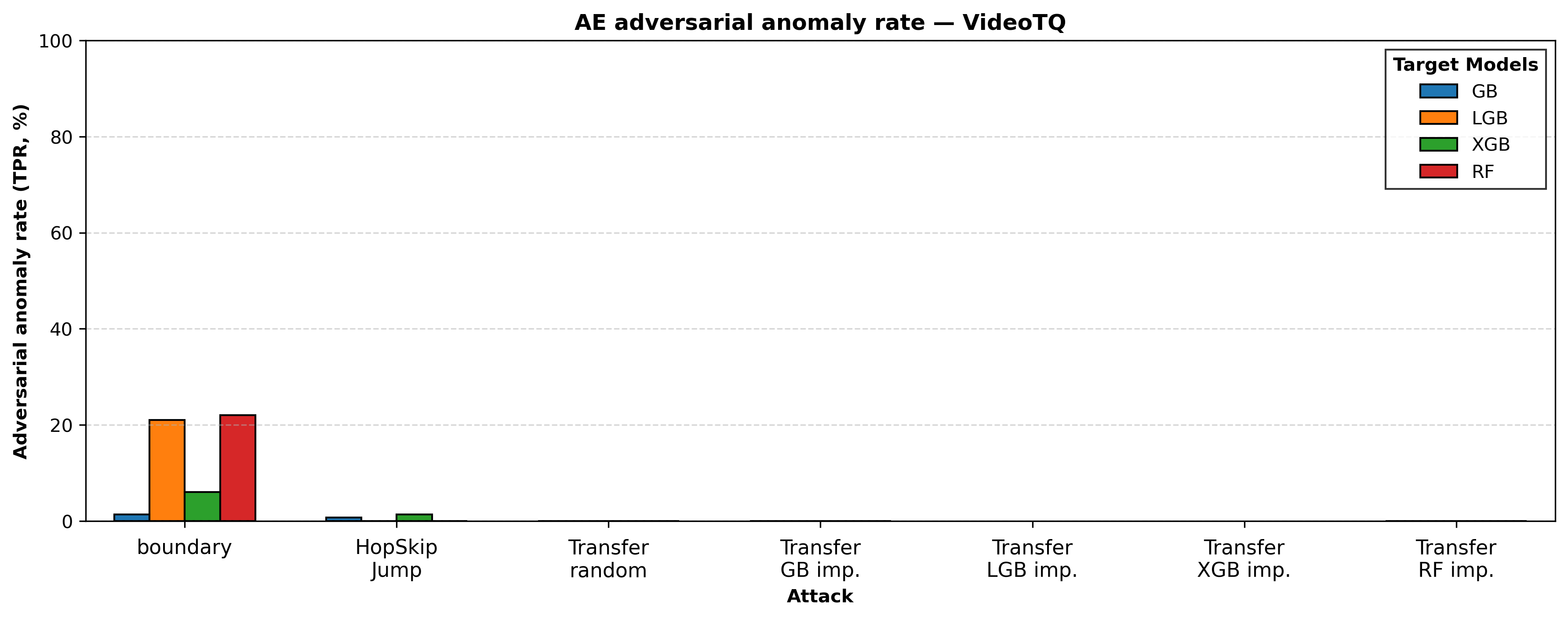}
    \caption{VideoTQ — AE}
    \label{fig:anomaly-videotq-ae}
  \end{subfigure}\hfill
  \begin{subfigure}[b]{0.48\textwidth}
    \centering
    \includegraphics[width=\linewidth]{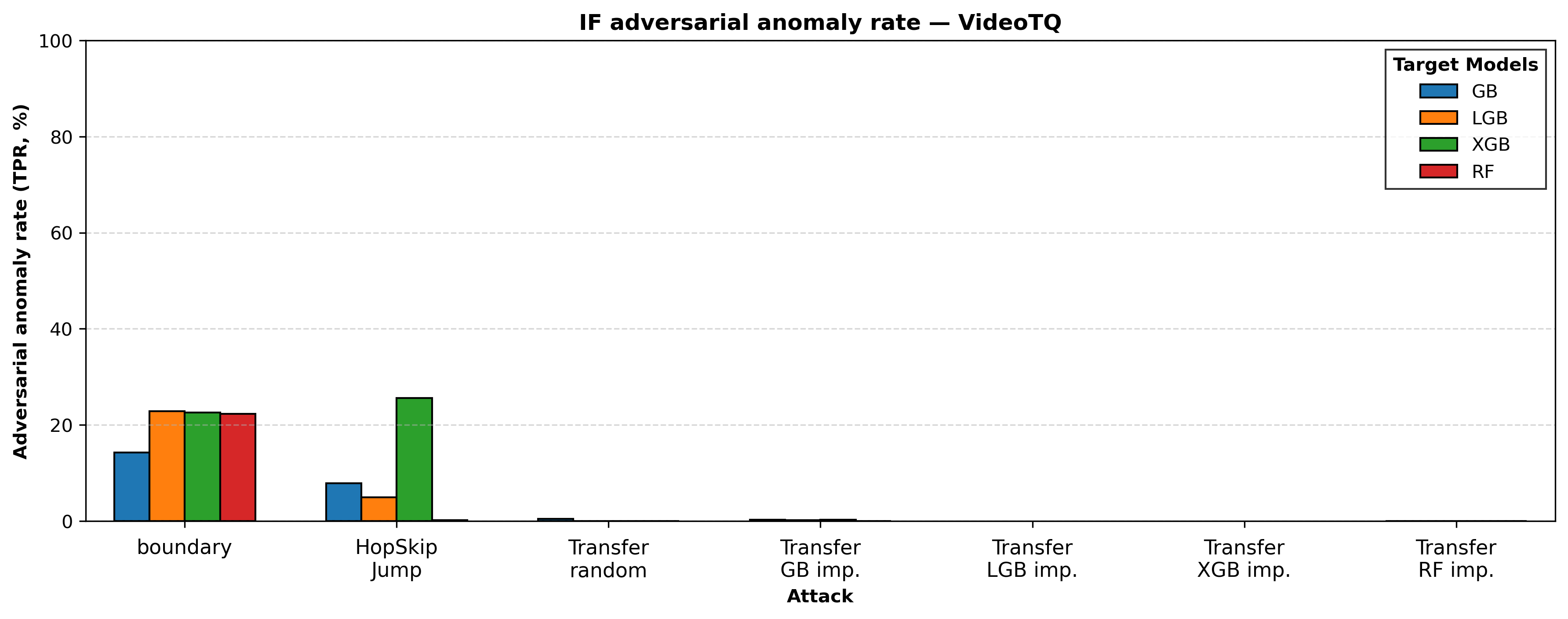}
    \caption{VideoTQ — IF}
    \label{fig:anomaly-videotq-if}
  \end{subfigure}

  \caption{\textbf{Attack quality} evaluated based on the feature space anomaly detection.
  Bars show the percentage of adversarial samples flagged as anomalies (TPR; higher is better) by an AE and an IF for each attack and target model, across datasets.
  Thresholds are fixed per dataset and class on a benign validation split; the resulting benign FPR is less than 5\% across all datasets and classes.}
  \label{fig:anomaly}
\end{figure*}

\cref{fig:anomaly} presents the anomaly detection rates obtained by applying IF anomaly scores and AE reconstruction errors to adversarial samples across different attacks, target models, and datasets. Each detection rate represents the true positive rate (TPR), indicating the percentage of adversarial samples identified as anomalous. For the AE, a sample is considered anomalous if its reconstruction error exceeds a predefined threshold (see ~\cref{equation:ae_threshold}). The resulting false positive rate (FPR) for benign test samples is small and fixed within a dataset: 0.73\% and 4.99\% for class 0 and class 1 on the Hate dataset, 3.06\% and 3.58\% on the ICU dataset, and 0.07\% and 1.45\% on the VideoTQ dataset. For comparability, IF was calibrated to the same FPR levels per dataset and class. Given that the FPR values are consistently very low (ranging between 0.07\% and 4.99\% across all datasets), the reported detection rates effectively represent the performance of each method; the false negative rate (FNR) can be directly obtained as $1-\mathrm{TPR}$.

On the Hate dataset, the boundary attack reveals a distinct contrast between the two detection methods: IF consistently achieved a $100\%$ detection rate across all target models, while AE reached only $50\%$, driven entirely by detecting adversarial samples from class~1, with no anomalies detected in class~0. For the HopSkipJump attack, both detectors achieved around $50\%$ anomaly detection when targeting the GB and LGB models, while anomaly detection dropped when targeting the XGB and RF models with AE outperforming IF. A similar pattern emerged, with anomalies detected only from class~1 and complete failure on class~0. In contrast, transferability-based attacks achieved much lower detection rates than query-based attacks. For random-based selection attacks, AE was moderately more sensitive than IF, with detection rates of approximately $20$–$40\%$ for AE compared to $14$–$29\%$ for IF; detections were almost exclusively from class~1. For importance-based selection attacks, both detectors performed poorly: GB and RF importance-based attacks (GB imp., RF imp.) yielded at most $15$–$25\%$ anomaly detection by AE and $10$–$20\%$ by IF, while LGB and XGB importance-based selection attacks (LGB imp., XGB imp.) were essentially undetected by either detector.

On the ICU dataset, the boundary attack demonstrated similar performance trends as observed for the Hate dataset, with IF achieving very high anomaly detection rates of $95$–$97\%$ and AE reaching $50$–$78\%$. 
In contrast, the HopSkipJump attack was less detectable, with IF averaging $20$–$30\%$ detection and AE rarely exceeding $10$–$15\%$. 
The transferability-based attacks also led to substantially fewer detected anomalies, with the exception of the random-based selection attack, which was identified at moderate levels by IF ($50$–$65\%$) but only weakly by AE. 
Importance-based selection attacks were largely ineffective for both detectors, with detection rates below $30\%$. 

On the VideoTQ dataset, query-based attacks achieved only low anomaly detection rates, with boundary attacks reaching at most $\approx20$–$25\%$ and HopSkipJump slightly lower, where IF achieved better detection rates than AE. 
In contrast, transferability-based attacks were essentially undetected by both detectors, with detection rates close to $0\%$ across all variants.

According to proportions z–tests with Holm correction, the observed differences in anomaly detection rates were statistically significant, with IF detecting query-based attacks at substantially higher rates than transferability-based attacks across all datasets ($p<0.001$), with only isolated exceptions, such as the random-based selection attack on the ICU dataset. A similar trend was observed for AE: detection rates for query-based attacks were consistently and significantly higher than for transferability-based attacks ($p<0.001$), despite AE achieving overall lower detection levels than IF, particularly on boundary and HopSkipJump attacks. Full pairwise results for both detectors are provided in Appendix~\ref{appendix:appendixC}.

As shown in \cref{fig:anomaly}, the two query-based attacks demonstrate distinct behaviors in terms of the anomaly detection rate.
Most adversarial samples generated by the boundary attack were detected as anomlies, particularly by IF. 
In contrast, adversarial samples from the HopSkipJump attack proved much harder to detect by both models, closer in magnitude to those observed for transferability-based attacks.

These results reveal two consistent patterns. 
First, query-based attacks, particularly the boundary attack, are substantially easier to detect, with IF outperforming AE. 
Transferability-based attacks, by contrast, remain less detectable by both models across all datasets, with only isolated cases where AE achieves a small advantage. 
Second, the comparison between AE and IF indicates that IF is systematically more effective in capturing adversarial samples that are ``far'' or structurally shifted in the feature space, as reflected in its dominance on boundary and random-based selection attacks. AE, in contrast, showed only sporadic and inconsistent advantages, with its effectiveness largely limited outside boundary attack scenarios. 
The substantial gap between query- and transferability-based anomaly detection underscores the persistent challenge posed by transferability-based gradient attacks, particularly in realistic settings where these attacks are almost entirely undetectable as anomalies.

Notably, despite HopSkipJump and transferability-based attacks often producing adversarial samples with lower $L_2$ distances (as detailed in \cref{subsec:results_risk}), these samples were still identifiable as anomalous in some cases. 
This suggests that even subtle perturbations introduced by these attacks can be identified as anomalous.

\textbf{Model Interpretation Stability.}\label{subsub:shap}

\begin{figure*}[!htbp]
    \centering
    
    \begin{subfigure}[b]{0.47\textwidth}
        \centering
        \includegraphics[width=\textwidth]{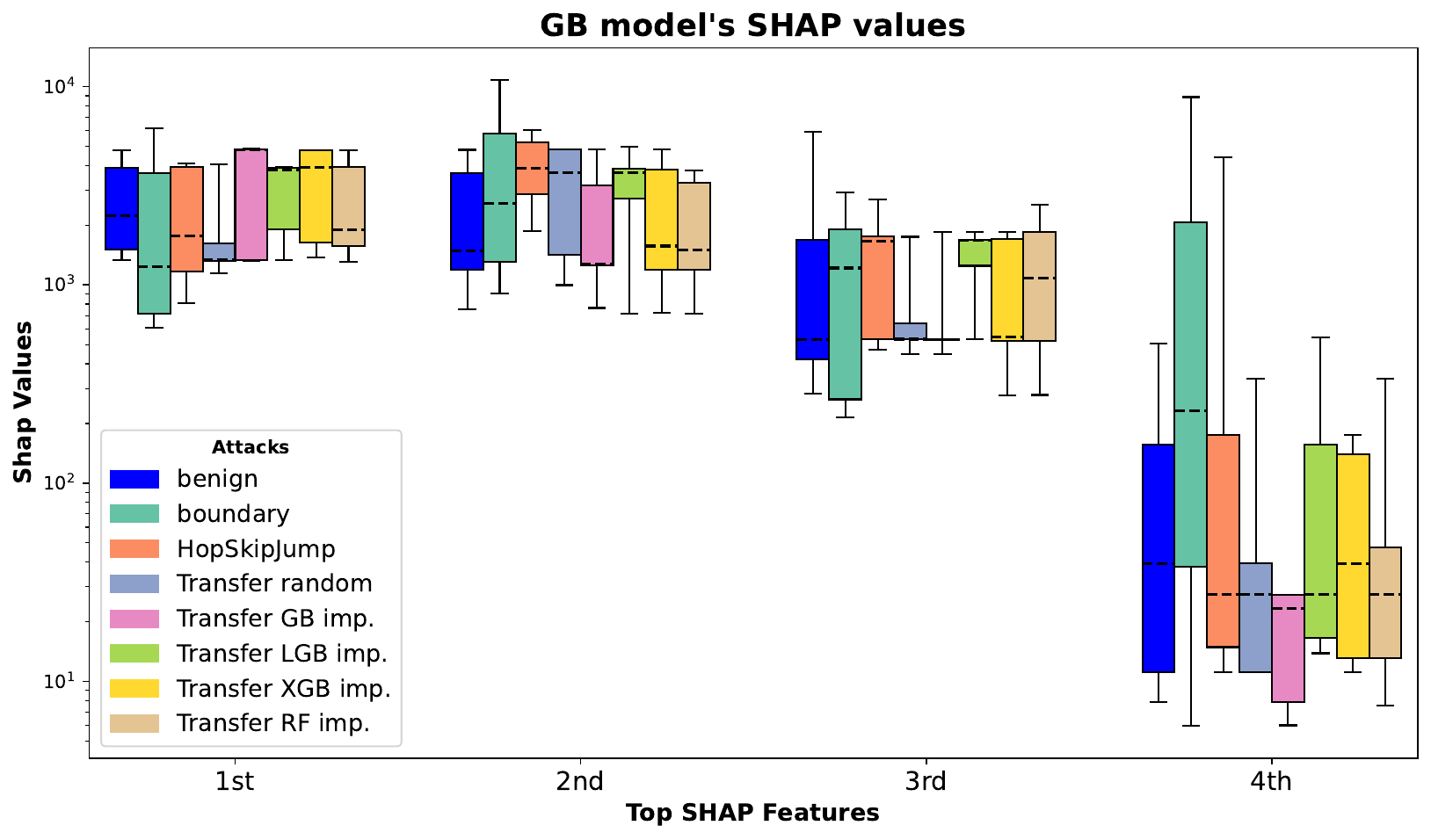}
        \label{fig:shap_gb_hate}
    \end{subfigure}
    \begin{subfigure}[b]{0.47\textwidth}
        \centering
        \includegraphics[width=\textwidth]{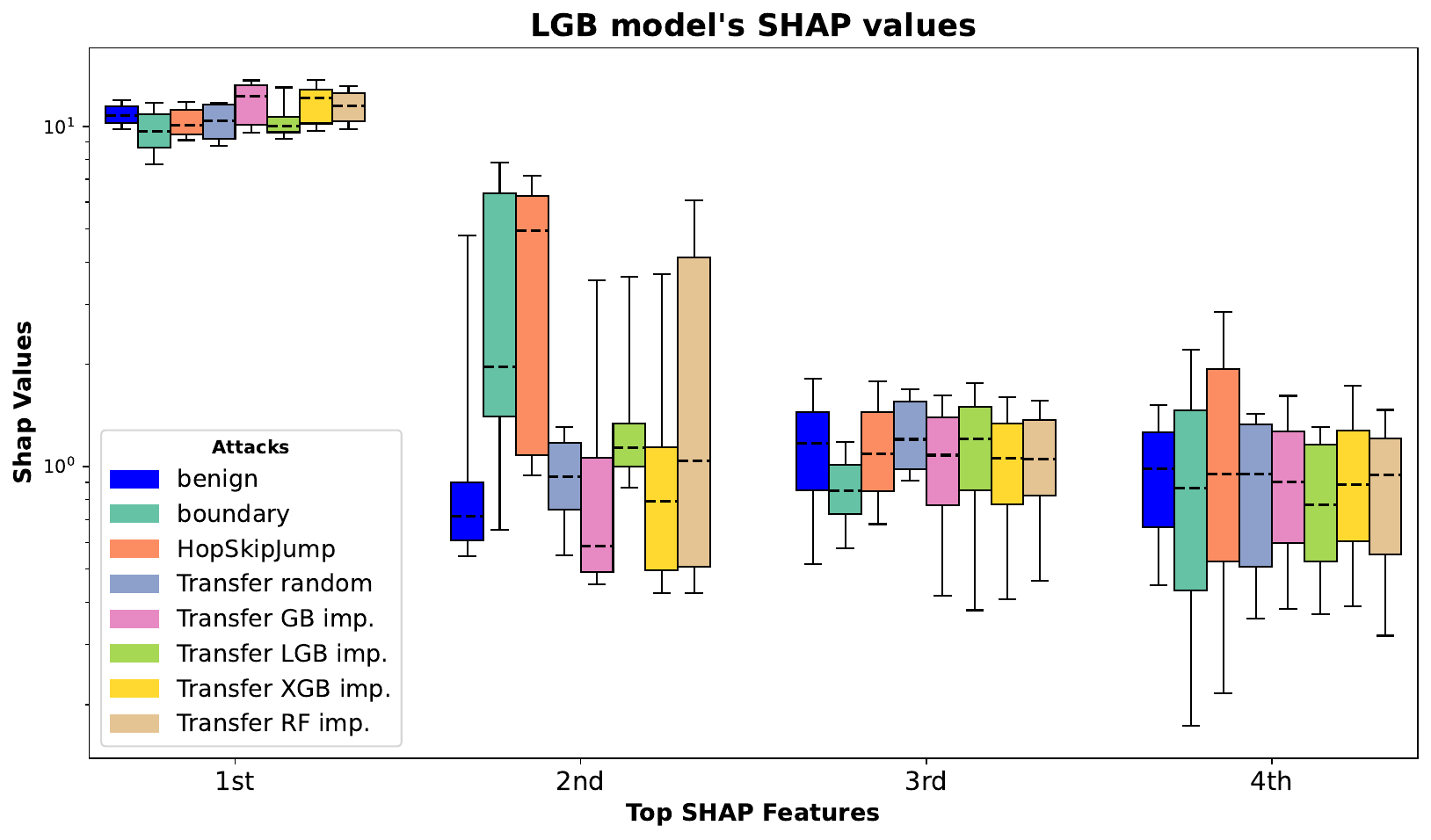}
        \label{fig:shap_lgb_hate}
    \end{subfigure}
    \begin{subfigure}[b]{0.47\textwidth}
        \centering
        \includegraphics[width=\textwidth]{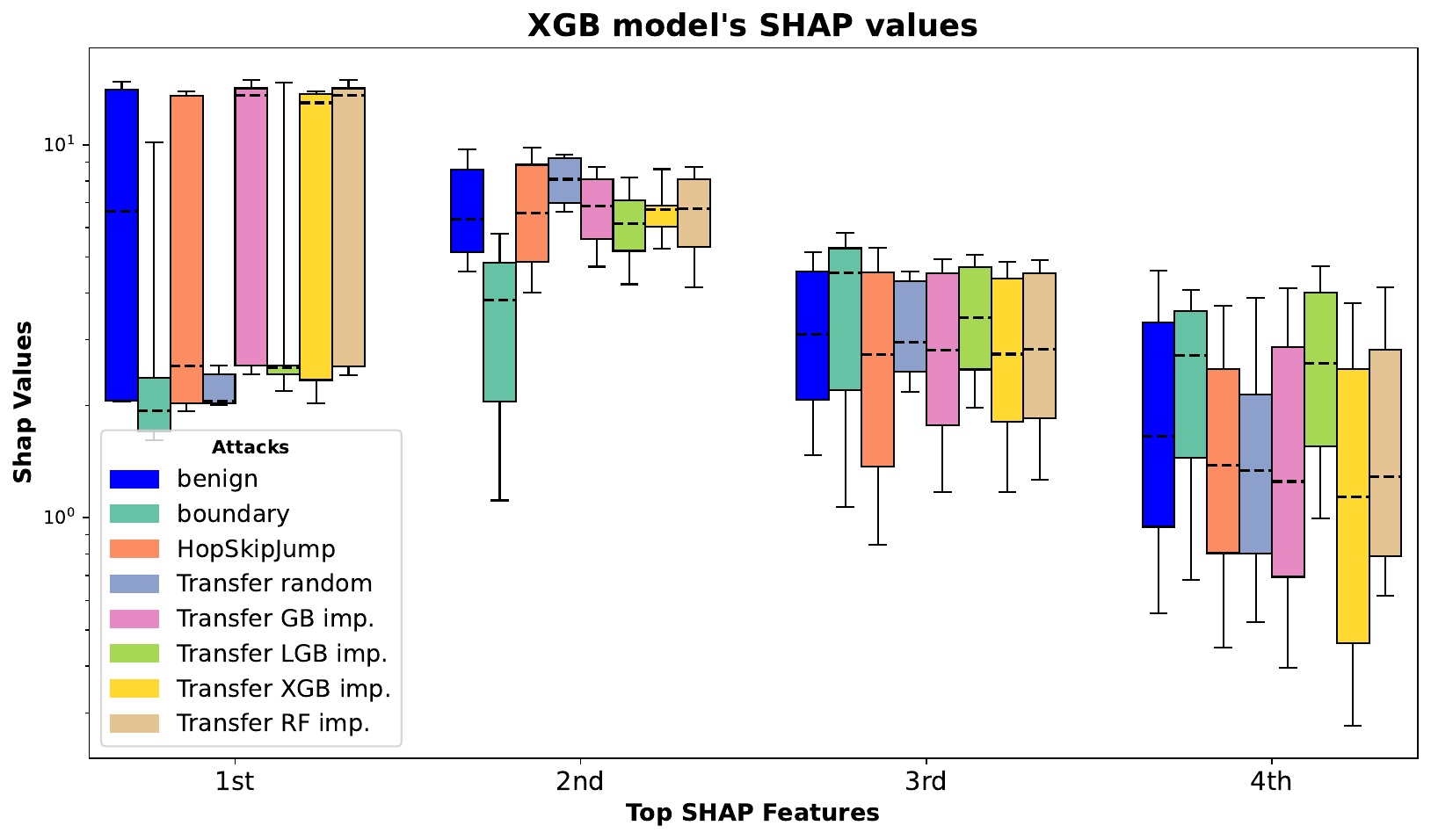}
        \label{fig:shap_xgb_hate}
    \end{subfigure}
    \begin{subfigure}[b]{0.47\textwidth}
        \centering
        \includegraphics[width=\textwidth]{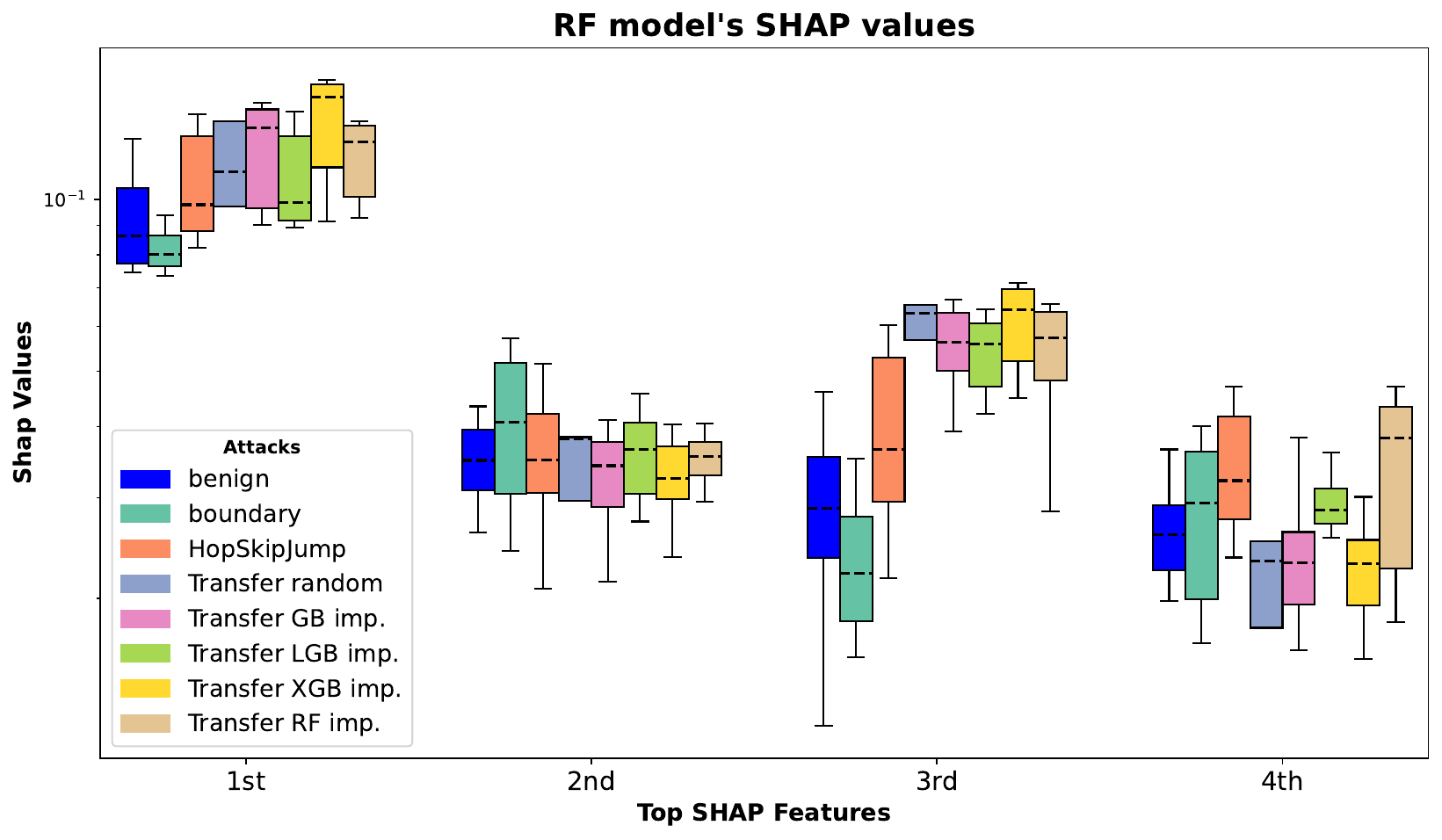}
        \label{fig:shap_rf_hate}
    \end{subfigure}
    
    \caption{\textbf{Attack quality} evaluated based on the impact on the target model's decision-making process; SHAP value distribution for benign (blue) and adversarial samples (on the \textbf{Hate dataset}), across the top-four most important features selected based on their average SHAP values computed on the benign training dataset.}
    \label{fig:shap_hate}
\end{figure*}

\begin{figure*}[!htbp]
    \centering
    
    \begin{subfigure}[b]{0.47\textwidth}
        \centering
        \includegraphics[width=\textwidth]{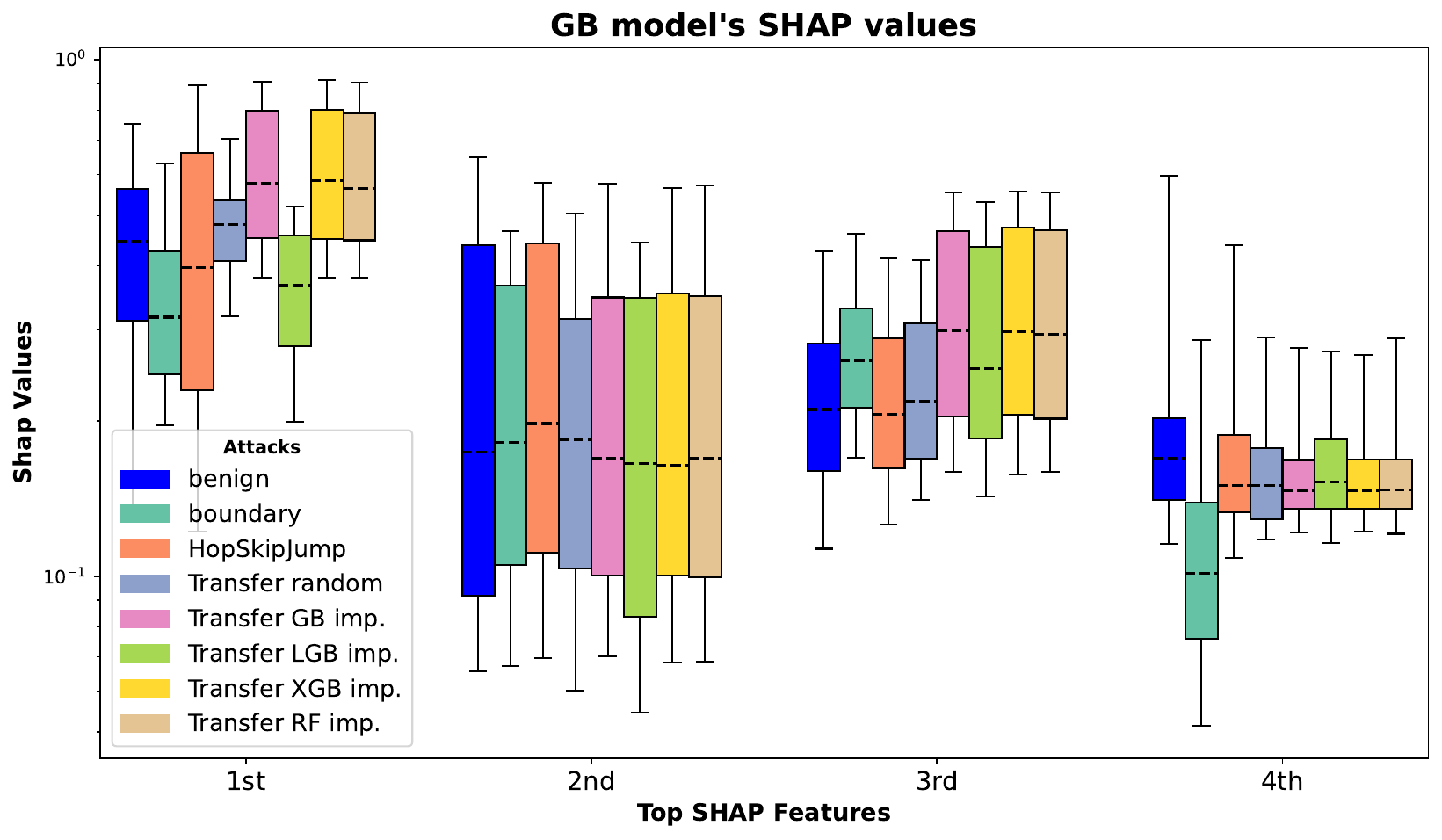}
        \label{fig:shap_gb_icu}
    \end{subfigure}
    \begin{subfigure}[b]{0.47\textwidth}
        \centering
        \includegraphics[width=\textwidth]{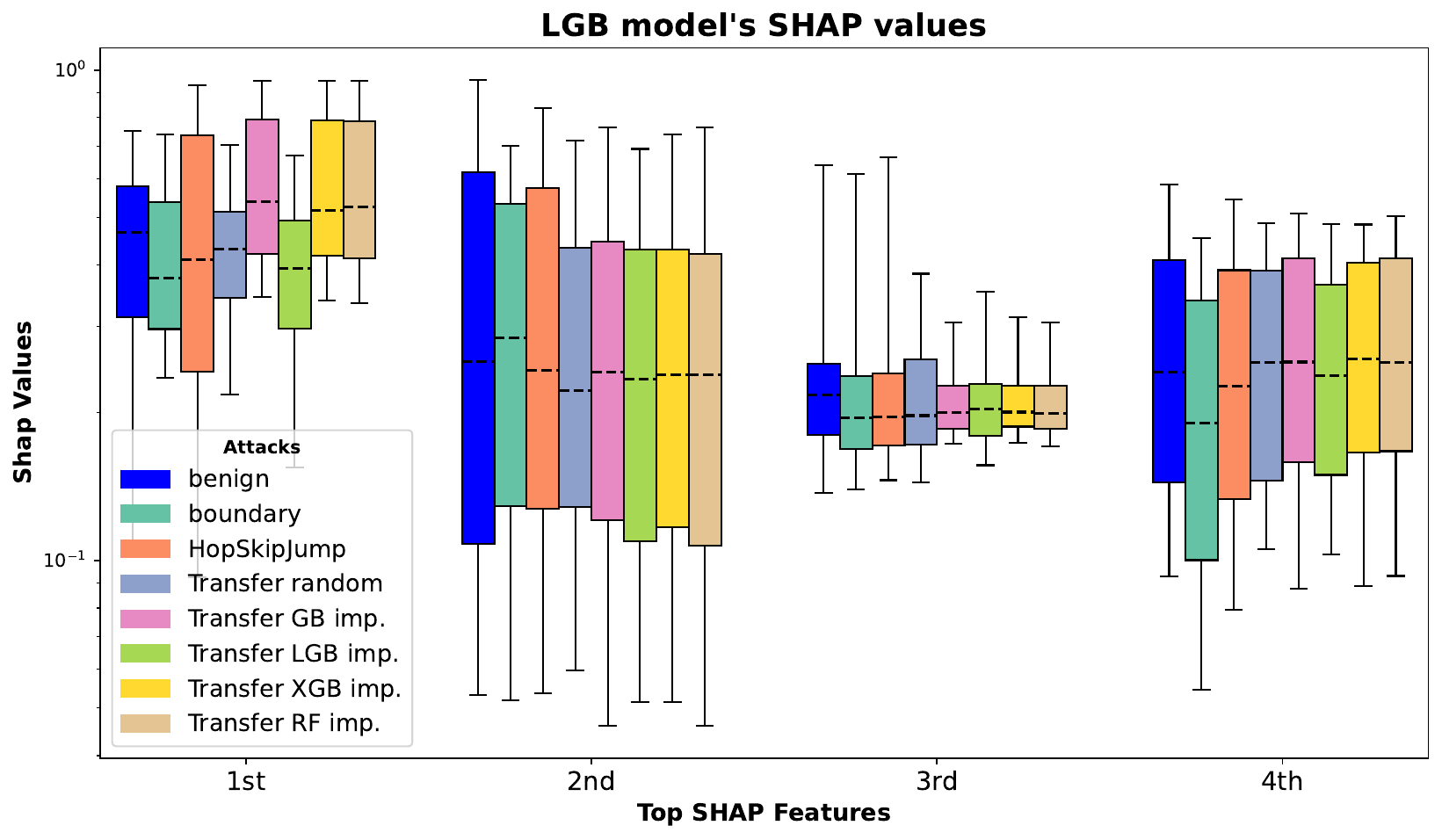}
        \label{fig:shap_lgb_icu}
    \end{subfigure}
    \begin{subfigure}[b]{0.47\textwidth}
        \centering
        \includegraphics[width=\textwidth]{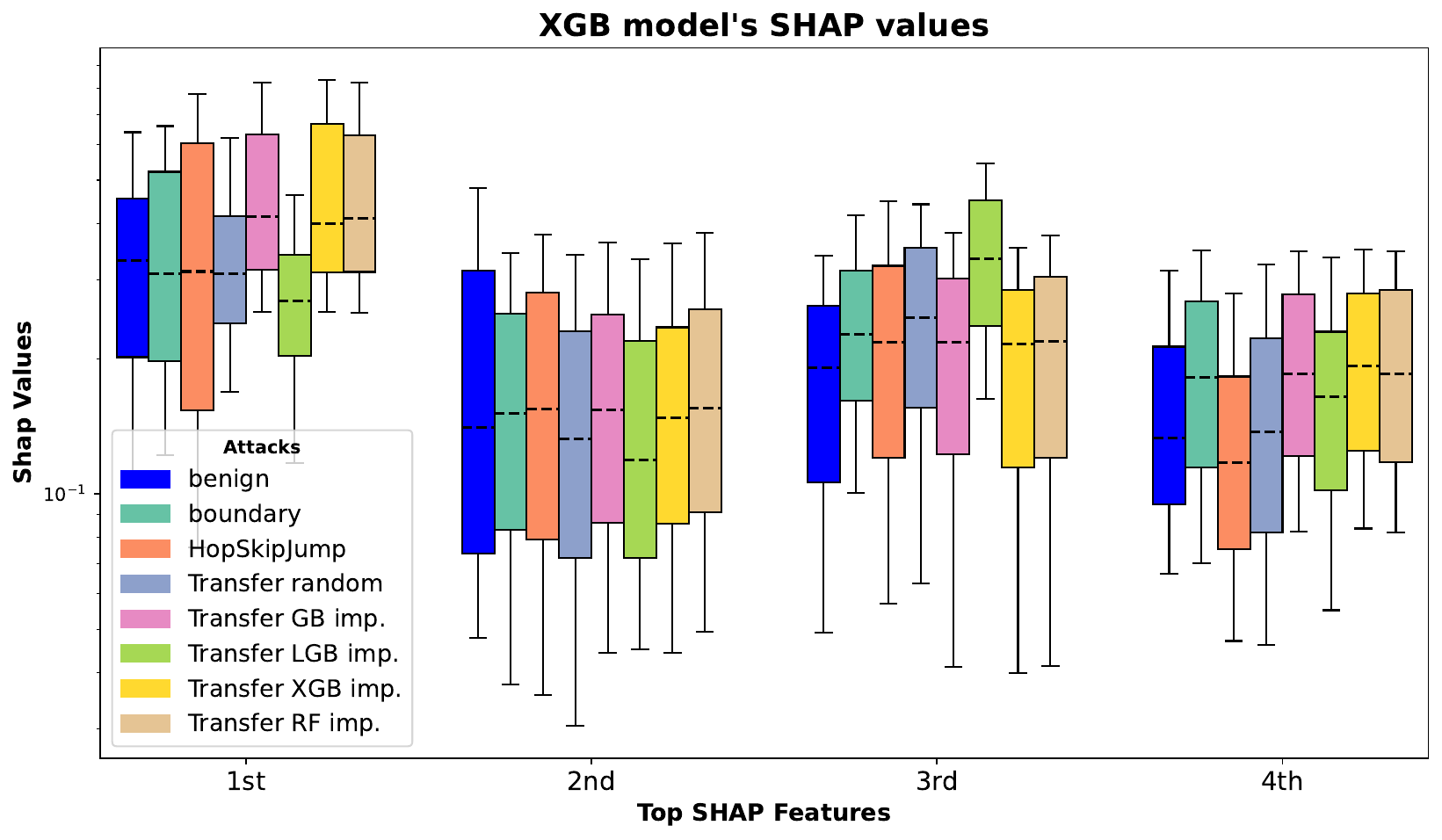}
        \label{fig:shap_xgb_icu}
    \end{subfigure}
    \begin{subfigure}[b]{0.47\textwidth}
        \centering
        \includegraphics[width=\textwidth]{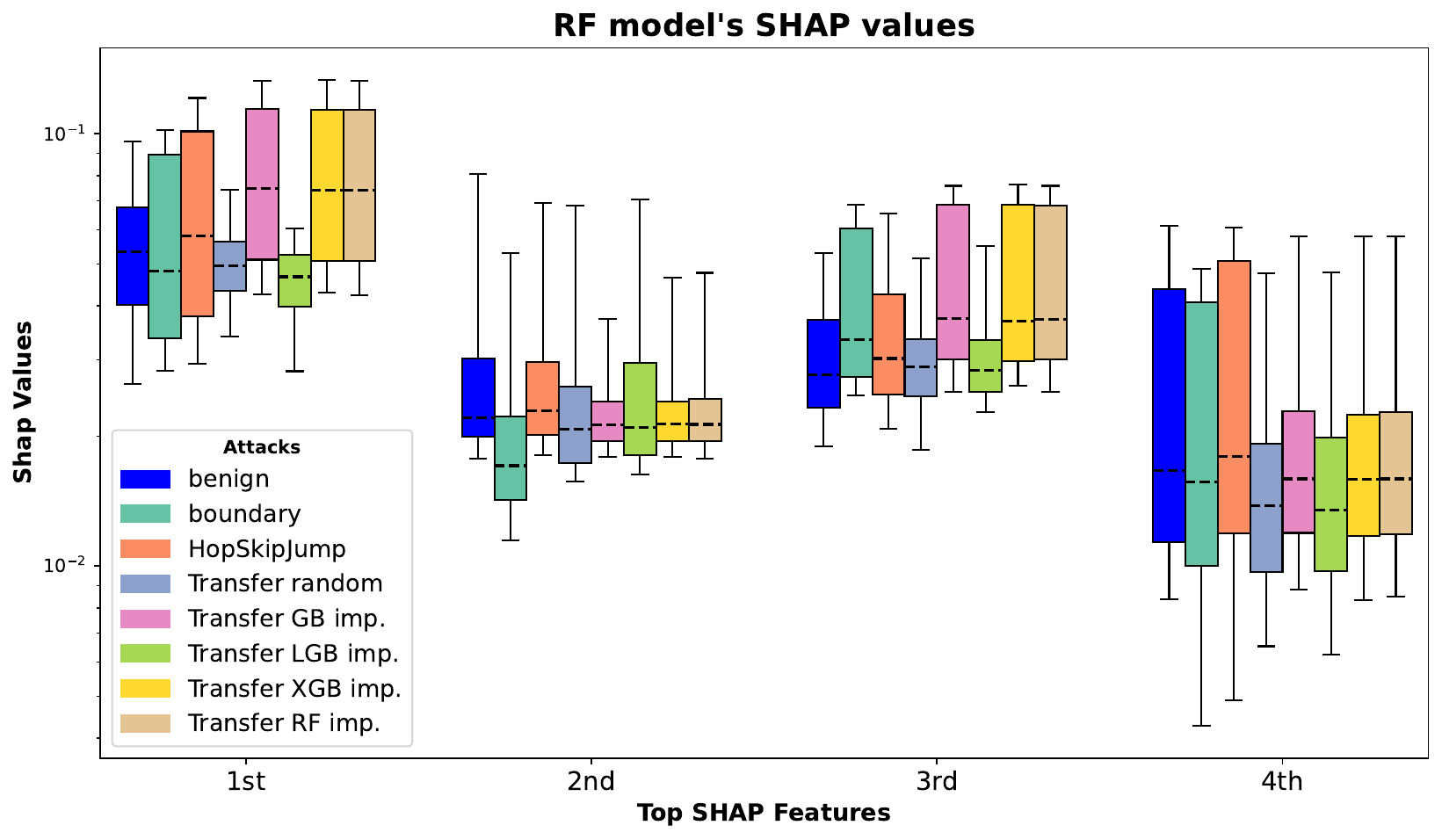}
        \label{fig:shap_rf_icu}
    \end{subfigure}
    
    \caption{\textbf{Attack quality} evaluated based on the impact on the target model's decision-making process; SHAP value distribution for benign (blue) and adversarial samples (on the \textbf{ICU dataset}), across the top-four most important features selected based on their average SHAP values computed on the benign training dataset.}
    \label{fig:shap_icu}
\end{figure*}

\begin{figure*}[!htbp]
    \centering
    
    \begin{subfigure}[b]{0.47\textwidth}
        \centering
        \includegraphics[width=\textwidth]{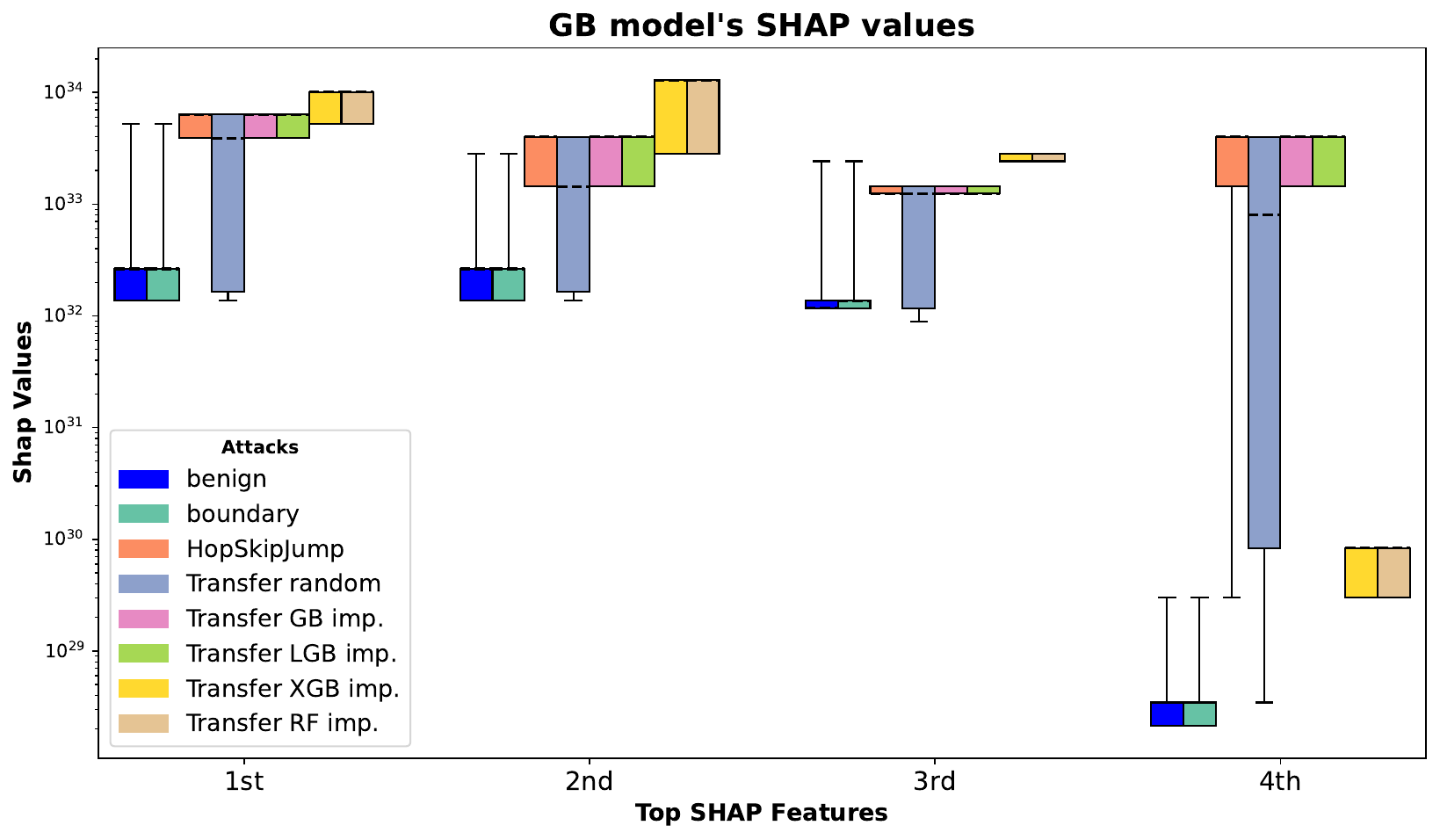}
        \label{fig:shap_gb_VideoTQ}
    \end{subfigure}
    \begin{subfigure}[b]{0.47\textwidth}
        \centering
        \includegraphics[width=\textwidth]{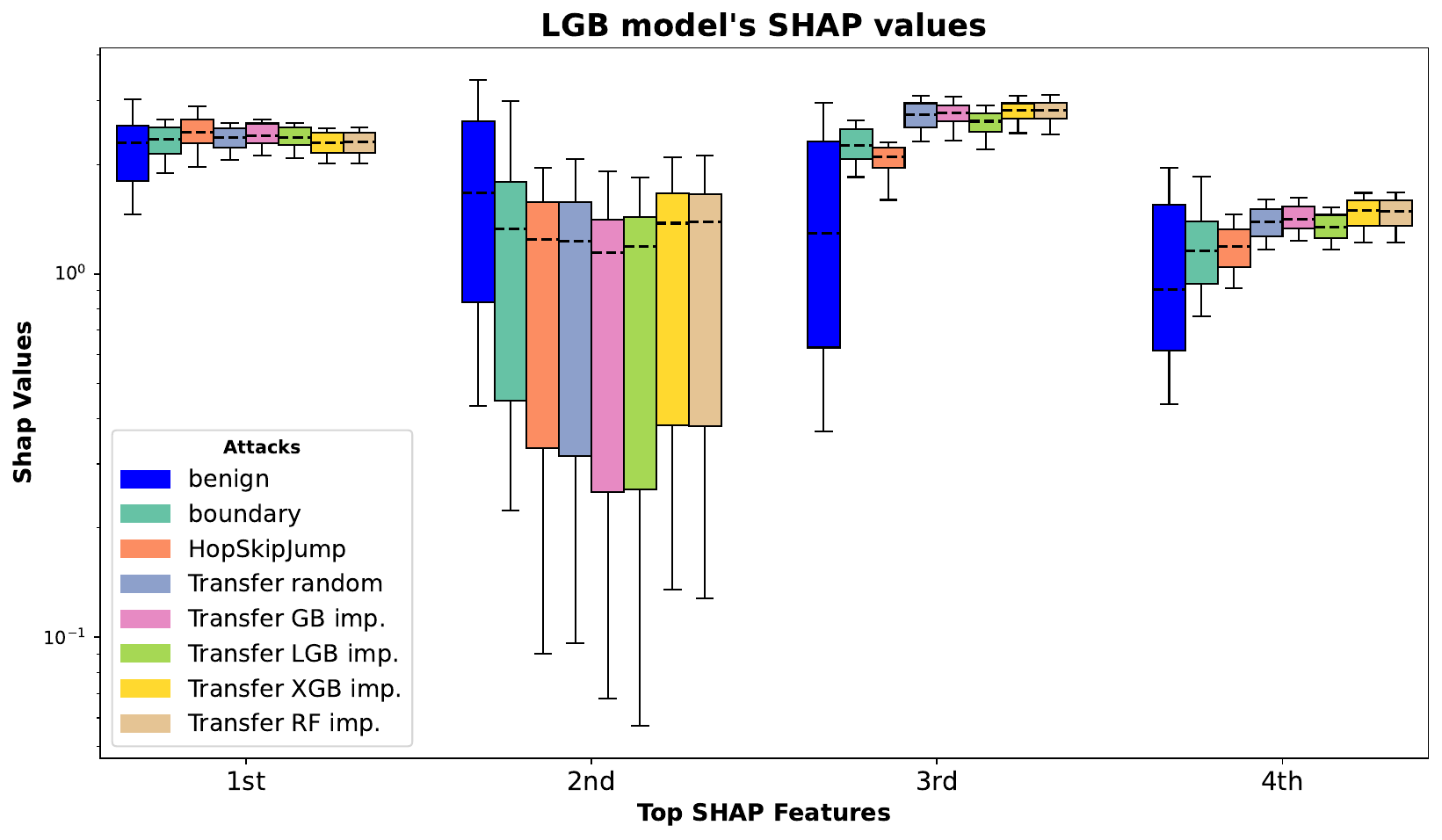}
        \label{fig:shap_lgb_VideoTQ}
    \end{subfigure}
    \begin{subfigure}[b]{0.47\textwidth}
        \centering
        \includegraphics[width=\textwidth]{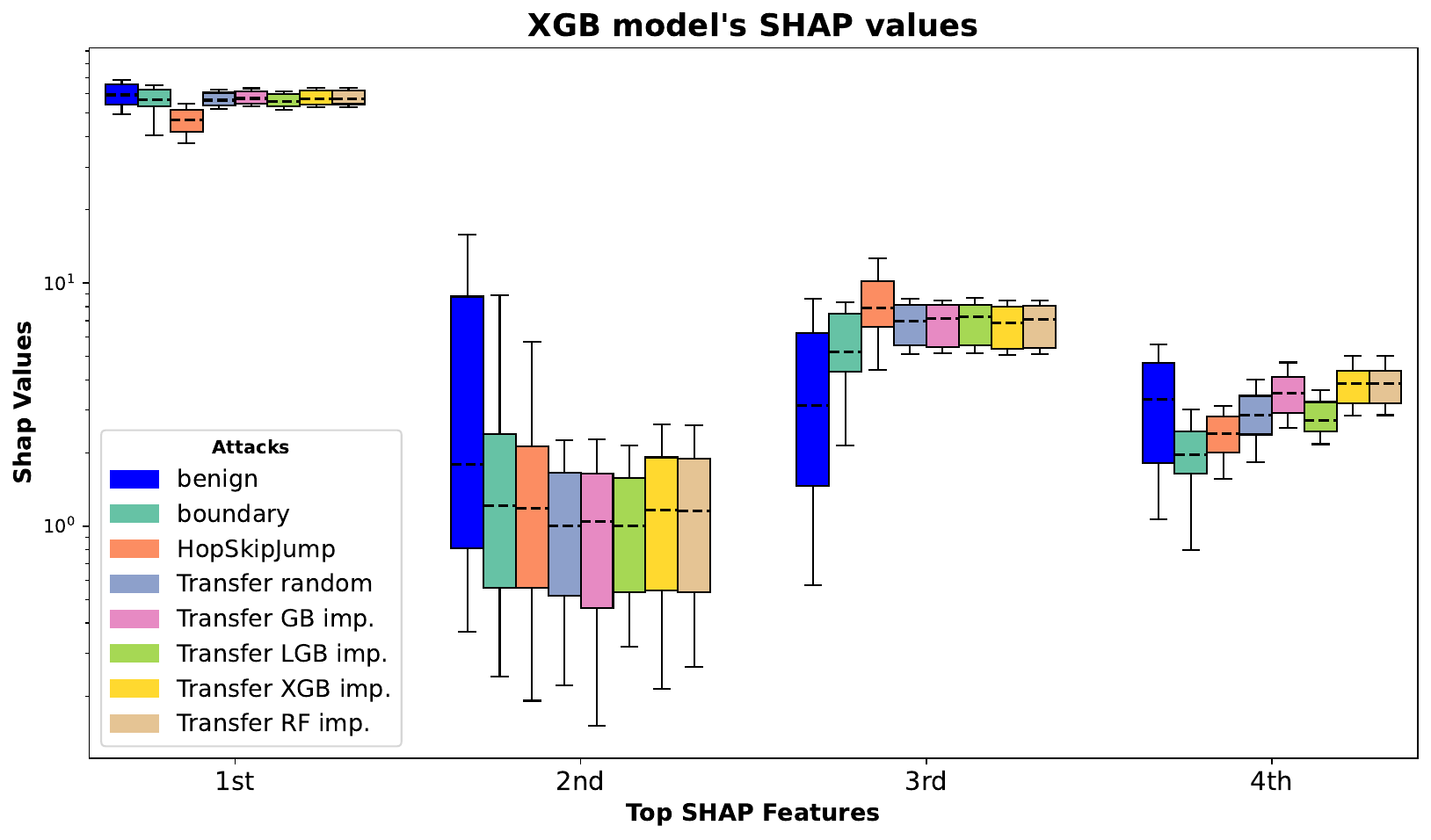}
        \label{fig:shap_xgb_VideoTQ}
    \end{subfigure}
    \begin{subfigure}[b]{0.47\textwidth}
        \centering
        \includegraphics[width=\textwidth]{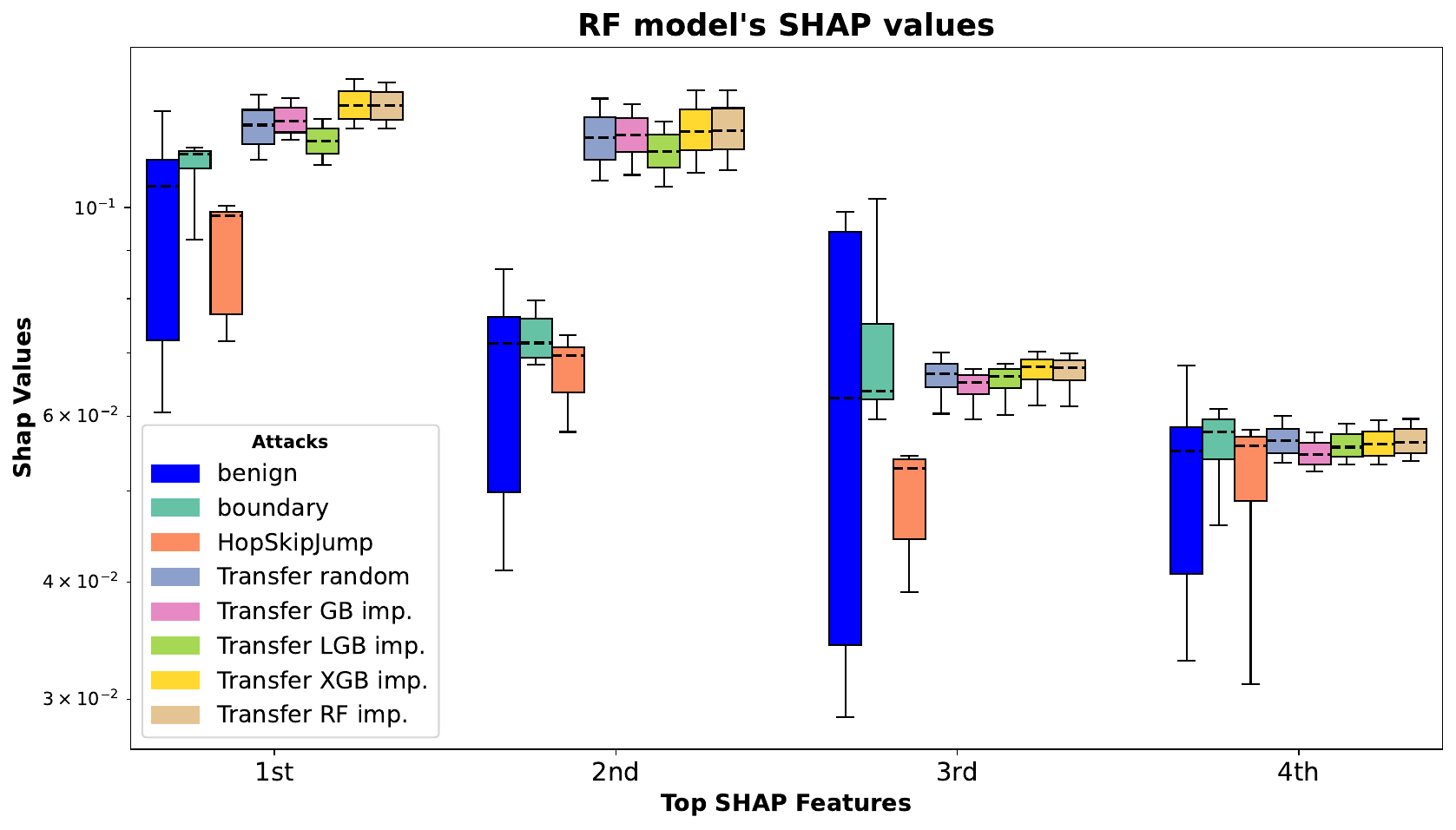}
        \label{fig:shap_rf_VideoTQ}
    \end{subfigure}
    
    \caption{\textbf{Attack quality} evaluated based on the impact on the target model's decision-making process; SHAP value distribution for benign (blue) and adversarial samples (on the \textbf{VideoTQ dataset}), across the top-four most important features selected based on their average SHAP values computed on the benign training dataset.}
    \label{fig:shap_VideoTQ}
\end{figure*}

~\cref{fig:shap_hate,fig:shap_icu,fig:shap_VideoTQ} presents the distribution of SHAP values for adversarial versus benign samples across different attacks and datasets for each target model.
To simplify the presentation, we focus on the top-four most important features, selected based on their average SHAP values computed on the benign training dataset.
As shown in the figures, for most target models and datasets, there is at least one feature where the SHAP value distribution across attacks varies significantly. Analyzing the SHAP-based feature importance of a queried sample can therefore be an effective technique for identifying attacks.

\cref{tab:shap_anomaly} presents the results of our evaluation of attack quality using SHAP-based metrics.
For each attack and target model, we report the Importance-Based Anomaly Detection Rate (percentage of samples with at least one anomalous feature based on SHAP values) and the Average Anomalous SHAP Features per Sample (as described in \cref{subsec-metrics}). Lower values indicate that there is less impact on the model’s decision-making process, meaning that adversarial samples were more consistent with benign samples in terms of feature importance (i.e., the model relies on similar features for prediction).

\begin{table*}
\caption{\textbf{Attack quality:} results based on the \textit{Importance-Based Anomaly Detection Rate} (percentage of samples with at least one feature with an anomalous SHAP value) and the \textit{Average Anomalous SHAP Features per Sample} computed according to the features' SHAP values. \textbf{Bold} values are samples with the fewest changes in SHAP values, indicating high-quality samples in terms of having minimal impact on the target model’s decision process.}
\centering
\resizebox{1\textwidth}{!}{
    \begin{tabular}
    {@{}p{0.07\textwidth}p{0.06\textwidth}ccccccc@{}}
    \toprule
    \multirow{3}{0.07\textwidth}{\textbf{Dataset}} &
    \multirow{3}{0.06\textwidth}{\textbf{Target Model}} &
    \multicolumn{2}{c}{\textbf{Query Attacks}} &
    \multicolumn{5}{c}{\textbf{Transferability Attacks}} \\ 
    \cmidrule(l){3-4} \cmidrule(l){5-9} 
      &&
      \textbf{boundary} &
      \textbf{HopSkipJump} &
      \textbf{random} &
      \textbf{GB imp.} &
      \textbf{LGB imp.} &
      \textbf{XGB imp.} &
      \textbf{RF imp.} \\ 
      ~ & ~ & \multicolumn{7}{c}{Importance-Based Anomaly Detection Rate (\%) / Average Anomalous SHAP Features per Sample} \\ \toprule
\textbf{Hate}   & GB  & 100.0 / 7.7         & 96.7 / 5.6          & 100.0 / 8.4        & 100.0 / 8.9         & 81.2 / 3.3         & \textbf{27.3 / 0.4} & \textbf{25.6 / 0.5} \\
                & LGB & 58.2 / 1.1          & 78.0 / 1.7          & 64.3 / 1.6         & 79.7 / 2.0          & 76.8 / 1.7         & 69.9 / 1.4          & \textbf{40.5 / 0.7} \\
                & XGB & \textbf{26.4 / 0.4} & 67.6 / 1.1          & 86.4 / 1.8         & 92.1 / 1.9          & 80.3 / 1.6         & \textbf{22.0 / 0.3} & \textbf{24.8 / 0.3} \\
                & RF  & 63.2 / \textbf{1.0} & \textbf{30.8 / 0.5} & 100.0 / 3.8        & 100.0 / 4.2         & 75.7 / 3.4         & 67.3 / \textbf{0.9} & \textbf{50.0 / 0.6} \\  \midrule
\textbf{ICU}    & GB  & 29.0 / 0.4          & 19.4 / 0.3          & \textbf{7.3 / 0.1} & \textbf{10.8 / 0.1} & \textbf{1.8 / 0.0} & \textbf{3.5 / 0.0}  & \textbf{3.5 / 0.0}  \\
                & LGB & \textbf{4.1 / 0.0}  & \textbf{3.9 / 0.0}  & 22.3 / 0.2         & 17.5 / 0.2          & \textbf{2.4 / 0.0} & \textbf{6.9 / 0.1}  & \textbf{5.6 / 0.1}  \\
                & XGB & \textbf{3.5 / 0.0}   & \textbf{6.0 / 0.1} & \textbf{6.5 / 0.1} & \textbf{6.4 / 0.1}  & \textbf{4.2 / 0.0} & \textbf{5.1 / 0.1}  & \textbf{3.3 / 0.0} \\
                & RF  & \textbf{0.1 / 0.0}   & \textbf{0.2 / 0.0} & \textbf{4.0 / 0.0} & \textbf{0.5 / 0.0}  & \textbf{0.0 / 0.0} & \textbf{0.0 / 0.0}  & \textbf{0.7 / 0.0} \\  \midrule
\textbf{VideoTQ}& GB  & \textbf{5.3 / 0.1}  & \textbf{9.4 / 0.2} & \textbf{0.4 / 0.0} & \textbf{1.0 / 0.0}  & 51.4 / 0.8         & 90.3 / 1.5          & 64.3 / 1.1          \\
                & LGB & 7.3 / 0.1           & \textbf{1.5 / 0.0}  & 3.4 / 0.0          & 3.4 / 0.0           & \textbf{0.8 / 0.0} & \textbf{0.6 / 0.0}  & \textbf{1.0 / 0.0}  \\
                & XGB & 38.9 / 0.4          & 25.2 / 0.3          & \textbf{0.4 / 0.0} & \textbf{1.0 / 0.0}  & \textbf{21.6 / 0.2} & 31.5 / 0.3        & \textbf{17.8 / 0.2}  \\
                & RF  & 66.6 / 0.7          & 87.6 / 0.9          & \textbf{0.0 / 0.0} & \textbf{0.0 / 0.0}  & \textbf{0.0 / 0.0} & 68.8 / 0.7          & 79.0 / 0.8          \\ \bottomrule
        \end{tabular} }

\label{tab:shap_anomaly}
\end{table*}

On the Hate dataset, transferability-based attacks using XGB and RF feature importance techniques (referred to in the table as XGB imp. and RF imp.) resulted in lower importance-based anomaly detection rates and lower values of average anomalous SHAP features per sample compared to other transferability- and query-based attacks. The boundary attack on the XGB model and HopSkipJump attack on the RF model also resulted in both a low importance-based anomaly detection rate and a low average number of anomalous features per sample. This indicates that the changes to feature importance were less distinguishable for those attacks. 

On the ICU dataset, both query- and transferability-based attacks generally produced less anomalous SHAP values for most target models (LGB, XGB, and RF), with importance-based anomaly detection rates typically below 7\% and very low values of average anomalous SHAP features per sample ranging from 0.0 to 0.1.
However, when targeting the GB model, transferability-based attacks resulted in notably lower importance-based anomaly detection rates (1.8\% to 10.8\%) compared to query-based attacks (29.0\% and 19.4\% for the boundary and HopSkipJump attacks, respectively). In this case, the average number of anomalous SHAP features per sample was slightly higher for query-based attacks (0.3 to 0.4) than for transferability-based attacks (0.0 to 0.1), although still relatively low overall.

On the VideoTQ dataset, transferability-based attacks resulted in less anomalous SHAP values than query-based attacks in most cases, specifically for the GB imp., LGB imp., and RF imp. feature importance techniques, where most transferability attacks resulted in importance-based anomaly detection rates below 1\%. Notably, for the GB target model, both query-based attacks resulted in lower anomalous SHAP values (5.3\% and 9.4\% for boundary and HopSkipJump, respectively), while some transferability-based attacks resulted in higher importance-based anomaly detection rates (51.4\%, 90.3\%, and 64.3\% for GB imp., LGB imp., and RF imp., respectively). All attacks resulted in very few anomalous SHAP features per sample, and in most cases, the number of anomalous features per sample was close to 0.0.

According to Mann–Whitney U tests (for the average anomalous SHAP features per sample) and proportions z–tests (for the importance-based anomaly rate), both with Holm correction, the observed differences were statistically significant across all datasets ($p<0.001$). On the Hate and ICU datasets, query-based attacks yielded substantially higher importance-based anomaly rates than transferability-based attacks, with medium effect sizes. On VideoTQ, query-based attacks also showed higher anomaly rates overall, although the differences in the average anomalous SHAP features per sample were smaller, with significance observed across all target models except under the random-based selection attack when targeting the GB model. Within the transferability-based group, omnibus tests revealed heterogeneity among importance-based selection variants, but no single strategy consistently outperformed the others. Among the query-based methods, the boundary attack generally produced more anomalous SHAP values than HopSkipJump, a difference that reached statistical significance in several cases. Full test results are provided in Appendix~\ref{appendix:appendixC}.

Based on the results in the table, we conclude that while the examined transferability-based attacks were designed to craft subtle and well-suited perturbations and generally result in a low importance-based anomaly detection rate, they do not always outperform query-based methods in generating coherent and consistent adversarial samples. Their effectiveness depends on several key factors, including the importance-based selection method used (i.e., the model from which feature importance is computed), dataset-specific constraints and characteristics, and the target model's architecture.

In highly constrained datasets like the ICU dataset where many features cannot be modified and over 30\% of the features are categorical, transferability attacks tend to produce perturbations of similar magnitude to query-based attacks on categorical features, making query-based approaches competitive in these scenarios. The limited number of editable features forces both query- and transferability-based attacks to make fewer changes, causing less anomalous behavior overall. The random-based selection attack is an exception, as it compromises coherence by arbitrarily modifying features, often resulting in anomalous values and noticeable changes in SHAP values.
These findings suggest that defense mechanisms should consider not only the attack strategy but also the specific dataset and model vulnerabilities.

Our evaluation of attack quality—employing both the anomaly detection rate to assess sample coherence and the importance-based anomaly detection rate to evaluate the impact of adversarial samples on the target model's decision-making process—reveals complementary insights. The Importance-Based Anomaly Detection Rate (computed based on SHAP values, see \cref{tab:shap_anomaly}), can help identify adversarial samples that appear normal when evaluating just their feature space values. For example, on the Hate dataset under the boundary attack against the XGB model, traditional anomaly detection (particularly with IF) achieved nearly perfect detection, whereas the SHAP-based anomaly rate remained very low (see \cref{fig:anomaly,tab:shap_anomaly}). In contrast, on the VideoTQ dataset under the LGB importance-based selection attack, the SHAP-based anomaly rate exceeded $90\%$, while traditional anomaly detection failed almost entirely (see \cref{fig:anomaly,tab:shap_anomaly}). 

These contrasting results underscore the importance of considering anomaly detection rates based on both sample values and feature importance when assessing the impact of adversarial attacks on model robustness. Each approach captures different aspects of adversarial perturbation, providing a more comprehensive understanding of the attack quality and potential defensive strategies.

\subsection{Effectiveness of the Class-Specific Anomaly Detection (CSAD) Approach}

\cref{tab:CSAD_vs_all} presents the results of two approaches for computing the anomaly detection rate: (1) the CSAD approach, which measures the anomalousness of SHAP values in adversarial samples in relation to benign SHAP values of the same predicted class only, and (2) the standard approach, which compares adversarial samples to the entire benign SHAP value distribution across all classes, enabling a comparative analysis of detection effectiveness between class-specific and general anomaly detection methods.

\begin{table*}
\caption{CSAD versus standard approach. The standard approach compares adversarial samples to all benign samples regardless of class, while CSAD only compares adversarial samples to benign samples of the same predicted class. Higher values indicate better detection results.}
\centering
\resizebox{1\textwidth}{!}{
    \begin{tabular}
    {@{}p{0.07\textwidth}p{0.06\textwidth}ccccccc@{}}
    \toprule
    \multirow{3}{0.07\textwidth}{\textbf{Dataset}} &
    \multirow{3}{0.06\textwidth}{\textbf{Target Model}} &
    \multicolumn{2}{c}{\textbf{Query Attacks}} &
    \multicolumn{5}{c}{\textbf{Transferability Attacks}} \\ 
    \cmidrule(l){3-4} \cmidrule(l){5-9} 
      &&
      \textbf{boundary} &
      \textbf{HopSkipJump} &
      \textbf{random} &
      \textbf{GB imp.} &
      \textbf{LGB imp.} &
      \textbf{XGB imp.} &
      \textbf{RF imp.} \\ 
      ~ & ~ & \multicolumn{7}{c}{Standard Approach (\%) / Class-Specific Anomaly detection (CSAD) Approach (\%)} \\ \toprule
\textbf{Hate}    & GB  & 87.9 / \textbf{100.0} & 64.1 / \textbf{96.7} & 94.4 / \textbf{100.0} & 96.7 / 100.0 & 100.0 / 81.2 & 10.0 / \textbf{27.3} & 32.0 / 25.6 \\
                 & LGB & 8.6 / \textbf{58.2}   & 40.4 / \textbf{78.0} & 52.6 / \textbf{64.3}  & 50.0 / \textbf{79.7}  & 20.0 / \textbf{76.8}  & 22.0 / \textbf{69.9} & 28.6 / \textbf{40.5} \\
                 & XGB & 0.0 / \textbf{26.4}   & 6.6 / \textbf{67.6}  & 70.3 / \textbf{86.4}  & 56.0 / \textbf{92.1}  & 50.0 / \textbf{80.3}  & 4.5 / 22.0  & 13.6 / \textbf{24.8} \\
                 & RF  & 37.9 / \textbf{63.2}  & 23.5 / \textbf{30.8} & 62.6 / \textbf{100.0} & 74.0 / \textbf{100.0} & 75.0 / \textbf{75.7}  & 37.8 / \textbf{67.3} & 20.0 / \textbf{50.0} \\\midrule
\textbf{ICU}     & GB  & 8.4 / \textbf{29.0}   & 5.1 / \textbf{19.4}  & 0.3 / \textbf{7.3}    & 1.2 / \textbf{10.8}   & 0.6 / \textbf{1.8}    & 0.9 / \textbf{3.5}   & 1.2 / \textbf{3.5}   \\
                 & LGB & 1.0 / \textbf{4.1}    & 1.2 / \textbf{3.9}   & 3.8 / \textbf{22.3}   & 4.1 / \textbf{17.5}   & 0.9 / \textbf{2.4}    & 4.3 / \textbf{6.9}   & 4.5 / \textbf{5.6}   \\
                 & XGB & 2.6 / \textbf{3.5}    & 3.9 / \textbf{6.0}   & 1.0 / \textbf{6.2}    & 2.2 / \textbf{6.4}    & 0.0 / \textbf{4.2}    & 4.2 / \textbf{5.1}   & 3.3 / 3.3   \\
                 & RF  & 0.0 / \textbf{0.1}    & 0.0 / \textbf{0.2}   & 0.0 / \textbf{4.0}    & 0.0 / \textbf{0.5}    & 0.0 / 0.0    & 0.0 / 0.0   & 0.0 / \textbf{0.7}   \\\midrule
\textbf{VideoTQ} & GB  & 1.2 / \textbf{5.3}    & 0.7 / \textbf{9.4}   & 0.0 / \textbf{0.4}    & 0.8 / \textbf{1.0}    & 43.8 / \textbf{51.4}  & 90.2 / \textbf{90.3} & 64.3 / 64.3 \\
                 & LGB & 6.7 / \textbf{7.3}    & 0.0 / \textbf{1.5}   & 4.1 / 3.4    & 10.6 / 3.4   & 0.8 / 0.8    & 1.0 / 0.6   & 0.8 / 1.0   \\
                 & XGB & 47.0 / 38.9  & 21.4 / \textbf{25.2} & 0.0 / \textbf{0.4}    & 0.9 / \textbf{1.0}    & 21.9 / 21.6  & 30.9 / \textbf{31.5} & 17.2 / \textbf{17.8} \\
                 & RF  & 0.0 / \textbf{66.6}   & 0.0 / \textbf{87.6}  & 13.8 / 0.0   & 17.0 / 0.0   & 0.0 / 0.0    & 0.0 / \textbf{68.8}  & 0.0 / \textbf{79.0}  \\ \bottomrule
        \end{tabular} }

\label{tab:CSAD_vs_all}
\end{table*}

For each attack and target model, the table presents the percentage of samples with at least one anomalous feature based on SHAP values using both approaches. Higher values indicate greater success in detecting compromised adversarial samples. For each pair of results, the approach with the better detection rate is highlighted in bold.
The results demonstrate that the CSAD approach substantially improves the detection of adversarial samples in most cases, often achieving a detection rate that is over twice that of the standard approach. Particularly notable are cases involving the RF target model on the VideoTQ dataset, where the CSAD approach achieved detection rates of 66.6\% and 87.6\% for the boundary and HopSkipJump attacks respectively, compared to 0\% detection with the standard approach.

Statistical validation using McNemar exact tests (with Holm correction) on the importance-based anomaly rate confirmed that CSAD consistently and significantly outperforms the standard approach across all datasets ($p<0.001$). Effect sizes were uniformly large (Cohen’s $g \approx 1.0$), indicating a strong and systematic advantage of CSAD. Complementary analyses of average anomalous SHAP features per sample (Wilcoxon signed-rank, Holm-corrected) led to the same conclusion, with rank-biserial correlations also close to 1.0.
These results highlight that the superiority of CSAD is statistically significant and practically substantial, underscoring its robustness across datasets and attack scenarios.
Full test results are provided in Appendix~\ref{appendix:appendixC}.

These findings confirm that assessing anomalies in data samples with respect to their specific target class is essential, particularly when the data distribution across features is not uniform. By comparing adversarial samples to benign samples of the same class, the CSAD approach accounts for class-specific patterns, leading to improved anomaly detection. This approach is especially valuable for datasets in which the normal patterns in the features of the classes differ. Subtle changes that only matter for specific classes may go undetected when a global threshold for all classes is used.

\section{Discussion}\label{sec:discussion}

The results from our evaluation of the attacker's risk highlight the trade-offs between different attack strategies. The examined query-based attacks achieved near-perfect success rates (often exceeding $98\%$). However, this consistency comes at the cost of larger distortions and high computational overhead that is reflected in increased $L_0$ and $L_2$ distances. Transferability-based approaches offered efficiency advantages but struggled with success rates, particularly on complex datasets like the ICU dataset. The lower attack success rates for transferability-based attacks on the evaluated datasets highlight their limitations in adversarial scenarios.
Importantly, when defenders rely on query-based attacks as part of defense strategies (e.g., adversarial training), it is crucial to recognize that such attacks may not capture the same degree of feature coherence exhibited by transferability-based attacks, potentially leaving gaps in robustness.

On datasets with a limited number of editable features like ICU and VideoTQ, transferability attacks struggled more in terms of success rate than the more flexible Hate dataset which has a high number of editable features.
However, from a defender's perspective, transferability-based attacks present a unique challenge. Despite their lower overall success rate, when adversarial samples generated by these attacks do succeed in deceiving the target model, they may evade traditional detection mechanisms, underscoring the complexity of defending against these types of attacks.

Interestingly, while small $L_2$ distances are commonly regarded in the literature as indicative of imperceptibility, our findings suggest this may not hold for tabular data. In some cases, transferability-based attacks that achieved very small $L_2$ distances were still detected as anomalies by the AE. This finding calls into question the assumption that lower $L_2$ norm correlate with indistinguishability in tabular data domains, where imperceptibility may depend on other factors besides the $L_2$ distance.

Our comprehensive evaluation, which assessed both feature space coherency and the stability of model interpretation (via SHAP), showed that adversarial perturbations can affect feature values and model interpretations independently. Some attacks (like the HopSkipJump attack when targeting XGB and RF models on the Hate dataset) preserved normal feature distributions in the sample values while significantly altering the model's decision-making as reflected in the anomalous SHAP values. Others (like several transferability-based attacks on the VideoTQ dataset) showed the opposite pattern with anomalous sample values but relatively normal SHAP distributions. This finding suggests that comprehensive defense systems should monitor both the raw feature values and their impact on model decision-making, since focusing solely on one could cause sophisticated attacks targeting the other to go undetected.

The superiority of the CSAD approach seen in our evaluation demonstrates that context-specific thresholds substantially outperform global thresholds for identifying adversarial samples. This finding likely extends beyond our specific implementation to other defense mechanisms, where class-conditional modeling could improve detection accuracy.

Regarding anomaly detection models, IF consistently outperformed AE across datasets, particularly under boundary and random-based attacks. 
While AEs are designed to capture complex feature interactions, in our evaluation their effectiveness was limited and inconsistent, with meaningful detection primarily observed under the query-based scenarios. 
These results suggest that tree-based methods like IF may provide stronger baselines for tabular anomaly detection, while neural approaches such as AEs require further refinement to reliably capture adversarial perturbations.

The generalizability of our methodological contributions extends naturally to multi-class classification scenarios, despite our empirical evaluation focusing on binary classification datasets. Our regression-based perturbation technique preserves feature relationships regardless of the number of target classes, as it maintains coherence based on feature dependencies rather than classification boundaries. The CSAD approach inherently supports multi-class settings by comparing adversarial samples against benign samples from the same predicted class, regardless of whether they are binary or multi-class. Similarly, the SHAP-based evaluation metrics apply directly to multi-class models, as SHAP provides feature importance scores for each class prediction. The focus on binary classification enables thorough analysis of the proposed evaluation framework while maintaining computational feasibility for comprehensive comparison across multiple attack strategies and models; the use of established benchmark datasets facilitates direct comparison with prior tabular adversarial research.

Finally, we note an important limitation of our regression-based perturbation technique. In our evaluation, the method was applied to datasets where at least some feature dependencies could be identified in advance, either from domain knowledge or through direct analysis of the training data. In scenarios where such dependencies are mutual, non-trivial, or entirely latent, an additional preliminary stage is required to infer them before perturbation. This detection phase could employ statistical dependency measures, feature clustering, or model-based approaches to uncover hidden relationships. While our framework is theoretically compatible with integrating such a stage, its design and implementation remain an open avenue for future research.

\section{Conclusion}
\label{conc}

In this study we compared effectiveness of different black-box adversarial attacks strategies on ML models for tabular data. Our evaluation framework considers the attacker’s risk, effort, and the quality of adversarial samples, providing insights for both attack optimization and defense strategies.

The results obtained using diverse target models and datasets highlight distinct trade-offs: query-based attacks, while highly effective, often sacrifice coherence by causing significant distortions, and require substantial computational resources. Transferability-based attacks, although less successful in some cases, achieve greater coherence and consistency with fewer feature modifications, resulting in subtler impact on model decision-making. This is evidenced by lower anomaly detection rates and minimal changes in feature importance.

Designing adversarial attacks for tabular data thus requires balancing sample coherence and the attack success rate. This balance is especially important in the context of query-based attacks, particularly when these attacks are used by defenders as part of adversarial training.
To support the necessary balancing, we introduced novel metrics for assessing those aspects of adversarial samples and attacks in the tabular domain. These metrics independently capture anomalies in both feature space values and model interpretation (as reflected in SHAP values), representing the surface-level coherence of samples as well as their internal impact on the model's decision-making process.
Our evaluation framework, which combines traditional anomaly detection with SHAP-based analysis, demonstrates that perturbations can independently affect each of these aspects. This underscores the importance of monitoring both aspects when designing or defending against adversarial attacks.

Furthermore, Our CSAD approach, which employs context-specific thresholds, markedly improved detection rates over approaches that utilize a global threshold, emphasizing the importance of examining class-specific patterns in identifying adversarial samples. 

These findings and metrics lay the groundwork for future research, enabling the development of more balanced attack strategies that maximize effectiveness while preserving feature space coherency. This, in turn, can help defenders bolster their systems against such threats.

Future work could perform ablation studies on alternative preprocessing pipelines to better understand their impact on adversarial attack effectiveness and model robustness. Another promising direction is to extend the regression-based perturbation framework to scenarios where feature dependencies are mutual, non-trivial, or entirely latent. This would require incorporating a preliminary dependency discovery stage, using, for example, statistical measures, feature clustering, or model-based inference, to identify and model hidden relationships before perturbation. Investigating adversarial robustness in such settings, including cases where models incorporate built-in defenses or where attackers lack full knowledge of the target model’s features, could provide deeper insights and further refine both attack and defense strategies in the tabular domain.

\section{Data and Code Availability}

To promote reproducibility and facilitate future research, we provide open access to all research artifacts generated in this study. This includes: (1) the preprocessed benchmark datasets, (2) the implementation code for the evaluated attack methods and defense mechanisms, (3) the trained target models, and (4) the adversarial samples generated. All resources are available at \url{https://github.com/yael87/Framework-for-adversarial-in-tabular-domain}.

\section{Declarations}
The authors declare that they have no known competing financial interests or personal relationships that could have appeared to influence the work reported in this paper.

\begin{appendices}
\section{Pseudocode for Tabular Attacks}
\label{appendix:appendixA}

In this research, we applied two types of query-based attacks: the boundary and HopSkipJump attacks, along with five versions of transferability-based attacks, all adapted for the tabular domain and outlined below. All attacks include a generic constraint modification procedure to ensure that the generated adversarial samples remain valid in the tabular data domain.

\subsection{Tabular Constraint Modification}

The tabular domain requires specific constraint handling to ensure that adversarial samples maintain realistic and valid feature values. Our generic constraint modification procedure~\cref{pro:tabularmod} addresses four key aspects: (1) clipping feature values to their valid ranges (e.g., age cannot be negative), (2) imposing data type constraints (e.g., integer values for discrete features), (3) preserving immutable features that cannot be modified by an attacker (e.g., gender, age), and (4) correcting dependent features using pretrained regression models to maintain realistic relationships within the rest of the sample.

\setcounter{AlgoLine}{0}
\begin{algorithm*}
\DontPrintSemicolon

\caption{A generic procedure for applying constraints in tabular data domains: clipping each feature value to its valid range, imposing realistic values according to feature constraints (e.g., integer, positive, normalized), ensuring immutable features maintain their original values, and correcting dependent features based on corresponding regression models.}
\label{pro:tabularmod}
\KwIn{The original sample $x$, adversarial sample $x_{adv}$, set of immutable features $I$, the set of each feature constraints $Constraints$, and set of regression models $regression\_models$(if exist)}

\SetKwProg{myproc}{Procedure}{}{}{}
\nl \myproc{Tabular\_Modify(x, $x_{adv}$, I, Constraints, regression\_models)}{
     \nl Clip (min, max) \\
     \nl Impose realistic values (constraints) \\
     \nl Impose immutable features (I) \\
     \nl Correct dependent features (regression\_models) \\
     \nl \KwRet{adversarial}} 

\end{algorithm*}

\subsection{The Boundary Attack}
In the boundary attack presented in~\cref{alg:bondaryadv}, the attacker randomly selects a sample from the defined target class, regardless of its proximity to the original sample (lines 2-3).
The attacker then optimizes the modified sample's values using an orthogonal perturbation, which iteratively adjusts the feature values to minimize the model's confidence in the original class while ensuring that the predicted class is the target class (lines 4-6).
During the crafting process, tabular constraints are applied to the sample (see~\cref{pro:tabularmod}), projecting each feature value onto its closest legitimate value to ensure input validity (line 7).

\begin{algorithm*}
\SetAlgoLined
\DontPrintSemicolon
\caption{The boundary adversarial attack for tabular data}
\label{alg:bondaryadv}
\KwIn{The target model $M$, original sample $x$, original sample's true label $y$, set of immutable features $I$, set of each feature constraints $Constraints$, and the similarity threshold $\alpha$}
 
\nl $x_{adv} \gets x$\\
\nl \While{$i<max\_init\_steps \land M(x_{adv})=y$}
    {\nl $x_{adv}\gets$ random perturbation from distribution $x \sim \mathcal{N}(\mu, \sigma^2)$}
\nl \While{$i<max\_iteration\_steps \land M(x_{adv})=y \land$ Similarity($x,x_{adv})<\alpha$}
        {\nl \While{$j<max\_similarity\_steps$}
             {\nl $x_{adv} \gets x_{adv}$ + Orthogonal\_Perturbation($x, x_{adv}$)}
        
        \nl $x_{adv} \gets$ Tabular\_Modify(x, $x_{adv}$, I, Constraints, regression\_models)}

\nl \KwRet{$x_{adv}$}

\end{algorithm*}

\subsection{The HopSkipJump Attack}
In the HopSkipJump attack, which is presented in~\cref{alg:hopadv}, the attacker employs binary search and proceeds in the target class gradients' direction to find a sample from the defined target class with the closest proximity to the original sample (lines 5-10).
During the binary search, tabular constraints are applied to the sample (see~\cref{pro:tabularmod}), projecting each feature value onto its closest legitimate value to ensure input validity (line 11).

\setcounter{AlgoLine}{0}
\begin{algorithm*}
\SetAlgoLined
\DontPrintSemicolon
\caption{The HopSkipJump adversarial attack for tabular data}
\label{alg:hopadv}

\KwIn{The target model $M$, original sample $x$, original sample's true label $y$, set of immutable features $I$, set of each feature constraints $Constraints$, and the similarity threshold $\alpha$}

\nl $x_{adv} \gets x$ \\
\nl \While{$i< max\_init\_steps \land M(x_{adv})=y$}
        {\nl $x_{adv} \gets$ random perturbation from distribution $x \sim \mathcal{N}(\mu, \sigma^2$) \\
        \nl $x_{adv} \gets$ Binary\_Search(x, $x_{adv})$}

\nl \While{$i<max\_iteration\_steps \land M(x_{adv})=y \land$ Similarity(x,$x_{adv})<\alpha$}
        {\nl \While{$j<max\_similarity\_steps$}
            {\nl $x_{adv} \gets$ Binary\_Search(x, $x_{adv}$) \\
            \nl $update \gets$ Compute\_Update(x, $x_{adv}$) \\
            \nl $potential\_sample \gets$ run step size search ($x_{adv}, update$) until predicts satisfying \\
            \nl $x_{adv} \gets potential\_sample$}
        \nl $x_{adv} \gets$ Tabular\_Modify(x, $x_{adv}$, I, Constraints)}
\nl \KwRet{$x_{adv}$}

\end{algorithm*}

\subsection{The Transferability-Based Attacks}
The transferability-based attacks leverage surrogate models to generate adversarial samples that transfer to black-box target models. These attacks iteratively select and perturb features based on different selection strategies until successful misclassification is achieved or the maximum perturbation budget is exhausted.

In our evaluation, we implemented untargeted transferability-based gradient attacks based on the architecture in~\cite{mathov2022not,grolman2022hateversarial}, with the pseudocode presented in~\cref{alg:transferadv}.
In each iteration of these attacks, the attacker first selects features to pertrube, excluding immutable features that cannot be changed by the attacker (line 4).
The feature selection process (lines 10-14) differs across attack variants and follows one of five strategies:
\textit{1) Random-based selection:} selecting $k$ random features and the $n$ features that are most correlated with them.
\textit{2-5) Importance-based selection:} selecting the $k$ features most likely to impact the model's prediction and the $n$ features most correlated with them. Feature importance is determined using SHAP values computed from trained ML classifiers, with each attack variant using a different classifier (gradient boosting, XGBoost, LightGBM, or random forest) to compute these SHAP values ~\cite{grolman2022hateversarial}.

After selecting features to perturb, a perturbation vector $p$ is calculated for the selected features (line 5) by using an optimizer that minimizes the adversary’s objective function $L_{adv}$ which is defined in \cref{sec:exp_setup} of the paper.
Finally, tabular constraints are applied to the sample using~\cref{pro:tabularmod}, to ensure that the adversarial sample maintains valid feature values (line 7).

The feature selection process is detailed in the Select\_Features procedure, which maintains a set $F$ of already-selected features to avoid redundant selections across iterations.

\setcounter{AlgoLine}{0}
\begin{algorithm*}
\DontPrintSemicolon
\caption{An untargeted transferability-based adversarial attack for tabular data}
\label{alg:transferadv}

\KwIn{The surrogate model $M'$, original sample $x$, original sample's true label $y$, adversary's objective loss function $L_{adv}$, set of immutable features $I$, set of constraints for each feature  $Constraints$,  maximum allowed $L_{0}$-perturbation noise $\lambda$, number of most important features to perturb $k$, and number of features most correlated with the $k$ features chosen $n$}
\nl $x_{adv} \gets x$ \\
\nl $F \gets \emptyset$ \\
\nl \While{$M'(x_{adv})=y \land |F|<\lambda$}
    {\nl $F\gets F\cup$ Select\_Features($M', x_{adv}, I, F, k, n$) \\
    \nl $p \gets$ Compute\_Perturbation($M', L_{adv}, x_{adv}, F$) \\
    \nl $x_{adv} \gets x_{adv} + p$ \\
    \nl $x_{adv} \gets$ Tabular\_Modify($x, x_{adv}, I, Constraints$)}
\nl \KwRet{$x_{adv}$} \\

\SetKwProg{myproc}{Procedure}{}{}
  \myproc{Select\_Features($M', x_{adv}, I, F, k, n$)}{
      \nl \KwIn{Surrogate model $M'$, current adversarial sample $x_{adv}$, immutable features $I$, previously selected features $F$, number of features to select $k$, number of correlated features per selection $n$}
     \nl 1) \textbf{Random-Based Selection:} Select $k$ random features from available features not in $F \cup I$, then select $n$ most correlated features for each \\
     \nl 2) \textbf{GB Feature Importance-Based Selection:} Compute SHAP values using gradient boosting model, select top $k$ features not in $F \cup I$, then $n$ correlated features \\
     \nl 3) \textbf{XGB Feature Importance-Based Selection:} Compute SHAP values using XGBoost model, select top $k$ features not in $F \cup I$, then $n$ correlated features \\
     \nl 4) \textbf{LGB Feature Importance-Based Selection:} Compute SHAP values using LightGBM model, select top $k$ features not in $F \cup I$, then $n$ correlated features \\
     \nl \textbf{5) RF Feature Importance-Based Selection:} Compute SHAP values using random forest model, select top $k$ features not in $F \cup I$, then $n$ correlated features \\
     \nl \KwRet{Set of newly selected features}}

\end{algorithm*}

\section{Computational Complexity Analysis}
\label{appendix:appendixB}

This appendix provides a structured and detailed comparison of the computational complexity of the CSAD training strategy versus the traditional single-model approach.  
Two main factors affecting the overall complexity are as follows:

\begin{enumerate}
    \item \textbf{Class distribution} - whether the dataset is \emph{balanced} (each class contains roughly the same number of samples) or \emph{imbalanced} (some classes contain many more samples than others).
    \item \textbf{Algorithmic complexity} - characterized by the exponent $\alpha$ in the training complexity $O(m^\alpha)$ for $m$ samples. We distinguish between:
    \begin{itemize}
        \item \emph{Linear} complexity ($\alpha = 1$)
        \item \emph{Super-linear} complexity ($\alpha > 1$)
        \item \emph{Sub-linear} complexity ($\alpha < 1$)
    \end{itemize}
\end{enumerate}

The analysis performed covers all combinations of these factors.

\subsection{General Definitions}
Let:
\begin{itemize}
    \item $n$ - total number of training samples
    \item $k$ - number of classes
    \item $n_i$ - number of samples in class $i$ ($\sum_{i=1}^k n_i = n$)
    \item $\alpha > 0$ - complexity exponent of the training algorithm
\end{itemize}


\paragraph{Traditional approach:}  
A single model is trained on all $n$ samples:
\[
C_{\text{traditional}} = O(n^\alpha)
\]
This cost remains constant regardless of $k$.

\paragraph{CSAD approach:}  
A separate model is trained for each class $i$ on $n_i$ samples:
\[
C_{\text{CSAD}} = \sum_{i=1}^k O(n_i^\alpha)
\]
The exact cost depends on the class distribution $\{n_i\}$.

\subsection{Balanced Classes} \label{Balanced}

If the classes are balanced, i.e., $n_i \approx \frac{n}{k}$ for all $i$, the computational cost of CSAD is:
\[
\resizebox{\columnwidth}{!}{$\displaystyle
C_{\text{CSAD}} = k \cdot O\!\left( \left( \frac{n}{k} \right)^\alpha \right)
= O\left(k \cdot \frac{n^\alpha}{k^\alpha}\right)
= O\!\left( k^{1-\alpha} \cdot n^\alpha \right)$}
\]
Therefore, the complexity ratio relative to the traditional approach is:
\[
\frac{C_{\text{CSAD}}}{C_{\text{traditional}}} = k^{1-\alpha}
\]

The implications are as follows:
\begin{itemize}
    \item For $\alpha > 1$, CSAD provides computational savings since $k^{1-\alpha} < 1$. The speedup factor (defined as the inverse of the complexity ratio) is $k^{\alpha - 1}$.
    \item For $\alpha = 1$, both approaches exhibit the same computational complexity $O(n)$.
    \item For $\alpha < 1$, Relative to the traditional method, CSAD incurs a higher computational cost.
\end{itemize}

\begin{figure*}[!htbp]
    \centering
    \includegraphics[width=0.65\linewidth]{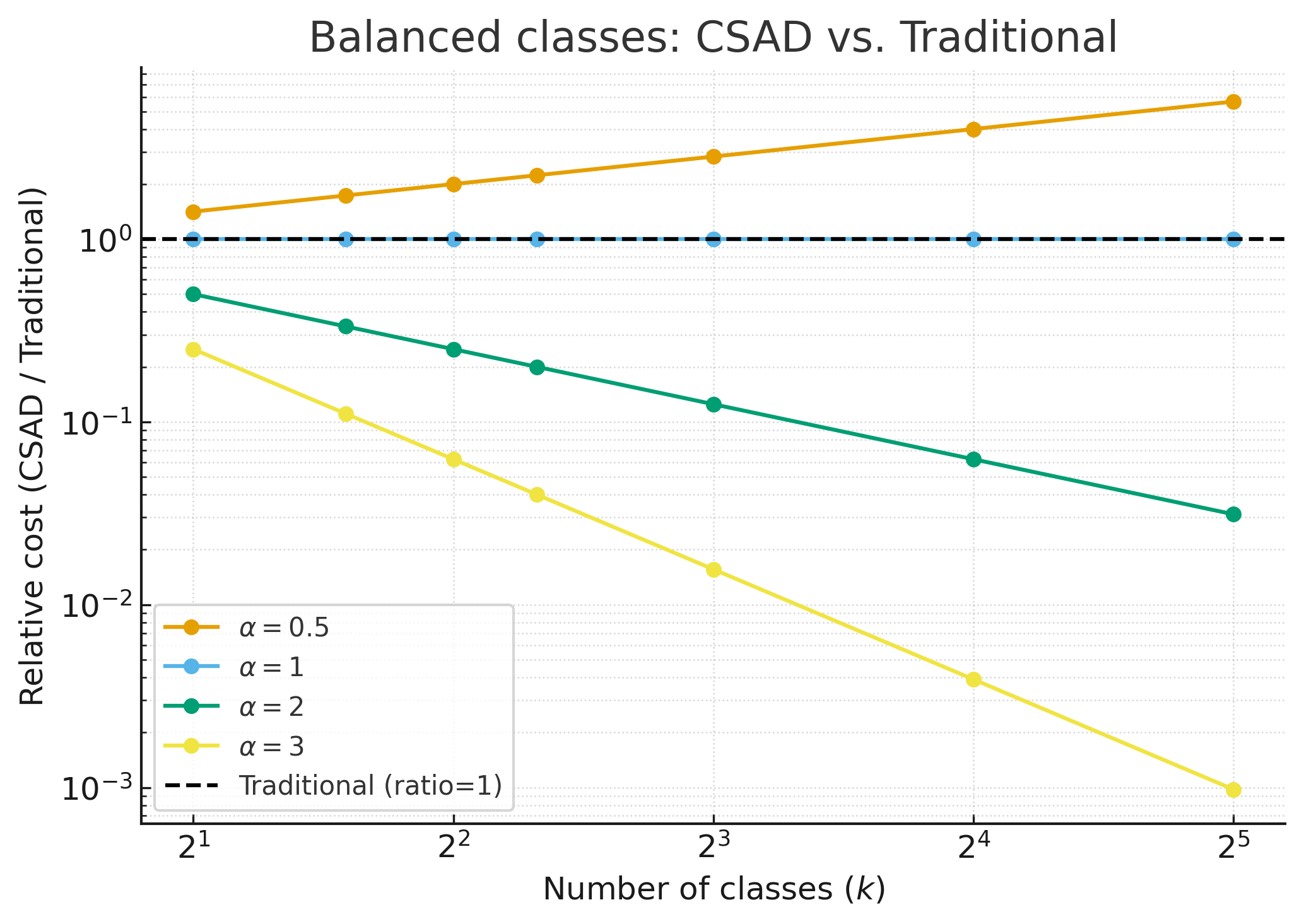}
    \caption{Training complexity comparison between CSAD and the traditional approach for balanced classes. The curves show the theoretical relative training cost as a function of the number of classes $k$, under the assumption that all classes contain the same number of samples. For $\alpha < 1$, CSAD becomes more expensive as the number of classes increases. At $\alpha = 1$, both approaches exhibit the same complexity. For $\alpha > 1$, CSAD achieves increasing computational savings with larger $k$, with steeper reductions for larger $\alpha$. The traditional approach maintains a constant cost regardless of $k$, as it trains a single model on the full dataset.}
    \label{fig:balanced_traditional_vs_CSAD}
\end{figure*}

The relative training cost for balanced classes across different values of $\alpha$ is presented in Figure~\ref{fig:balanced_traditional_vs_CSAD}. As shown in the figure, for $\alpha > 1$, CSAD demonstrates significant computational advantages that increase with the number of classes, while for $\alpha < 1$, the traditional approach remains more efficient.

Table~\ref{tab:complexity_summary} summarizes the relative computational complexity and savings for the balanced case.

\begin{table*}
\centering
\caption{Summary of computational complexity for balanced datasets.}
\begin{tabular}{lccc}
\toprule
\textbf{$\boldsymbol{\alpha}$ value} & \textbf{Traditional Cost} & \textbf{CSAD Cost (balanced)} & \textbf{Relative Savings} \\
\midrule
$\alpha > 1$ & $O(n^\alpha)$ & $O(k^{1-\alpha} \cdot n^\alpha)$ & Significant \\
$\alpha = 1$ & $O(n)$ & $O(n)$ & None \\
$\alpha < 1$ & $O(n^\alpha)$ & $O(k^{1-\alpha} \cdot n^\alpha)$ & Negative (higher cost) \\
\bottomrule
\end{tabular}
\label{tab:complexity_summary}
\end{table*}

\subsection{Unbalanced Classes} \label{Unbalanced}

In unbalanced scenarios, the ratio becomes:
\[
\frac{C_{\text{CSAD}}}{C_{\text{traditional}}} = \frac{\sum_{i=1}^k n_i^\alpha}{n^\alpha}
\]
For $\alpha > 1$, the convexity of $x^\alpha$ and Jensen's inequality~\cite{cover2006elements} imply:
\[
\sum_{i=1}^k n_i^\alpha \geq k\left( \frac{n}{k} \right)^\alpha
\]
Therefore, the balanced case minimizes $C_{\text{CSAD}}$ for fixed $n$ and $k$.

In the extreme case, where one class dominates ($n_{\max} \approx n$):
\[
C_{\text{CSAD}} \approx O(n^\alpha) = C_{\text{traditional}}
\]
Therefore, the computational cost improvement is minimal or negligible, and the cost of CSAD approaches that of the traditional single-model approach.

\begin{figure*}[!htbp]
    \centering
    \includegraphics[width=0.65\linewidth]{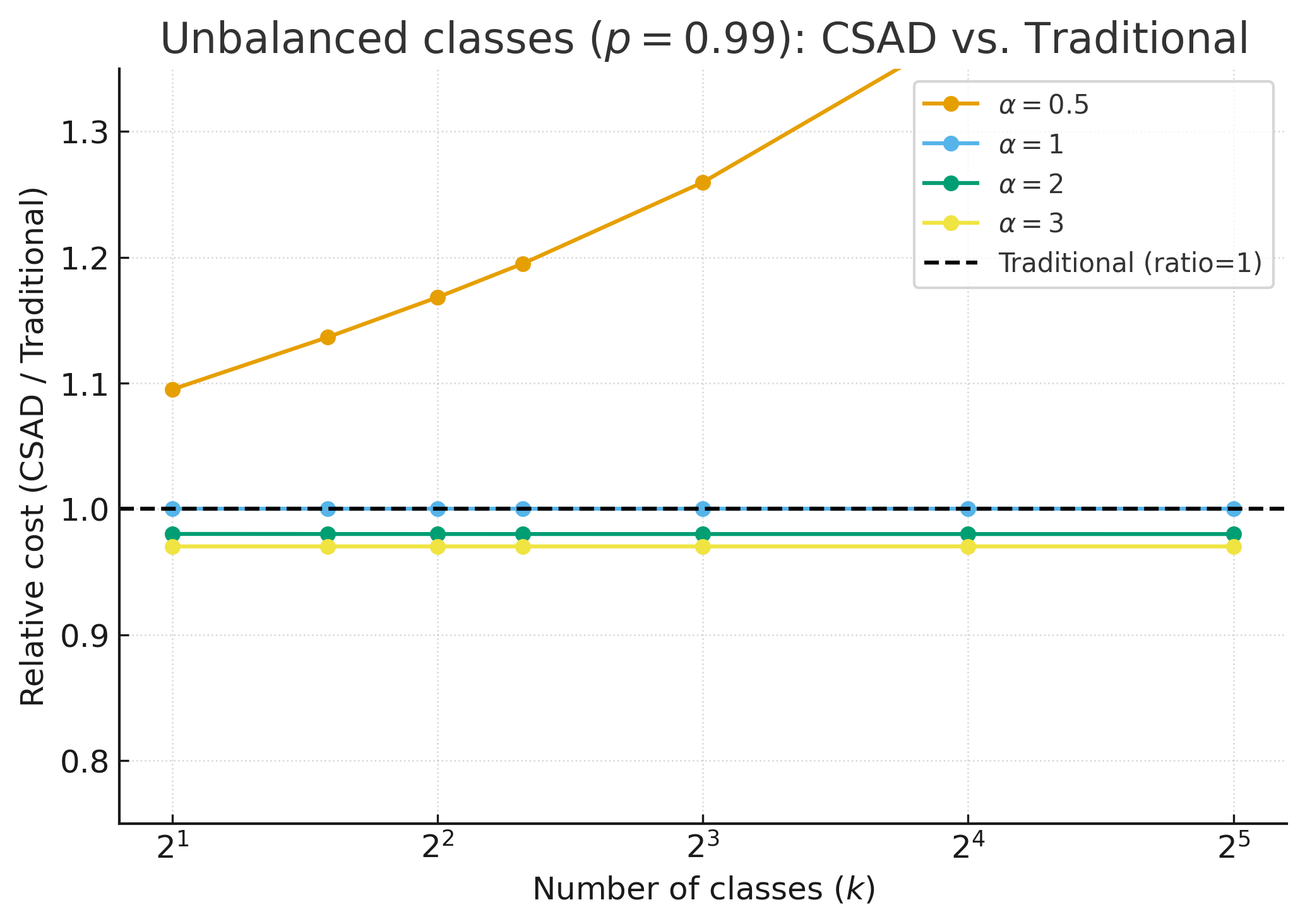}
    \caption{Training complexity comparison between CSAD and traditional approaches for unbalanced classes. The curves illustrate the worst case where a dominant class contains most of the samples (99\% in this example). For $\alpha < 1$, CSAD incurs higher cost as the number of classes increases. For $\alpha = 1$, both approaches exhibit equivalent complexity. For $\alpha > 1$, the computational advantage of CSAD largely vanishes due to the dominant class, making its cost comparable to the traditional approach.}
    \label{fig:unbalanced_traditional_vs_CSAD}
\end{figure*}

The relative training cost for unbalanced classes across different values of $\alpha$ is presented in Figure~\ref{fig:unbalanced_traditional_vs_CSAD}. As shown in the figure, for a dominant class that contains most of the samples (99\% in this example), CSAD's computational advantage diminishes, because the largest class determines the training time. In such scenarios, CSAD and the traditional approach exhibit equivalent computational complexity.

\begin{figure*}[!htbp]
    \centering
    \includegraphics[width=0.65\textwidth]{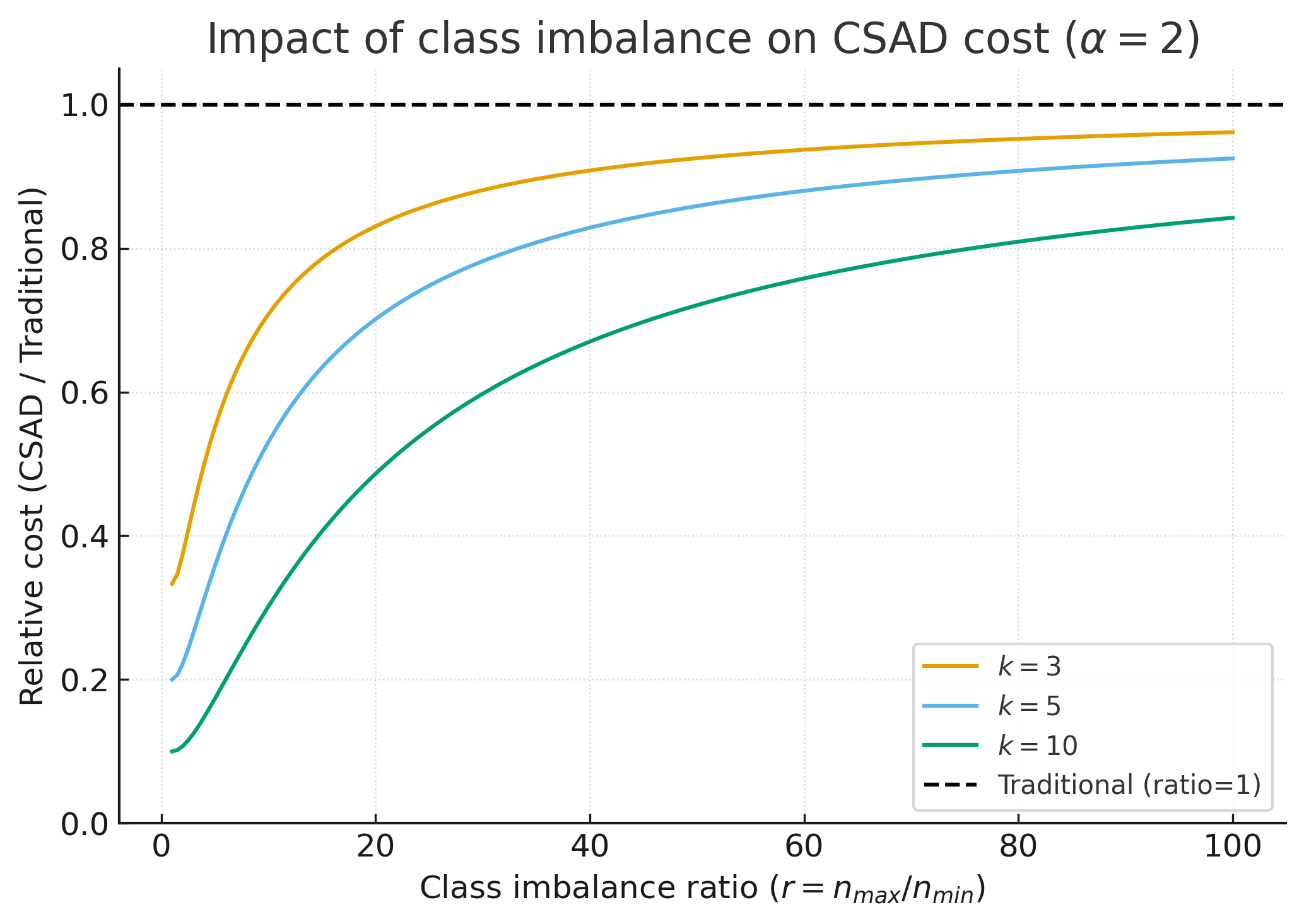}
    \caption{The impact of class imbalance on the relative benefit of CSAD over the traditional approach, for $\alpha > 1$ and a fixed number of classes $k$. When classes are nearly balanced, CSAD achieves substantial computational savings over the traditional method. As the imbalance ratio (largest/smallest class) increases, this advantage diminishes, reaching parity when a single class dominates.}
    \label{fig:unbalanced_class_size_ratio}
\end{figure*}

Figure~\ref{fig:unbalanced_class_size_ratio} illustrates the relative advantage of CSAD over the traditional single-model approach when $\alpha>1$; diminishes progressively as class imbalance increases, reaching parity in the extreme case where a single class dominates.

\subsection{Linear vs. Super-linear Cases}

For \textbf{linear} algorithms ($\alpha = 1$), such as IF~\cite{liu2008isolation} and AEs~\cite{sakurada2014anomaly}, the cost remains:
\[
C_{\text{traditional}} = O(n), \quad C_{\text{CSAD}} = O(n)
\]
regardless of the class distribution. Partitioning the data into subsets has no impact on the total cost, since the cost grows exactly in proportion to the number of processed samples.

For \textbf{super-linear} algorithms ($\alpha > 1$), such as One-Class SVM with RBF kernels~\cite{scholkopf2001estimating}, the savings in balanced datasets can be substantial, as discussed in~\cref{Balanced}.  
In unbalanced datasets, however, the advantage diminishes as shown in~\cref{Unbalanced}.

\subsection{Numerical Examples}
  
Table~\ref{tab:complexity_examples} presents specific examples for both balanced and unbalanced datasets, including an extreme case where one class contains 97\% of the samples.
 
\begin{table*}
\centering
\begin{tabular}{lccccc}
\toprule
\textbf{Case} & $\boldsymbol{n}$ & $\boldsymbol{k}$ & $\boldsymbol{\alpha}$ & $\boldsymbol{\frac{C_{\text{CSAD}}}{C_{\text{traditional}}}}$ & \textbf{Savings} \\
\midrule
Balanced, super-linear & $100{,}000$ & $4$ & $2$   & $0.25$  & 75\% \\
Balanced, near-linear & $100{,}000$ & $4$ & $1.1$ & $0.87$  & 13\% \\
Balanced, linear      & $100{,}000$ & $4$ & $1$   & $1.00$  & 0\% \\
Unbalanced (97\%|1\%|1\%|1\%), $\alpha=2$ & $100{,}000$ & $4$ & $2$ & $\approx 0.94$ & 6\% \\
Unbalanced (97\%|1\%|1\%|1\%), $\alpha=1.1$ & $100{,}000$ & $4$ & $1.1$ & $\approx 0.99$ & 1\% \\
\bottomrule
\end{tabular}
\caption{Training complexity ratio $\frac{C_{\text{CSAD}}}{C_{\text{traditional}}}$ for balanced and unbalanced scenarios, including an extreme case where one out of four classes contains 97\% of the data. Ratios $< 1$ indicate computational savings achieved by CSAD.}

\label{tab:complexity_examples}
\end{table*}

\subsection{Summary}

\begin{itemize}
    \item \textbf{Balanced, $\boldsymbol{\alpha > 1}$:} CSAD achieves a speedup factor of $k^{\alpha - 1}$ compared to the traditional approach.
    \item \textbf{Unbalanced:} The computational savings of CSAD decreases as class imbalance grows; in the extreme case, CSAD's cost approaches that of the traditional approach.
    \item \textbf{$\boldsymbol{\alpha = 1}$:} CSAD and the traditional approach exhibit equivalent computational complexity, regardless of the class distribution.
    \item \textbf{$\boldsymbol{\alpha < 1}$:} CSAD incurs higher computational cost than the traditional approach.
\end{itemize}

\section{Statistical Validation of Adversarial Comparisons}
\label{appendix:appendixC}

This appendix reports the full statistical validation results for all datasets.
The goal is to provide a rigorous comparison between query-based attacks 
(boundary, HopSkipJump) and transferability attacks (GB imp., LGB imp., 
XGB imp., RF imp., Random) across multiple evaluation metrics.

\textbf{Statistical tests.}
For each metric we select the statistical test based on the nature of the data and 
whether the compared samples are paired (same adversarial examples) or independent 
(different sets of successful adversarial examples):

\begin{itemize}
    \item \textbf{Continuous perturbation metrics ($L_0$, $L_2$, average anomalous SHAP features):} 
    pairwise \emph{Mann--Whitney $U$} tests (independent samples, non-parametric), 
    with Cliff's $\delta$ as effect size.
    \item \textbf{Proportion-type metrics (overall success rate, anomaly detection rates with IF or AE, 
    importance-based anomaly rates):} 
    two-sample \emph{$z$-tests for proportions} (independent samples), 
    with Cohen's $h$ as effect size.
    \item \textbf{Class-specific anomaly detection (CSAD) vs.\ standard approach:} 
    since both measures are computed for the same adversarial examples, 
    we use \emph{paired tests}: McNemar’s exact test for binary anomaly flags, 
    and Wilcoxon signed-rank tests for anomaly counts. 
    Effect sizes are reported as Cohen’s $g$ and rank-biserial correlation $r_{rb}$.
\end{itemize}

All $p$-values are adjusted with Holm’s sequential Bonferroni correction to control 
the family-wise error rate.

\textbf{Table reporting and interpretation.}
Each table reports adjusted $p$-values, significance (\checkmark{} if significant, 
``ns'' otherwise), and the effect size category (S/M/L). 
Row groups show:
(i) boundary vs.\ transferability attacks,
(ii) HopSkipJump vs.\ transferability attacks,
(iii) boundary vs.\ HopSkipJump attacks,
(iv) Internal heterogeneity inside transferability attacks (\checkmark if significant differences exist, ns otherwise).


\subsection{Hate Dataset}

\subsubsection{Metric: Overall Success Rate}
Table~\ref{tab:hate_overall_sr} reports the results of pairwise tests (Proportions z-test) with Holm correction for the Overall Success Rate across different attacks. 
The results show that both \emph{boundary vs. transferability attacks} and \emph{HopSkipJump vs. transferability attacks} comparisons are highly significant, with large effect sizes (L) for Random, GB imp., and LGB imp., and medium effect sizes (M) for XGB imp. and RF imp. 
The \emph{boundary vs. HopSkipJump} comparison is also significant, but the effect size is small (S). 
In addition, the omnibus test indicates no significant internal heterogeneity within the transferability attacks group.

\begin{table*}
\centering
\resizebox{\linewidth}{!}{%
\begin{tabular}{lccccc}
\toprule
Comparison & Random & GB imp. & LGB imp. & XGB imp. & RF imp. \\
\midrule
boundary vs. transferability attacks & $1.19\text{e-264}$/\checkmark/L & $1.46\text{e-76}$/\checkmark/L & $2.11\text{e-148}$/\checkmark/L & $1.92\text{e-58}$/\checkmark/M & $1.75\text{e-39}$/\checkmark/M \\
HopSkipJump vs. transferability attacks & $1.19\text{e-264}$/\checkmark/L & $1.46\text{e-76}$/\checkmark/L & $2.11\text{e-148}$/\checkmark/L & $1.92\text{e-58}$/\checkmark/M & $1.75\text{e-39}$/\checkmark/M \\
boundary vs. HopSkipJump & \multicolumn{5}{c}{NaN/\checkmark/S} \\
Transferability internal heterogeneity & \multicolumn{5}{c}{ns (none)} \\
\bottomrule
\end{tabular}}
\caption{\textsc{Hate}: Overall attack success rate. Each cell reports $p$-value / significance / effect size.}
\label{tab:hate_overall_sr}
\end{table*}

\subsubsection{Metric: L0 Distance}
Table~\ref{tab:hate_l0} reports the results of pairwise tests (Mann–Whitney U) with Holm correction for the $L_0$ distance across different attacks. 
The results show that all comparisons are highly significant with large effect sizes (L). 
Moreover, unlike the success rate analysis, the omnibus test reveals significant internal heterogeneity within the transferability attacks group, indicating that $L_0$ distances vary substantially across different transferability attacks.

\begin{table*}
\centering
\resizebox{\linewidth}{!}{%
\begin{tabular}{lccccc}
\toprule
Comparison & Random & GB imp. & LGB imp. & XGB imp. & RF imp. \\
\midrule
boundary vs. transferability attacks & $6.80\text{e-42}$/\checkmark/L & $7.41\text{e-191}$/\checkmark/L & $5.46\text{e-131}$/\checkmark/L & $1.51\text{e-208}$/\checkmark/L & $1.76\text{e-216}$/\checkmark/L \\
HopSkipJump vs. transferability attacks & $2.75\text{e-41}$/\checkmark/L & $6.57\text{e-190}$/\checkmark/L & $5.50\text{e-130}$/\checkmark/L & $1.25\text{e-207}$/\checkmark/L & $1.08\text{e-215}$/\checkmark/L \\
boundary vs. HopSkipJump & \multicolumn{5}{c}{$1.69\text{e-120}$/\checkmark/L} \\
Transferability internal heterogeneity & \multicolumn{5}{c}{\checkmark (heterogeneity)} \\
\bottomrule
\end{tabular}}
\caption{\textsc{Hate}: $L_0$ distance of adversarial examples. Each cell reports $p$-value / significance / effect size.}
\label{tab:hate_l0}
\end{table*}

\subsubsection{Metric: L2 Distance}
Table~\ref{tab:hate_l2} reports the results of pairwise tests (Mann–Whitney U) with Holm correction for the $L_2$ distance across different attacks. 
The results show that all comparisons are highly significant with large effect sizes (L). 
Both \emph{boundary vs. transferability attacks} and \emph{HopSkipJump vs. transferability attacks} yield extremely small $p$-values across all Transferability variants. 
The \emph{boundary vs. HopSkipJump} comparison is also significant with a large effect size. 
In addition, the omnibus test indicates significant internal heterogeneity within the transferability attacks group, suggesting that $L_2$ distances differ substantially across different transferability attacks.

\begin{table*}
\centering
\resizebox{\linewidth}{!}{%
\begin{tabular}{lccccc}
\toprule
Comparison & Random & GB imp. & LGB imp. & XGB imp. & RF imp. \\
\midrule
boundary vs. transferability attacks & $3.11\text{e-40}$/\checkmark/L & $2.72\text{e-181}$/\checkmark/L & $1.82\text{e-127}$/\checkmark/L & $7.10\text{e-195}$/\checkmark/L & $4.09\text{e-209}$/\checkmark/L \\
HopSkipJump vs. transferability attacks & $1.44\text{e-08}$/\checkmark/L & $4.03\text{e-33}$/\checkmark/L & $1.78\text{e-23}$/\checkmark/L & $1.81\text{e-35}$/\checkmark/L & $3.11\text{e-40}$/\checkmark/L \\
boundary vs. HopSkipJump & \multicolumn{5}{c}{$5.54\text{e-238}$/\checkmark/L} \\
Transferability internal heterogeneity & \multicolumn{5}{c}{\checkmark (heterogeneity)} \\
\bottomrule
\end{tabular}}
\caption{\textsc{Hate}: $L_2$ distance of adversarial examples. Each cell reports $p$-value / significance / effect size.}
\label{tab:hate_l2}
\end{table*}

\subsubsection{Metric: Feature-Space Anomaly Detection Rate (IF-based)}
Table~\ref{tab:hate_if} reports the results of pairwise tests (Proportions z-test) with Holm correction for the feature-space anomaly detection rate using the Isolation Forest (IF) detector. 
The results show that comparisons of \emph{boundary vs. transferability attacks} are all highly significant with large effect sizes (L). 
\emph{HopSkipJump vs. transferability attacks} comparisons are also significant, though the effect sizes are consistently medium (M). 
The \emph{boundary vs. HopSkipJump} comparison is highly significant with a large effect size. 
The omnibus test further indicates significant internal heterogeneity within the transferability attacks group, suggesting that IF-based anomaly detection rates differ substantially across transferability attacks.

\begin{table*}
\centering
\resizebox{\linewidth}{!}{%
\begin{tabular}{lccccc}
\toprule
Comparison & Random & GB imp. & LGB imp. & XGB imp. & RF imp. \\
\midrule
boundary vs. transferability attacks & $1.44\text{e-142}$/\checkmark/L & $6.63\text{e-223}$/\checkmark/L & $1.04\text{e-164}$/\checkmark/L & $2.24\text{e-223}$/\checkmark/L & $3.44\text{e-210}$/\checkmark/L \\
HopSkipJump vs. transferability attacks & $3.53\text{e-02}$/\checkmark/M & $1.24\text{e-21}$/\checkmark/M & $6.52\text{e-05}$/\checkmark/M & $2.40\text{e-19}$/\checkmark/M & $1.48\text{e-11}$/\checkmark/M \\
boundary vs. HopSkipJump & \multicolumn{5}{c}{$1.75\text{e-157}$/\checkmark/L} \\
Transferability internal heterogeneity & \multicolumn{5}{c}{\checkmark (heterogeneity)} \\
\bottomrule
\end{tabular}}
\caption{\textsc{Hate}: Feature-space anomaly detection rate (Isolation Forest). Each cell reports $p$-value / significance / effect size.}
\label{tab:hate_if}
\end{table*}

\subsubsection{Metric: Feature-Space Anomaly Detection Rate (AE-based)}
Table~\ref{tab:hate_ae} reports the results of pairwise tests (Proportions z-test) with Holm correction for the feature-space anomaly detection rate using the Autoencoder (AE) detector. 
The results show that \emph{boundary vs. transferability attacks} comparisons are all significant with medium (M) effect sizes. 
\emph{HopSkipJump vs. transferability attacks} comparisons are also significant, mostly with medium (M) effect sizes, except for XGB imp.\ where the effect size is small (S). 
The \emph{boundary vs. HopSkipJump} comparison is not significant. 
The omnibus test indicates significant internal heterogeneity within the transferability attacks group, suggesting that AE-based anomaly detection rates differ across transferability attacks.

\begin{table*}
\centering
\resizebox{\linewidth}{!}{%
\begin{tabular}{lccccc}
\toprule
Comparison & Random & GB imp. & LGB imp. & XGB imp. & RF imp. \\
\midrule
boundary vs. transferability attacks & $9.52\text{e-05}$/\checkmark/M & $5.00\text{e-10}$/\checkmark/M & $2.58\text{e-09}$/\checkmark/M & $1.24\text{e-08}$/\checkmark/M & $2.13\text{e-26}$/\checkmark/M \\
HopSkipJump vs. transferability attacks & $6.13\text{e-04}$/\checkmark/M & $2.09\text{e-07}$/\checkmark/M & $3.22\text{e-07}$/\checkmark/M & $3.93\text{e-06}$/\checkmark/S & $8.97\text{e-22}$/\checkmark/M \\
boundary vs. HopSkipJump & \multicolumn{5}{c}{$1.0$/--/S} \\
Transferability internal heterogeneity & \multicolumn{5}{c}{\checkmark (heterogeneity)} \\
\bottomrule
\end{tabular}}
\caption{\textsc{Hate}: Feature-space anomaly detection rate (Autoencoder). Each cell reports $p$-value / significance / effect size.}
\label{tab:hate_ae}
\end{table*}

\subsubsection{Metric: Importance-Based Anomaly Detection Rate (SHAP-based)}
Table~\ref{tab:hate_shap_has} reports the results of pairwise tests (Proportions z-test) with Holm correction for the SHAP-based importance anomaly rate. 
The results show that \emph{boundary vs. transferability attacks} comparisons are all highly significant, with large effect sizes (L) except for LGB imp.\ (medium, M). 
\emph{HopSkipJump vs. transferability attacks} comparisons are significant as well, mostly with medium (M) effects, and a large (L) effect for XGB imp. 
The \emph{boundary vs. HopSkipJump} comparison is significant but with a small (S) effect size. 
In addition, the omnibus test indicates significant internal heterogeneity within the transferability attacks group, suggesting that SHAP-based anomaly rates differ across transferability attacks.

\begin{table*}
\centering
\resizebox{\linewidth}{!}{%
\begin{tabular}{lccccc}
\toprule
Comparison & Random & GB imp. & LGB imp. & XGB imp. & RF imp. \\
\midrule
boundary vs. transferability attacks & $4.25\text{e-27}$/\checkmark/L & $1.72\text{e-71}$/\checkmark/L & $4.57\text{e-32}$/\checkmark/M & $1.56\text{e-75}$/\checkmark/L & $1.38\text{e-67}$/\checkmark/L \\
HopSkipJump vs. transferability attacks & $8.38\text{e-07}$/\checkmark/M & $2.75\text{e-30}$/\checkmark/M & $5.41\text{e-07}$/\checkmark/M & $4.64\text{e-33}$/\checkmark/L & $2.74\text{e-27}$/\checkmark/M \\
boundary vs. HopSkipJump & \multicolumn{5}{c}{$3.18\text{e-13}$/\checkmark/S} \\
Transferability internal heterogeneity & \multicolumn{5}{c}{\checkmark (heterogeneity)} \\
\bottomrule
\end{tabular}}
\caption{\textsc{Hate}: Importance-based anomaly detection rate (SHAP-based). Each cell reports $p$-value / significance / effect size.}
\label{tab:hate_shap_has}
\end{table*}

\subsubsection{Metric: Average Anomalous SHAP Features per Sample}
Table~\ref{tab:hate_shap_mean} reports the results of pairwise tests (Mann–Whitney U) with Holm correction for the average number of anomalous SHAP features per sample. 
The results show that \emph{boundary vs. transferability attacks} comparisons are all highly significant with large effect sizes (L). 
\emph{HopSkipJump vs. transferability attacks} comparisons are also significant, with large (L) effects in most cases, except for LGB imp.\ where the effect size is medium (M). 
The \emph{boundary vs. HopSkipJump} comparison is highly significant with a large effect size (L). 
The omnibus test further indicates significant internal heterogeneity within the transferability attacks group, suggesting that the distribution of anomalous SHAP counts varies considerably across transferability attacks.

\begin{table*}
\centering
\resizebox{\linewidth}{!}{%
\begin{tabular}{lccccc}
\toprule
Comparison & Random & GB imp. & LGB imp. & XGB imp. & RF imp. \\
\midrule
boundary vs. transferability attacks & $7.73\text{e-20}$/\checkmark/L & $8.33\text{e-113}$/\checkmark/L & $1.18\text{e-60}$/\checkmark/L & $4.26\text{e-120}$/\checkmark/L & $9.30\text{e-126}$/\checkmark/L \\
HopSkipJump vs. transferability attacks & $9.93\text{e-06}$/\checkmark/L & $2.67\text{e-42}$/\checkmark/L & $1.97\text{e-11}$/\checkmark/M & $2.10\text{e-45}$/\checkmark/L & $2.40\text{e-43}$/\checkmark/L \\
boundary vs. HopSkipJump & \multicolumn{5}{c}{$1.47\text{e-58}$/\checkmark/L} \\
Transferability internal heterogeneity & \multicolumn{5}{c}{\checkmark (heterogeneity)} \\
\bottomrule
\end{tabular}}
\caption{\textsc{Hate}: Average number of anomalous SHAP features per sample. Each cell reports $p$-value / significance / effect size.}
\label{tab:hate_shap_mean}
\end{table*}

\subsubsection{Comparison Between CSAD and Standard Approaches}
Table~\ref{tab:csad_std_binary} reports the results of statistical tests comparing the CSAD and Standard approaches on the \emph{Importance-Based Anomaly Detection Rate}. 
The paired McNemar exact tests (Holm-adjusted) indicate that CSAD significantly outperforms the Standard approach across nearly all attack–target combinations, with consistently large effect sizes (Cohen’s $g$). 
This confirms that class-specific anomaly detection substantially improves the binary anomaly rate.

\begin{table*}
\centering
\begin{tabular}{lcccc}
\toprule
 & \multicolumn{4}{c}{Target Model} \\
\cmidrule(lr){2-5}
\textbf{Query Attacks} & GB & LGB & XGB & RF \\
\midrule
boundary & $3.91\text{e-2}$/\checkmark/L & $2.17\text{e-17}$/\checkmark/L & $1.36\text{e-11}$/\checkmark/L & $7.57\text{e-10}$/\checkmark/L \\
HopSkipJump & $2.16\text{e-12}$/\checkmark/L & $4.36\text{e-26}$/\checkmark/L & $2.70\text{e-33}$/\checkmark/L & $6.25\text{e-13}$/\checkmark/L \\
\midrule
\textbf{Transferability Attacks} & GB & LGB & XGB & RF \\
\midrule
random & $1.00\text{e0}$/--/L & $1.25\text{e-1}$/--/L & $3.75\text{e-1}$/--/L & $1.00\text{e0}$/--/L \\
GB imp. & $1.19\text{e-12}$/\checkmark/L & $1.02\text{e-14}$/\checkmark/L & $7.57\text{e-10}$/\checkmark/L & $9.77\text{e-4}$/\checkmark/L \\
LGB imp. & $2.10\text{e-9}$/\checkmark/L & $8.82\text{e-12}$/\checkmark/L & $7.45\text{e-9}$/\checkmark/L & $1.43\text{e-6}$/\checkmark/L \\
XGB imp. & $1.18\text{e-11}$/\checkmark/L & $1.49\text{e-13}$/\checkmark/L & $7.28\text{e-12}$/\checkmark/L & $4.77\text{e-6}$/\checkmark/L \\
RF imp. & $8.15\text{e-11}$/\checkmark/L & $1.25\text{e-14}$/\checkmark/L & $1.93\text{e-12}$/\checkmark/L & $2.38\text{e-13}$/\checkmark/L \\
\bottomrule
\end{tabular}
\caption{\textsc{Hate}: CSAD vs. Standard on importance-based anomaly detection rate (exact McNemar, Holm-adjusted). Each cell shows $p$-value / significance / effect size (Cohen’s $g$).}
\label{tab:csad_std_binary}
\end{table*}

Table~\ref{tab:csad_std_counts} reports complementary results for the \emph{Average Anomalous SHAP Features per Sample}. 
Here, Wilcoxon signed-rank tests (Holm-adjusted) confirm the same qualitative conclusion: CSAD consistently identifies more anomalous SHAP features than Standard, with large effect sizes (rank-biserial $r$). 
The random transferability attack shows mixed significance, but when significant the effect sizes remain large.

\begin{table*}
\centering
\begin{tabular}{lcccc}
\toprule
 & \multicolumn{4}{c}{Target Model} \\
\cmidrule(lr){2-5}
\textbf{Query Attacks} & GB & LGB & XGB & RF \\
\midrule
boundary & $4.28\text{e-30}$/\checkmark/L & $3.49\text{e-22}$/\checkmark/L & $9.50\text{e-27}$/\checkmark/L & $1.36\text{e-22}$/\checkmark/L \\
HopSkipJump & $3.44\text{e-23}$/\checkmark/L & $8.11\text{e-23}$/\checkmark/L & $1.69\text{e-30}$/\checkmark/L & $5.37\text{e-25}$/\checkmark/L \\
\midrule
\textbf{Transferability Attacks} & GB & LGB & XGB & RF \\
\midrule
random & $5.92\text{e-1}$/--/L & $3.40\text{e-4}$/\checkmark/L & $3.40\text{e-4}$/\checkmark/L & $7.61\text{e-1}$/--/L \\
GB imp. & $3.37\text{e-12}$/\checkmark/L & $1.72\text{e-10}$/\checkmark/L & $3.14\text{e-13}$/\checkmark/L & $1.22\text{e-7}$/\checkmark/L \\
LGB imp. & $1.55\text{e-10}$/\checkmark/L & $1.09\text{e-9}$/\checkmark/L & $1.62\text{e-11}$/\checkmark/L & $2.22\text{e-7}$/\checkmark/L \\
XGB imp. & $2.29\text{e-11}$/\checkmark/L & $1.63\text{e-10}$/\checkmark/L & $1.75\text{e-12}$/\checkmark/L & $6.11\text{e-6}$/\checkmark/L \\
RF imp. & $4.18\text{e-12}$/\checkmark/L & $2.48\text{e-10}$/\checkmark/L & $3.62\text{e-12}$/\checkmark/L & $1.06\text{e-10}$/\checkmark/L \\
\bottomrule
\end{tabular}
\caption{\textsc{Hate}: CSAD vs. Standard on average anomalous SHAP features per sample (Wilcoxon signed-rank, Holm-adjusted). Each cell shows $p$-value / significance / effect size (rank-biserial $r$).}
\label{tab:csad_std_counts}
\end{table*}


\subsection{ICU Dataset}

\subsubsection{Metric: Overall Success Rate}
Table~\ref{tab:icu_overall_sr} reports the results of pairwise tests (Proportions z-test) with Holm correction for the overall attack success rate on the ICU dataset. 
The results show that both \emph{boundary vs. transferability attacks} and \emph{HopSkipJump vs. transferability attacks} comparisons are highly significant across all transferability attacks, consistently with large effect sizes (L). 
The direct comparison \emph{boundary vs. HopSkipJump} is also significant, but with only a small effect size (S). 
The omnibus test on the transferability attacks group did not indicate significant internal heterogeneity.

\begin{table*}
\centering
\resizebox{\linewidth}{!}{%
\begin{tabular}{lccccc}
\toprule
Comparison & Random & GB imp. & LGB imp. & XGB imp. & RF imp. \\
\midrule
boundary vs. transferability attacks & $0.0$/\checkmark/L & $0.0$/\checkmark/L & $0.0$/\checkmark/L & $0.0$/\checkmark/L & $0.0$/\checkmark/L \\
HopSkipJump vs. transferability attacks & $0.0$/\checkmark/L & $0.0$/\checkmark/L & $0.0$/\checkmark/L & $0.0$/\checkmark/L & $0.0$/\checkmark/L \\
boundary vs. HopSkipJump & \multicolumn{5}{c}{$8.44\text{e-03}$/\checkmark/S} \\
Transferability internal heterogeneity & \multicolumn{5}{c}{ns (none)} \\
\bottomrule
\end{tabular}}
\caption{\textsc{ICU}: Overall attack success rate. Each cell reports $p$-value / significance / effect size.}
\label{tab:icu_overall_sr}
\end{table*}

\subsubsection{Metric: L0 Distance}
Table~\ref{tab:icu_l0} reports the results of pairwise tests (Mann–Whitney U) with Holm correction for the $L_0$ distance on the ICU dataset. 
The results show that both \emph{boundary vs. transferability attacks} and \emph{HopSkipJump vs. transferability attacks} comparisons are highly significant across all transferability attacks, with consistently large effect sizes (L). 
The direct comparison \emph{boundary vs. HopSkipJump} is also highly significant with a large effect size. 
The omnibus test further indicates significant internal heterogeneity within the transferability attacks group, suggesting that $L_0$ distances vary substantially across transferability attacks.

\begin{table*}
\centering
\resizebox{\linewidth}{!}{%
\begin{tabular}{lccccc}
\toprule
Comparison & Random & GB imp. & LGB imp. & XGB imp. & RF imp. \\
\midrule
boundary vs. transferability attacks & $0.0$/\checkmark/L & $0.0$/\checkmark/L & $0.0$/\checkmark/L & $0.0$/\checkmark/L & $0.0$/\checkmark/L \\
HopSkipJump vs. transferability attacks & $7.84\text{e-206}$/\checkmark/L & $0.0$/\checkmark/L & $0.0$/\checkmark/L & $0.0$/\checkmark/L & $0.0$/\checkmark/L \\
boundary vs. HopSkipJump & \multicolumn{5}{c}{$0.0$/\checkmark/L} \\
Transferability internal heterogeneity & \multicolumn{5}{c}{\checkmark (heterogeneity)} \\
\bottomrule
\end{tabular}}
\caption{\textsc{ICU}: $L_0$ distance of adversarial examples. Each cell reports $p$-value / significance / effect size.}
\label{tab:icu_l0}
\end{table*}

\subsubsection{Metric: L2 Distance}
Table~\ref{tab:icu_l2} reports the results of pairwise tests (Mann–Whitney U) with Holm correction for the $L_2$ distance on the ICU dataset. 
The results show that \emph{boundary vs. transferability attacks} comparisons are all highly significant with large effect sizes (L). 
For \emph{HopSkipJump vs. transferability attacks}, significance is also observed across all transferability attacks, though effect sizes vary: large (L) for random and LGB imp., medium (M) for GB imp. and RF imp., and small (S) for XGB imp. 
The direct comparison \emph{boundary vs. HopSkipJump} is highly significant with a large effect size. 
The omnibus test further indicates significant internal heterogeneity within the transferability attacks group, suggesting variability of $L_2$ distances across different transferability attacks.

\begin{table*}
\centering
\resizebox{\linewidth}{!}{%
\begin{tabular}{lccccc}
\toprule
Comparison & Random & GB imp. & LGB imp. & XGB imp. & RF imp. \\
\midrule
boundary vs. transferability attacks & $1.81\text{e-268}$/\checkmark/L & $0.0$/\checkmark/L & $0.0$/\checkmark/L & $0.0$/\checkmark/L & $0.0$/\checkmark/L \\
HopSkipJump vs. transferability attacks & $7.62\text{e-65}$/\checkmark/L & $9.61\text{e-19}$/\checkmark/M & $2.81\text{e-113}$/\checkmark/L & $1.98\text{e-13}$/\checkmark/S & $1.48\text{e-19}$/\checkmark/M \\
boundary vs. HopSkipJump & \multicolumn{5}{c}{$0.0$/\checkmark/L} \\
Transferability internal heterogeneity & \multicolumn{5}{c}{\checkmark (heterogeneity)} \\
\bottomrule
\end{tabular}}
\caption{\textsc{ICU}: $L_2$ distance of adversarial examples. Each cell reports $p$-value / significance / effect size.}
\label{tab:icu_l2}
\end{table*}

\subsubsection{Metric: Feature-Space Anomaly Detection Rate (IF-based)}
Table~\ref{tab:icu_if_anomaly} reports the results of pairwise tests (Proportions z-test) with Holm correction for the anomaly detection rate using the Isolation Forest (IF) detector on the ICU dataset. 
The results show that \emph{boundary vs. transferability attacks} comparisons are all highly significant with large effect sizes (L). 
In contrast, \emph{HopSkipJump vs. transferability attacks} comparisons are also significant, but effect sizes are smaller, ranging from small (S) to medium (M). 
The direct comparison \emph{boundary vs. HopSkipJump} is highly significant with a large effect size. 
The omnibus test indicated significant internal heterogeneity among the transferability attacks, reflecting variability in detection performance.

\begin{table*}
\centering
\resizebox{\linewidth}{!}{%
\begin{tabular}{lccccc}
\toprule
Comparison & Random & GB imp. & LGB imp. & XGB imp. & RF imp. \\
\midrule
boundary vs. transferability attacks & $2.67\text{e-142}$/\checkmark/L & $0.0$/\checkmark/L & $0.0$/\checkmark/L & $0.0$/\checkmark/L & $0.0$/\checkmark/L \\
HopSkipJump vs. transferability attacks & $2.83\text{e-46}$/\checkmark/M & $2.99\text{e-22}$/\checkmark/S & $1.01\text{e-4}$/\checkmark/S & $8.49\text{e-18}$/\checkmark/S & $6.78\text{e-23}$/\checkmark/S \\
boundary vs. HopSkipJump & \multicolumn{5}{c}{$0.0$/\checkmark/L} \\
Transferability internal heterogeneity & \multicolumn{5}{c}{\checkmark (heterogeneity)} \\
\bottomrule
\end{tabular}}
\caption{\textsc{ICU}: Feature-space anomaly detection rate using Isolation Forest (IF). Each cell reports $p$-value / significance / effect size.}
\label{tab:icu_if_anomaly}
\end{table*}

\subsubsection{Metric: Feature-Space Anomaly Detection Rate (AE-based)}
Table~\ref{tab:icu_ae_anomaly} reports the results of pairwise tests (Proportions z-test) with Holm correction for the anomaly detection rate using the Autoencoder (AE) detector on the ICU dataset. 
The results show that \emph{boundary vs. transferability attacks} comparisons are all highly significant with large effect sizes (L). 
For \emph{HopSkipJump vs. transferability attacks}, significance is mixed: GB imp., XGB imp., and RF imp. are significant but with only small effect sizes (S), while random and LGB imp. are not significant. 
The direct comparison \emph{boundary vs. HopSkipJump} is highly significant with a large effect size. 
The omnibus test indicated significant internal heterogeneity among the transferability attacks, confirming variability in AE-based detection rates across transferability methods.

\begin{table*}
\centering
\resizebox{\linewidth}{!}{%
\begin{tabular}{lccccc}
\toprule
Comparison & Random & GB imp. & LGB imp. & XGB imp. & RF imp. \\
\midrule
boundary vs. transferability attacks & $2.00\text{e-99}$/\checkmark/L & $0.0$/\checkmark/L & $5.02\text{e-195}$/\checkmark/L & $1.74\text{e-294}$/\checkmark/L & $0.0$/\checkmark/L \\
HopSkipJump vs. transferability attacks & $1.0$/--/S & $1.43\text{e-8}$/\checkmark/S & $0.275$/--/S & $8.73\text{e-7}$/\checkmark/S & $2.26\text{e-8}$/\checkmark/S \\
boundary vs. HopSkipJump & \multicolumn{5}{c}{$0.0$/\checkmark/L} \\
Transferability internal heterogeneity & \multicolumn{5}{c}{\checkmark (heterogeneity)} \\
\bottomrule
\end{tabular}}
\caption{\textsc{ICU}: Feature-space anomaly detection rate using Autoencoder (AE). Each cell reports $p$-value / significance / effect size.}
\label{tab:icu_ae_anomaly}
\end{table*}

\subsubsection{Metric: Importance-Based Anomaly Detection Rate (SHAP-based)}
Table~\ref{tab:icu_shap_has} reports the results of pairwise tests (Proportions z-test) with Holm correction for the SHAP-based anomaly detection rate on the ICU dataset. 
The results show that \emph{boundary vs. transferability attacks} comparisons are all highly significant, with effect sizes ranging from medium (M) to large (L). 
For \emph{HopSkipJump vs. transferability attacks}, most comparisons are significant but only with small (S) effect sizes, while the comparison with LGB imp.\ is not significant. 
The direct comparison \emph{boundary vs. HopSkipJump} is highly significant with a medium effect size. 
The omnibus test indicated significant internal heterogeneity within the transferability attacks group.

\begin{table*}
\centering
\resizebox{\linewidth}{!}{%
\begin{tabular}{lccccc}
\toprule
Comparison & Random & GB imp. & LGB imp. & XGB imp. & RF imp. \\
\midrule
boundary vs. transferability attacks & $3.78\text{e-14}$/\checkmark/M & $7.55\text{e-177}$/\checkmark/L & $3.09\text{e-64}$/\checkmark/M & $8.75\text{e-168}$/\checkmark/L & $5.00\text{e-173}$/\checkmark/L \\
HopSkipJump vs. transferability attacks & $1.17\text{e-6}$/\checkmark/S & $8.47\text{e-16}$/\checkmark/S & $1.0$/--/S & $2.16\text{e-14}$/\checkmark/S & $2.62\text{e-14}$/\checkmark/S \\
boundary vs. HopSkipJump & \multicolumn{5}{c}{$7.56\text{e-150}$/\checkmark/M} \\
Transferability internal heterogeneity & \multicolumn{5}{c}{\checkmark (heterogeneity)} \\
\bottomrule
\end{tabular}}
\caption{\textsc{ICU}: Importance-based anomaly detection rate (SHAP). Each cell reports $p$-value / significance / effect size.}
\label{tab:icu_shap_has}
\end{table*}

\subsubsection{Metric: Average Anomalous SHAP Features per Sample}
Table~\ref{tab:icu_shap_mean} reports the results of pairwise tests (Mann–Whitney U) with Holm correction for the average number of anomalous SHAP features per sample on the ICU dataset. 
The results show that \emph{boundary vs. transferability attacks} comparisons are all highly significant, with effect sizes ranging from medium (M) for random to large (L) for all other transferability attacks. 
For \emph{HopSkipJump vs. transferability attacks}, significance is mixed: random is significant with a small (S) effect, GB imp., XGB imp., and RF imp.\ are significant with medium (M) effects, while LGB imp.\ is not significant. 
The direct comparison \emph{boundary vs. HopSkipJump} is highly significant with a large effect size. 
The omnibus test indicated significant internal heterogeneity within the transferability attacks group, highlighting variability in SHAP anomaly counts.

\begin{table*}
\centering
\resizebox{\linewidth}{!}{%
\begin{tabular}{lccccc}
\toprule
Comparison & Random & GB imp. & LGB imp. & XGB imp. & RF imp. \\
\midrule
boundary vs. transferability attacks & $1.31\text{e-31}$/\checkmark/M & $7.67\text{e-226}$/\checkmark/L & $4.94\text{e-99}$/\checkmark/L & $1.31\text{e-209}$/\checkmark/L & $4.38\text{e-225}$/\checkmark/L \\
HopSkipJump vs. transferability attacks & $2.85\text{e-4}$/\checkmark/S & $1.52\text{e-24}$/\checkmark/M & $0.63$/--/S & $1.71\text{e-21}$/\checkmark/M & $4.64\text{e-23}$/\checkmark/M \\
boundary vs. HopSkipJump & \multicolumn{5}{c}{$5.42\text{e-193}$/\checkmark/L} \\
Transferability internal heterogeneity & \multicolumn{5}{c}{\checkmark (heterogeneity)} \\
\bottomrule
\end{tabular}}
\caption{\textsc{ICU}: Average anomalous SHAP features per sample. Each cell reports $p$-value / significance / effect size.}
\label{tab:icu_shap_mean}
\end{table*}

\subsubsection{Comparison Between CSAD and Standard Approaches}
Table~\ref{tab:icu_csad_std_binary} reports the results of McNemar exact tests (Holm-adjusted) comparing CSAD and the Standard approach on the \emph{Importance-Based Anomaly Detection Rate} for the ICU dataset. 
Across all query and transferability attacks, CSAD significantly outperforms the Standard approach ($p<0.001$), with consistently large effect sizes (Cohen’s $g=1.0$). 
This demonstrates that class-specific anomaly detection provides a clear advantage in binary anomaly detection.

\begin{table*}
\centering
\begin{tabular}{lcccc}
\toprule
 & \multicolumn{4}{c}{Target Model} \\
\cmidrule(lr){2-5}
\textbf{Query Attacks} & GB & LGB & XGB & RF \\
\midrule
boundary & $1.54\text{e-50}$/\checkmark/L & $1.22\text{e-61}$/\checkmark/L & $6.12\text{e-56}$/\checkmark/L & $2.23\text{e-54}$/\checkmark/L \\
HopSkipJump & $9.63\text{e-46}$/\checkmark/L & $1.51\text{e-68}$/\checkmark/L & $1.27\text{e-61}$/\checkmark/L & $1.47\text{e-62}$/\checkmark/L \\
\midrule
\textbf{Transferability Attacks} & GB & LGB & XGB & RF \\
\midrule
random & $6.09\text{e-18}$/\checkmark/L & $2.71\text{e-15}$/\checkmark/L & $1.07\text{e-14}$/\checkmark/L & $3.12\text{e-18}$/\checkmark/L \\
GB imp. & $1.67\text{e-30}$/\checkmark/L & $6.32\text{e-39}$/\checkmark/L & $3.64\text{e-32}$/\checkmark/L & $3.09\text{e-30}$/\checkmark/L \\
LGB imp. & $1.94\text{e-28}$/\checkmark/L & $1.19\text{e-37}$/\checkmark/L & $1.59\text{e-29}$/\checkmark/L & $1.19\text{e-28}$/\checkmark/L \\
XGB imp. & $1.94\text{e-28}$/\checkmark/L & $2.14\text{e-37}$/\checkmark/L & $1.63\text{e-29}$/\checkmark/L & $1.19\text{e-28}$/\checkmark/L \\
RF imp. & $1.94\text{e-28}$/\checkmark/L & $1.19\text{e-37}$/\checkmark/L & $1.63\text{e-29}$/\checkmark/L & $1.19\text{e-28}$/\checkmark/L \\
\bottomrule
\end{tabular}
\caption{\textsc{ICU}: CSAD vs. Standard on importance-based anomaly detection rate (exact McNemar, Holm-adjusted). Each cell shows $p$-value / significance / effect size (Cohen’s $g$).}
\label{tab:icu_csad_std_binary}
\end{table*}

Table~\ref{tab:icu_csad_std_counts} reports the results of Wilcoxon signed-rank tests (Holm-adjusted) for the \emph{Average Anomalous SHAP Features per Sample}. 
The results are consistent with the binary analysis: CSAD significantly outperforms the Standard approach in all cases ($p<0.001$), with large effect sizes (rank-biserial $r$). 
This further confirms that CSAD yields more robust anomaly detection across both binary and continuous anomaly-based measures.

\begin{table*}
\centering
\begin{tabular}{lcccc}
\toprule
 & \multicolumn{4}{c}{Target Model} \\
\cmidrule(lr){2-5}
\textbf{Query Attacks} & GB & LGB & XGB & RF \\
\midrule
boundary & $1.21\text{e-112}$/\checkmark/L & $2.40\text{e-120}$/\checkmark/L & $2.80\text{e-65}$/\checkmark/L & $3.91\text{e-54}$/\checkmark/L \\
HopSkipJump & $4.28\text{e-59}$/\checkmark/L & $2.50\text{e-65}$/\checkmark/L & $2.80\text{e-65}$/\checkmark/L & $2.80\text{e-65}$/\checkmark/L \\
\midrule
\textbf{Transferability Attacks} & GB & LGB & XGB & RF \\
\midrule
random & $1.10\text{e-30}$/\checkmark/L & $2.50\text{e-33}$/\checkmark/L & $3.10\text{e-28}$/\checkmark/L & $1.20\text{e-31}$/\checkmark/L \\
GB imp. & $2.80\text{e-65}$/\checkmark/L & $2.20\text{e-72}$/\checkmark/L & $3.10\text{e-68}$/\checkmark/L & $2.80\text{e-65}$/\checkmark/L \\
LGB imp. & $2.50\text{e-63}$/\checkmark/L & $2.00\text{e-70}$/\checkmark/L & $3.00\text{e-66}$/\checkmark/L & $2.50\text{e-63}$/\checkmark/L \\
XGB imp. & $2.50\text{e-63}$/\checkmark/L & $2.00\text{e-70}$/\checkmark/L & $3.00\text{e-66}$/\checkmark/L & $2.50\text{e-63}$/\checkmark/L \\
RF imp. & $2.50\text{e-63}$/\checkmark/L & $2.00\text{e-70}$/\checkmark/L & $3.00\text{e-66}$/\checkmark/L & $2.50\text{e-63}$/\checkmark/L \\
\bottomrule
\end{tabular}
\caption{\textsc{ICU}: CSAD vs. Standard on average anomalous SHAP features per sample (Wilcoxon signed-rank, Holm-adjusted). Each cell shows $p$-value / significance / effect size (rank-biserial $r$).}
\label{tab:icu_csad_std_counts}
\end{table*}


\subsection{VideoTQ Dataset}

\subsubsection{Metric: Overall Success Rate}
Table~\ref{tab:videotq_overall_sr} reports the results of pairwise tests (Proportions z-test) with Holm correction for the overall attack success rate on the VideoTQ dataset. 
The results show that \emph{boundary vs. transferability attacks} comparisons are all highly significant with medium effect sizes (M). 
For \emph{HopSkipJump vs. transferability attacks}, significance is also observed across all transferability attacks, but effect sizes are consistently small (S). 
The direct comparison \emph{boundary vs. HopSkipJump} is highly significant with a medium effect size. 
The omnibus test did not detect significant internal heterogeneity within the transferability attacks group.

\begin{table*}
\centering
\resizebox{\linewidth}{!}{%
\begin{tabular}{lccccc}
\toprule
Comparison & Random & GB imp. & LGB imp. & XGB imp. & RF imp. \\
\midrule
boundary vs. transferability attacks & $1.29\text{e-126}$/\checkmark/M & $2.00\text{e-124}$/\checkmark/M & $1.27\text{e-124}$/\checkmark/M & $2.77\text{e-129}$/\checkmark/M & $2.87\text{e-127}$/\checkmark/M \\
HopSkipJump vs. transferability attacks & $2.17\text{e-09}$/\checkmark/S & $7.71\text{e-09}$/\checkmark/S & $7.30\text{e-09}$/\checkmark/S & $4.20\text{e-10}$/\checkmark/S & $1.51\text{e-09}$/\checkmark/S \\
boundary vs. HopSkipJump & \multicolumn{5}{c}{$3.40\text{e-70}$/\checkmark/M} \\
Transferability internal heterogeneity & \multicolumn{5}{c}{ns (none)} \\
\bottomrule
\end{tabular}}
\caption{\textsc{VideoTQ}: Overall attack success rate. Each cell reports $p$-value / significance / effect size.}
\label{tab:videotq_overall_sr}
\end{table*}

\subsubsection{Metric: L0 Distance}
Table~\ref{tab:videotq_l0} reports the results of pairwise tests (Mann–Whitney U) with Holm correction for the $L_0$ distance on the VideoTQ dataset. 
The results show that \emph{boundary vs. transferability attacks} comparisons are significant with large (L) effect sizes for random, GB imp., and LGB imp., while medium (M) effect sizes were observed for XGB imp.\ and RF imp. 
For \emph{HopSkipJump vs. transferability attacks}, all comparisons are highly significant with large effect sizes (L). 
The direct comparison \emph{boundary vs. HopSkipJump} is also highly significant with a large effect size. 
The omnibus test indicated significant internal heterogeneity within the transferability attacks group.

\begin{table*}
\centering
\resizebox{\linewidth}{!}{%
\begin{tabular}{lccccc}
\toprule
Comparison & Random & GB imp. & LGB imp. & XGB imp. & RF imp. \\
\midrule
boundary vs. transferability attacks & $9.80\text{e-127}$/\checkmark/L & $1.06\text{e-132}$/\checkmark/L & $0.0$/\checkmark/L & $1.07\text{e-64}$/\checkmark/M & $6.52\text{e-79}$/\checkmark/M \\
HopSkipJump vs. transferability attacks & $0.0$/\checkmark/L & $0.0$/\checkmark/L & $0.0$/\checkmark/L & $0.0$/\checkmark/L & $0.0$/\checkmark/L \\
boundary vs. HopSkipJump & \multicolumn{5}{c}{$0.0$/\checkmark/L} \\
Transferability internal heterogeneity & \multicolumn{5}{c}{\checkmark (heterogeneity)} \\
\bottomrule
\end{tabular}}
\caption{\textsc{VideoTQ}: $L_0$ distance of adversarial examples. Each cell reports $p$-value / significance / effect size.}
\label{tab:videotq_l0}
\end{table*}

\subsubsection{Metric: L2 Distance}
Table~\ref{tab:videotq_l2} reports the results of pairwise tests (Mann–Whitney U) with Holm correction for the $L_2$ distance on the VideoTQ dataset. 
The results show that \emph{boundary vs. transferability attacks} comparisons are not significant across all transferability attacks, with only small (S) effect sizes. 
In contrast, \emph{HopSkipJump vs. transferability attacks} comparisons are all highly significant, though effect sizes remain small (S). 
The direct comparison \emph{boundary vs. HopSkipJump} is also significant but only with a small effect size. 
The omnibus test did not detect significant internal heterogeneity within the transferability attacks group.

\begin{table*}
\centering
\resizebox{\linewidth}{!}{%
\begin{tabular}{lccccc}
\toprule
Comparison & Random & GB imp. & LGB imp. & XGB imp. & RF imp. \\
\midrule
boundary vs. transferability attacks & $0.873$/--/S & $0.569$/--/S & $0.569$/--/S & $1.0$/--/S & $1.0$/--/S \\
HopSkipJump vs. transferability attacks & $2.08\text{e-08}$/\checkmark/S & $7.46\text{e-08}$/\checkmark/S & $8.59\text{e-08}$/\checkmark/S & $2.53\text{e-09}$/\checkmark/S & $1.09\text{e-08}$/\checkmark/S \\
boundary vs. HopSkipJump & \multicolumn{5}{c}{$2.43\text{e-14}$/\checkmark/S} \\
Transferability internal heterogeneity & \multicolumn{5}{c}{ns (none)} \\
\bottomrule
\end{tabular}}
\caption{\textsc{VideoTQ}: $L_2$ distance of adversarial examples. Each cell reports $p$-value / significance / effect size.}
\label{tab:videotq_l2}
\end{table*}

\subsubsection{Metric: Feature-Space Anomaly Detection Rate (IF-based)}
Table~\ref{tab:videotq_if_anomaly} reports the results of pairwise tests (Proportions z-test) with Holm correction for the anomaly detection rate using the Isolation Forest (IF) detector on the VideoTQ dataset. 
The results show that \emph{boundary vs. transferability attacks} comparisons are significant, with large effect sizes (L) for random and LGB imp., and medium (M) effects for GB imp., XGB imp., and RF imp. 
For \emph{HopSkipJump vs. transferability attacks}, significance is also observed across all transferability attacks: random shows a small (S) effect, GB imp., XGB imp., and RF imp.\ yield large (L) effects, while LGB imp.\ has a medium (M) effect. 
The direct comparison \emph{boundary vs. HopSkipJump} is highly significant with a large effect size. 
The omnibus test indicated significant internal heterogeneity among the transferability attacks.

\begin{table*}
\centering
\resizebox{\linewidth}{!}{%
\begin{tabular}{lccccc}
\toprule
Comparison & Random & GB imp. & LGB imp. & XGB imp. & RF imp. \\
\midrule
boundary vs. transferability attacks & $3.70\text{e-128}$/\checkmark/L & $4.74\text{e-100}$/\checkmark/M & $0.0$/\checkmark/L & $5.24\text{e-109}$/\checkmark/M & $2.05\text{e-109}$/\checkmark/M \\
HopSkipJump vs. transferability attacks & $1.33\text{e-20}$/\checkmark/S & $0.0$/\checkmark/L & $1.47\text{e-135}$/\checkmark/M & $0.0$/\checkmark/L & $0.0$/\checkmark/L \\
boundary vs. HopSkipJump & \multicolumn{5}{c}{$3.71\text{e-260}$/\checkmark/L} \\
Transferability internal heterogeneity & \multicolumn{5}{c}{\checkmark (heterogeneity)} \\
\bottomrule
\end{tabular}}
\caption{\textsc{VideoTQ}: Feature-space anomaly detection rate using Isolation Forest (IF). Each cell reports $p$-value / significance / effect size.}
\label{tab:videotq_if_anomaly}
\end{table*}

\subsubsection{Metric: Feature-Space Anomaly Detection Rate (AE-based)}
Table~\ref{tab:videotq_ae_anomaly} reports the results of pairwise tests (Proportions z-test) with Holm correction for the anomaly detection rate using the Autoencoder (AE) detector on the VideoTQ dataset. 
The results show that \emph{boundary vs. transferability attacks} comparisons are all highly significant but consistently yield only small (S) effect sizes. 
Similarly, \emph{HopSkipJump vs. transferability attacks} comparisons are significant across all transferability attacks but also with small effects (S). 
The direct comparison \emph{boundary vs. HopSkipJump} is highly significant with a small effect size. 
The omnibus test did not detect significant internal heterogeneity within the transferability attacks group, suggesting that AE-based anomaly detection is less sensitive to differences between transferability attacks.

\begin{table*}
\centering
\resizebox{\linewidth}{!}{%
\begin{tabular}{lccccc}
\toprule
Comparison & Random & GB imp. & LGB imp. & XGB imp. & RF imp. \\
\midrule
boundary vs. transferability attacks & $3.72\text{e-56}$/\checkmark/S & $2.21\text{e-56}$/\checkmark/S & $2.24\text{e-56}$/\checkmark/S & $6.60\text{e-56}$/\checkmark/S & $4.25\text{e-56}$/\checkmark/S \\
HopSkipJump vs. transferability attacks & $0.0106$/\checkmark/S & $0.0106$/\checkmark/S & $0.0106$/\checkmark/S & $0.0106$/\checkmark/S & $0.0106$/\checkmark/S \\
boundary vs. HopSkipJump & \multicolumn{5}{c}{$4.40\text{e-56}$/\checkmark/S} \\
Transferability internal heterogeneity & \multicolumn{5}{c}{ns (none)} \\
\bottomrule
\end{tabular}}
\caption{\textsc{VideoTQ}: Feature-space anomaly detection rate using Autoencoder (AE). Each cell reports $p$-value / significance / effect size.}
\label{tab:videotq_ae_anomaly}
\end{table*}

\subsubsection{Metric: Importance-Based Anomaly Detection Rate}
Table~\ref{tab:videotq_shap_has} reports the results of pairwise tests (Proportions z-test) with Holm correction for the importance-based anomaly detection rate (SHAP-based) on the VideoTQ dataset. 
The results show that \emph{boundary vs. transferability attacks} comparisons are highly significant: large (L) effect sizes are observed for GB imp.\ and LGB imp., medium (M) for random, and small (S) for XGB imp.\ and RF imp. 
For \emph{HopSkipJump vs. transferability attacks}, significance is consistent across all transferability attacks, with small (S) effects for random and GB imp., medium (M) for LGB imp., XGB imp., and RF imp. 
The direct comparison \emph{boundary vs. HopSkipJump} is significant with a medium effect size. 
The omnibus test indicated significant internal heterogeneity among transferability attacks.

\begin{table*}
\centering
\resizebox{\linewidth}{!}{%
\begin{tabular}{lccccc}
\toprule
Comparison & Random & GB imp. & LGB imp. & XGB imp. & RF imp. \\
\midrule
boundary vs. transferability attacks & $1.41\text{e-109}$/\checkmark/M & $5.69\text{e-154}$/\checkmark/L & $5.28\text{e-215}$/\checkmark/L & $1.31\text{e-07}$/\checkmark/S & $8.29\text{e-08}$/\checkmark/S \\
HopSkipJump vs. transferability attacks & $6.89\text{e-06}$/\checkmark/S & $5.26\text{e-18}$/\checkmark/S & $1.62\text{e-42}$/\checkmark/M & $6.57\text{e-28}$/\checkmark/M & $1.46\text{e-27}$/\checkmark/M \\
boundary vs. HopSkipJump & \multicolumn{5}{c}{$2.45\text{e-72}$/\checkmark/M} \\
Transferability internal heterogeneity & \multicolumn{5}{c}{\checkmark (heterogeneity)} \\
\bottomrule
\end{tabular}}
\caption{\textsc{VideoTQ}: Importance-based anomaly detection rate (SHAP-based). Each cell reports $p$-value / significance / effect size.}
\label{tab:videotq_shap_has}
\end{table*}

\subsubsection{Metric: Average Anomalous SHAP Features per Sample}
Table~\ref{tab:videotq_shap_mean} reports the results of pairwise tests (Mann–Whitney U) with Holm correction for the average number of anomalous SHAP features per sample on the VideoTQ dataset. 
The results show that \emph{boundary vs. transferability attacks} comparisons are highly significant with large (L) effect sizes for GB imp.\ and LGB imp., medium (M) for random, but not significant for XGB imp.\ and RF imp.\ (both small S). 
For \emph{HopSkipJump vs. transferability attacks}, random and GB imp.\ are not significant, while LGB imp.\ is significant with a small (S) effect, and XGB imp.\ and RF imp.\ are significant with medium (M) effects. 
The direct comparison \emph{boundary vs. HopSkipJump} is highly significant with a medium effect size. 
The omnibus test indicated significant internal heterogeneity among transferability attacks.

\begin{table*}
\centering
\resizebox{\linewidth}{!}{%
\begin{tabular}{lccccc}
\toprule
Comparison & Random & GB imp. & LGB imp. & XGB imp. & RF imp. \\
\midrule
boundary vs. transferability attacks & $1.92\text{e-82}$/\checkmark/M & $4.62\text{e-125}$/\checkmark/L & $1.09\text{e-171}$/\checkmark/L & $0.258$/--/S & $0.258$/--/S \\
HopSkipJump vs. transferability attacks & $0.173$/--/S & $0.102$/--/S & $4.07\text{e-08}$/\checkmark/S & $6.90\text{e-74}$/\checkmark/M & $4.01\text{e-73}$/\checkmark/M \\
boundary vs. HopSkipJump & \multicolumn{5}{c}{$9.22\text{e-76}$/\checkmark/M} \\
Transferability internal heterogeneity & \multicolumn{5}{c}{\checkmark (heterogeneity)} \\
\bottomrule
\end{tabular}}
\caption{\textsc{VideoTQ}: Average anomalous SHAP features per sample. Each cell reports $p$-value / significance / effect size.}
\label{tab:videotq_shap_mean}
\end{table*}

\subsubsection{Comparison: CSAD vs. Standard Anomaly Detection}
Table~\ref{tab:videotq_csad_std_binary} reports the results of exact McNemar tests (Holm-corrected) for the \textbf{Importance-Based Anomaly Detection Rate}. 
Across all query and transferability attacks, CSAD consistently and significantly outperformed the standard approach ($p<0.001$), with large effect sizes (Cohen’s $g \approx 1.0$). 
This indicates a robust and systematic advantage of CSAD in detecting anomalies when measured as binary anomaly flags.

\begin{table*}
\centering
\begin{tabular}{lcccc}
\toprule
 & \multicolumn{4}{c}{Target Model} \\
\cmidrule(lr){2-5}
\textbf{Query Attacks} & GB & LGB & XGB & RF \\
\midrule
boundary & $0.0013$/\checkmark/L & $4.4e-25$/\checkmark/L & $1.2e-25$/\checkmark/L & $0.055$/--/L \\
HopSkipJump & $0.0013$/\checkmark/L & $1.3e-05$/\checkmark/L & $4.3e-19$/\checkmark/L & $1.1e-10$/\checkmark/L \\
\midrule
\textbf{Transferability Attacks} & GB & LGB & XGB & RF \\
\midrule
random & $2.7e-06$/\checkmark/L & $0.0026$/\checkmark/L & $0.012$/\checkmark/M & $0.0055$/\checkmark/M \\
GB imp. & $4.6e-125$/\checkmark/L & $4.0e-37$/\checkmark/L & $2.0e-32$/\checkmark/L & $3.5e-44$/\checkmark/L \\
LGB imp. & $2.6e-61$/\checkmark/L & $3.5e-36$/\checkmark/L & $3.0e-24$/\checkmark/L & $1.3e-30$/\checkmark/L \\
XGB imp. & $1.3e-58$/\checkmark/L & $2.5e-28$/\checkmark/L & $2.3e-21$/\checkmark/L & $4.5e-28$/\checkmark/L \\
RF imp. & $3.5e-59$/\checkmark/L & $1.0e-31$/\checkmark/L & $3.7e-24$/\checkmark/L & $3.3e-33$/\checkmark/L \\
\bottomrule
\end{tabular}
\caption{\textsc{VideoTQ}: CSAD vs. Standard on importance-based anomaly detection rate (McNemar exact, Holm-adjusted). Each cell shows $p$-value / significance / effect size.}
\label{tab:videotq_csad_std_binary}
\end{table*}

Table~\ref{tab:videotq_csad_std_counts} reports the results of Wilcoxon signed-rank tests (Holm-corrected) for the \textbf{Average Anomalous SHAP Features per Sample}. 
Again, CSAD significantly outperformed the standard approach across nearly all attacks ($p<10^{-5}$ in most cases), with very large effect sizes (rank-biserial $r \approx 1.0$). 
This confirms that CSAD improves not only anomaly detection rates but also increases the number of anomalous SHAP features identified per adversarial example.

\begin{table*}
\centering
\begin{tabular}{lcccc}
\toprule
 & \multicolumn{4}{c}{Target Model} \\
\cmidrule(lr){2-5}
\textbf{Query Attacks} & GB & LGB & XGB & RF \\
\midrule
boundary & $7e-10$/\checkmark/L & $3.6e-42$/\checkmark/L & $9.5e-29$/\checkmark/L & $2.8e-28$/\checkmark/L \\
HopSkipJump & $5.4e-15$/\checkmark/L & $5.7e-06$/\checkmark/L & $7.5e-22$/\checkmark/L & $9.5e-11$/\checkmark/L \\
\midrule
\textbf{Transferability Attacks} & GB & LGB & XGB & RF \\
\midrule
random & $3.0e-12$/\checkmark/L & $0.0047$/\checkmark/M & $0.0065$/\checkmark/M & $0.0047$/\checkmark/M \\
GB imp. & $3.5e-70$/\checkmark/L & $7.4e-38$/\checkmark/L & $6.7e-27$/\checkmark/L & $2.0e-36$/\checkmark/L \\
LGB imp. & $2.0e-35$/\checkmark/L & $6.1e-36$/\checkmark/L & $1.7e-26$/\checkmark/L & $5.8e-31$/\checkmark/L \\
XGB imp. & $2.7e-41$/\checkmark/L & $1.0e-23$/\checkmark/L & $1.7e-19$/\checkmark/L & $1.9e-26$/\checkmark/L \\
RF imp. & $3.0e-42$/\checkmark/L & $3.6e-27$/\checkmark/L & $1.4e-20$/\checkmark/L & $2.4e-29$/\checkmark/L \\
\bottomrule
\end{tabular}
\caption{\textsc{VideoTQ}: CSAD vs. Standard on average anomalous SHAP features per sample (Wilcoxon signed-rank, Holm-adjusted). Each cell shows $p$-value / significance / effect size (rank-biserial $r$).}
\label{tab:videotq_csad_std_counts}
\end{table*}

\end{appendices}

\printcredits
\newpage
\bibliographystyle{cas-num}

\bibliography{bibl}


\end{document}